\documentclass{article}

\PassOptionsToPackage{numbers, compress}{natbib}
\usepackage[final]{neurips_2023}

\usepackage[utf8]{inputenc} 
\usepackage[T1]{fontenc}    
\usepackage{hyperref}       
\usepackage{url}            
\usepackage{booktabs}       
\usepackage{amsfonts}       
\usepackage{nicefrac}       
\usepackage{microtype}      
\usepackage{xcolor}         
\usepackage{graphicx}
\usepackage{wrapfig}
\usepackage{amsmath}
\usepackage{caption}
\usepackage{subcaption}
\usepackage{authblk}
\usepackage{listings}

\title{STG-MTL: Scalable Task Grouping\\For Multi-Task Learning Using Data Maps}

\author[1]{Ammar Sherif \thanks{Corresponding Author <ammarsherif90 [at] gmail [dot] com>}\ \ }
\author[2]{Abubakar Abid}
\author[1]{Mustafa Elattar}
\author[1]{Mohamed ElHelw}
\affil[1]{Nile University}
\affil[2]{Hugging Face}

\begin{document}

\maketitle

\begin{abstract}
  Multi-Task Learning (MTL) is a powerful technique that has gained popularity due to its performance improvement over traditional Single-Task Learning (STL). However, MTL is often challenging because there is an exponential number of possible task groupings, which can make it difficult to choose the best one because some groupings might produce performance degradation due to negative interference between tasks. That is why existing solutions are severely suffering from scalability issues, limiting any practical application. In our paper, we propose a new data-driven method that addresses these challenges and provides a scalable and modular solution for classification task grouping based on a re-proposed data-driven features, Data Maps, which capture the training dynamics for each classification task during the MTL training. Through a theoretical comparison with other techniques, we manage to show that our approach has the superior scalability. Our experiments show a better performance and verify the method's effectiveness, even on an unprecedented number of tasks (up to 100 tasks on CIFAR100). Being the first to work on such number of tasks, our comparisons on the resulting grouping shows similar grouping to the mentioned in the dataset, CIFAR100. Finally, we provide a modular implementation\footnote{\url{https://github.com/ammarSherif/STG-MTL}} for easier integration and testing, with examples from multiple datasets and tasks.
\end{abstract}

\section{Introduction}
Multi-Task Learning (MTL) has emerged as a powerful technique in deep learning \cite{MTL_Survey,crawshaw2020MTL_DL} that allows for joint training of multiple related tasks, leading to improved model performance compared to traditional Single-Task Learning (STL). By leveraging shared representations and knowledge across tasks, MTL enhances generalization and mitigates overfitting. Furthermore, MTL promotes faster learning of related tasks and alleviates the computational requirements of deep learning, making it particularly valuable in scenarios with limited task-specific data. That is why MTL has gained significant attention in various domains, including computer vision \cite{vision_mtl_vision_recognition,cross_stitch_mtl,which_tasks_to_learn}, natural language processing \cite{mtl_nlp_survey, mtl_nlp_medical,mtl_nlp_sentiment,instance_based_mtl_hiv}, speech recognition \cite{mtl_speech_nlp,mtl_speech_emotion}, and healthcare \cite{mtl_nlp_medical,mtl_covid,mtl_surgery}, and has shown promising results in improving accuracy, robustness, and efficiency. However, effectively harnessing the potential of MTL poses several challenges, including the identification of optimal task groupings \cite{MTG_task_grouping_meta_learner,TAG,which_tasks_to_learn} and the management of negative interference between tasks \cite{mtl_optimization, wu2020understanding, maninis2019attentive}.

The task grouping problem in MTL is particularly challenging due to the exponential number of possible task combinations \cite{aribandi2021ext5,TAG,which_tasks_to_learn,MTG_task_grouping_meta_learner}. What makes it worse for the exhaustive search is that each trial involves a complete training and evaluation procedure, leading to computational and optimization burden. Moreover, inappropriate task groupings may result in performance degradation due to negative transfer between tasks \cite{mtl_optimization, wu2020understanding, maninis2019attentive}. Existing solutions, which address these challenges, often suffer from scalability and modularity issues, making their practical application in real-world scenarios nearly infeasible.

In this paper, we propose a novel data-driven method for task grouping in MTL for classification tasks, which overcomes the scalability and modularity limitations. Our method utilizes Data Maps \cite{data_maps}, data-driven features that capture the training dynamics of each classification task during an MTL training. By analyzing these data maps, we can identify task groupings, both hard and soft ones, that promote positive transfer and mitigate negative interference as much as possible. We demonstrate the effectiveness of our method through extensive experimentation, including experiments on an unprecedented number of tasks, scaling up to 100 tasks to emphasize the practicality of our approach, where the generated groupings aligned with the dataset categorization. We also illustrate the performance improvement of the method in comparison to using both MTL and STL for training. Furthermore, we emphasize the practicality of our approach by providing a modular code implementation, making it easier for researchers and developers to integrate and test our method on their own datasets and tasks.

The contributions of this paper are summarized below:
\begin{itemize}
    \item We propose a novel data-driven method for classification task grouping in MTL, addressing the challenges of scalability and modularity, by re-proposing the usage of data maps as task features.
    \item We utilize soft-clustering weights to enable model specialization via loss weighting.
    \item We conduct extensive experiments, demonstrating the effectiveness of our method, even on a large number of tasks (scaling up to 100 classification tasks).
    \item We provide a modular code implementation of our method, facilitating its adoption and usage by both the research and the industry communities.
\end{itemize}

\begin{figure}[t]
    \begin{center}
    \centerline{\includegraphics[width=\textwidth]{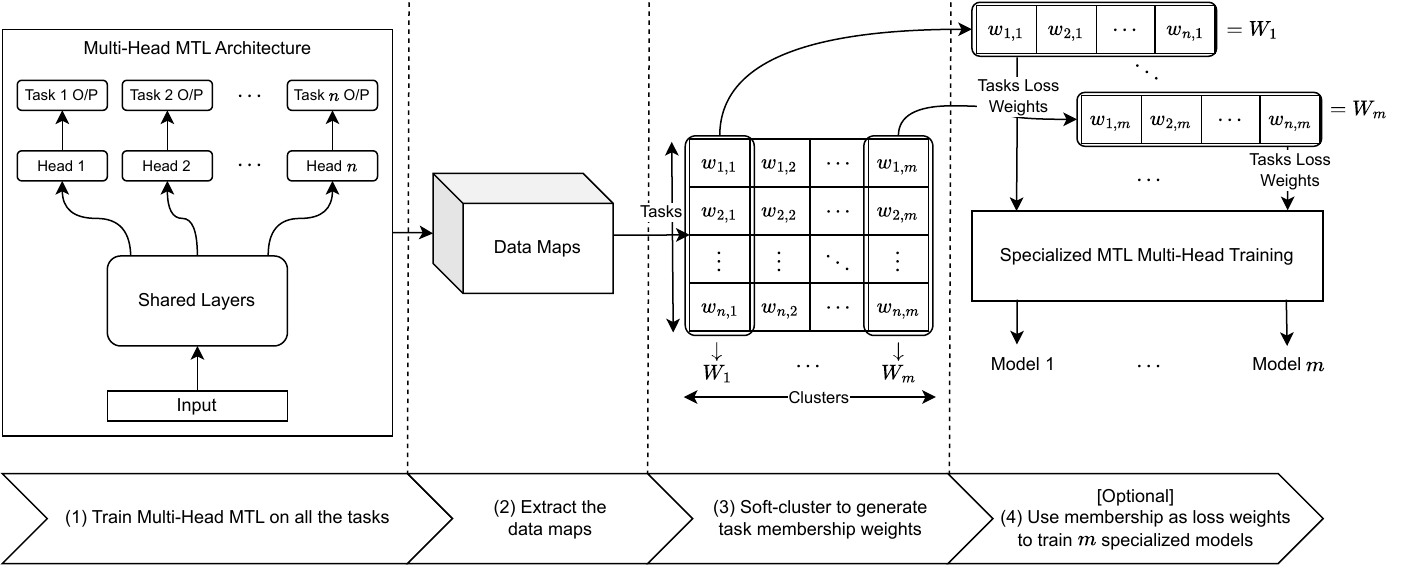}}
    \caption{\textbf{Overview} of our method to cluster the tasks using Data Maps. $(1)$ we use a single Multi-head Multi-Task Learning architecture to jointly train all the tasks. Each head is task-specific layers. $(2)$ we extract the data maps of all the tasks across the epochs in $E$. $(3)$ we use the data maps to cluster the tasks using kmeans and generate the memberships according to Equation \ref{eq_fuzzification}. $(4)$ to evaluate our clustering results, we train $m$ models where each model represents a cluster focusing on particular tasks using the memberships as loss weights.}
    \label{fig_overview_train}
    \end{center}
\end{figure}

\section{Related Work}

MTL has been extensively studied to leverage the benefits of information sharing among related tasks, which can serve as an inductive bias to improve modeling performance \cite{mtl_97_inductive_bias, MTL_Survey}. Another perspective on MTL is that it enables more efficient utilization of the model capacity by focusing on learning relevant features and reducing the impact of irrelevant signals, which contributes to overfitting, leading to better generalization. However, when tasks lack shared information, they compete for the limited model capacity, resulting in performance degradation \cite{mtl_optimization, wu2020understanding, maninis2019attentive}. To address this challenge, task grouping has emerged as a promising solution to identify subsets of tasks that can be trained together. This helps with avoiding negative interference and promoting improved performance.

Traditionally, the decision of task grouping has been approached through costly cross-validation techniques or human expert knowledge \cite{MTL_Survey}. However, these methods have limitations when applied to different problem domains and do not scale well. Therefore, some simply used correlation, either between model task predictions \cite{correlation_method_1}, data labels, or task weights \cite{correlation_method_2}, to recognize related tasks. For instance, in medical imaging, these methods have been applied to improve predictions for Cardiac Indices \cite{correlation_method_1,correlation_method_2}. However, these approaches are ineffective in scenarios lacking explicit correlations between task labels, in the dataset itself, such as exclusive membership binary classification tasks. Furthermore, both works aim to know task relationships to improve the model predictions of Cardiac Indices in multi-view or multi-modal settings, using a single model. This is quite different from our objective of finding the relations to split the tasks into different models to avoid negative interference.

Other attempts have been made to approach the problem differently enabling the models to automate the search over which parameters to share among particular tasks \cite{mtl_tree,cross_stitch_mtl}. Methods such as Neural Architecture Search \cite{liu2018progressive,huang2018gnas, chen2023mod, mtl_tree, NEURIPS2020_634841a6_AdaShare, what_layers_to_share}, Soft-Parameter Sharing \cite{ruder2019latent, long2017learning, cross_stitch_mtl}, and asymmetric information transfer \cite{lee2016asymmetric,lee2018deep,huang2022curriculum} have been developed. However, these models often exhibit poor generalization and struggle to perform well on diverse tasks and domains. Besides, they often require a large model capacity and do not thus scale well enough with a large number of tasks. 

Therefore, gradient-based approaches \cite{TAG,strezoski2019learning,task_vectors} have also been explored to determine task grouping in advance. The Task Affinity Grouping (TAG) approach \cite{TAG}, which leverages gradients to determine task similarity, is an example of such an approach. Nevertheless, it has complex training paradigm and requires $\Theta(n^2)$ more forward and backward passes to compute the inter task affinities, putting an issue with scalability even if we enhance the solution's modularity. Another method, called Higher-Order Approximation (HOA) \cite{which_tasks_to_learn}, reduces the exponential number of MTL training, from the exhaustive search, by considering only the quadratic pairs of task combinations. However, even with such relaxation, the scalability of HOA remains limited, particularly when dealing with a large number of tasks.

Recent studies have embraced the utilization of accumulated gradients to depict individual tasks as vectors \cite{task_vectors}, offering improved scalability by necessitating fine-tuning of only $\Theta(n)$ models. However, this method encounters significant limitations concerning modularity. The requirement to fine-tune a pre-trained model for each task results in a dependency where all tasks must share the same pre-trained model, which could pose challenges when dealing with a large number of tasks. It has also a constraint that all tasks must share the same architecture because task vectors are expressed in the model's weight space. Furthermore, the work focuses on using task vectors to enhance fine-tuning and transfer learning, rather than grouping the tasks into different MTL models to mitigate negative interference. 

The task grouping problem has as well been addressed differently through a Meta-Learning approach \cite{MTG_task_grouping_meta_learner}, aiming to create a meta-learner that can estimate task grouping gains. Nevertheless, the computational demands of this approach pose practical challenges for real-world applications; it requires training MTL networks for every chosen task combination in the training set for multiple iterations. It furthermore outputs all the possible gains of every task combination, whose numbers grow exponentially, and runs a search algorithm over these exponentially growing gains to find the optimal grouping. As a result, the scalability of this solution is severely limited, making it less feasible for a larger number of tasks.

\section{Task Clustering using Data Maps}
Now, we elaborate in the components in our method in the next sections. We start with stating the notations we will use along with our MTL architecture we are using in our experiments in Section \ref{section_preliminaries}. Then, we move on to illustrate the data maps, which is crucial component of our method in Section \ref{section_data_maps}. In Section \ref{section_task_clustering}, we talk regarding the approaches we use to cluster the tasks. We also introduce our evaluation mechanism of our task grouping in Section \ref{section_model_specialization}. Finally, we conclude this part with a simple theoretical comparison of our method and the literature from the perspective of scalability and modularity in Section \ref{section_theoretical_comparison}. Figure \ref{fig_overview_train} provides an overview of our method.

\subsection{Preliminaries} \label{section_preliminaries}

\textbf{Notations}\ In our paper, we use the following notations. The set of all tasks is denoted as $T=\{T_{1},\dots,T_{n}\}$, where $n$ represents the number of tasks and $|T|=n$. The total number of training data points is denoted as $N$. We calculate the data maps at specific epochs, and the set of epochs is represented as $E=\{E_1, \dots, E_e\}$, where $E_i$ corresponds to the $i^{th}$ epoch and $|E|=e$. The task clusters are denoted by $C=\{C_1, \dots, C_m\}$, and each cluster $C_i$ has an associated centroid $c_i$ and $|C|=m$. The association weight, soft-cluster weight, of each task $i$ to a cluster $j$ is represented by $w_{i,j}$, with the constraint that $\sum_{j=1}^{m}w_{i,j} = 1$; $W_{j}, WL_j$ are the weight vector and weighted loss of all the tasks in cluster $j$ respectively, while $L_j$ simply represents the resulting loss of model $j$ denoting cluster $j$. Notice that the values of $w_{i,j}$ range from $0$ to $1$, where $1$ signifies full membership and $0$ indicates no membership.

\textbf{MTL Architecture}\ Following the previous approaches \cite{TAG,which_tasks_to_learn,MTG_task_grouping_meta_learner}, we utilize a commonly employed hard-sharing multi-head architecture (Figure \ref{fig_overview_train}) for all our MTL experiments, where a single feature extractor is used to obtain shared representations, and separate task-specific heads are employed to output the result. Additionally, for all the experiments, we maintain the same data splits, via prior seeding, and keep the optimization algorithm and other hyperparameters fixed; this is to make sure any variability in the performance is only attributed to the task grouping and the corresponding weights if any.

\subsection{Data Maps as Task Features} \label{section_data_maps}
\begin{wrapfigure}{r}{0.5\textwidth}
\includegraphics[width=0.9\linewidth]{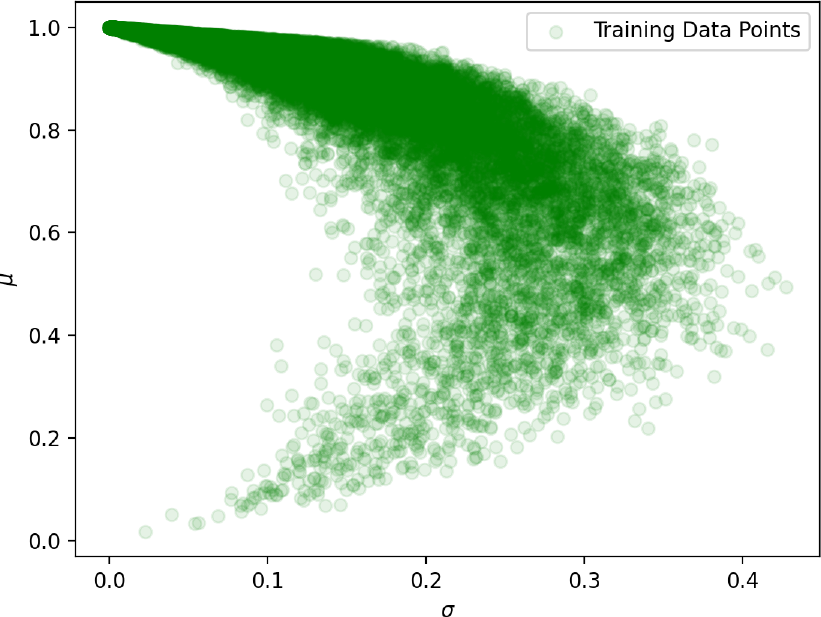} 
\caption{An example of a generated data map for the ``Living being'' task after 21 epochs of co-training on 15 tasks of G2 (Section \ref{section_datasets_tasks})}
\label{fig_data_map_exampple}
\end{wrapfigure}
Data Maps \cite{data_maps} has been originally developed as a model-based tool for characterizing and diagnosing NLP datasets. We re-propose using them as task features during training in an MTL settings given their ability to capture the model behavior of each individual training data point with respect to each task, which is a valuable characterization to the task itself. For that reason, we employ them, in our work, as task features due to their simplicity, scalability, and ability to extract them on the fly without prior knowledge of the model architecture, thus enhancing the modularity of our approach.

The concept behind Data Maps revolves briefly around extracting two values for each data point: the \textit{model confidence} ($\mu$) of the true class, which is the average probability of the true class over the epochs, and the \textit{variability} ($\sigma$) of this confidence, which is the standard deviation of the true class probabilities over the same epochs. For a particular task, the data map shape is $(N,2)$. Figure \ref{fig_data_map_exampple} shows a visualization of the resulting Data Map for an example task extracted from CIFAR10 dataset \cite{krizhevsky2009learning_cifar}.

Because their information is very task-dependent, we thought they can serve as task descriptors. To further enhance the expressiveness of the extracted features, we also extract data maps at various epochs, allowing us to gain insights into their evolution over time; the resulting shape in such case is $(n, e, N, 2)$. Therefore, by analyzing their characteristics, over the different epochs during training, we can capture crucial information about the relatedness of each task. 

In the extraction of data maps, we could have used two approaches. The first approach involves building a single MTL model that incorporates all tasks and extracting the data maps directly from this unified model. Alternatively, we can utilize the second approach, where individual models are constructed for each task, resulting in multiple STLs, and merging the data maps obtained from each model. Our results are primarily based on the first approach, as it offers the advantage of using a single training model, simplifying computational complexity, enhancing scalability, and streamlining experimentation, while having a comparable results to STL, as shown in Section \ref{section_evaluation_results}.

\subsection{Task Clustering} \label{section_task_clustering}

With the extracted data maps in hand, our next step is to group the tasks into clusters based on their similarity. We propose using soft clustering to obtain class membership weights, $w_{i,j}$. To obtain them, we represent each task as a vector by flattening the concatenated data maps, extracted at different epochs. We then employ the k-means algorithm \cite{lloyd1982least} to cluster these task vectors, aiming to identify distinct clusters of tasks, into hard clusters. To introduce a more nuanced representation of task similarities, we incorporate the fuzzification step \cite{bezdek1984fcm} into our approach, yet we modify it for better computational efficiency, equation \ref{eq_fuzzification}. This enables soft clustering, where $x_i$ represents the $i^{th}$ task vector of the corresponding data maps and $F>1$ represents the fuzzification index. This fuzzification process assigns soft memberships to tasks, allowing for more flexible and comprehensive clustering results. We predominantly rely on the soft clustering approach due to its effectiveness and reliability in addition to allowing model specialization via loss weighting, as explained in Section \ref{section_model_specialization}.

\begin{align}
    w_{i,j} &= \frac{1}{\sum_{k=1}^{n} \left(\frac{\lVert x_i - c_j \rVert}{\lVert x_i - c_k \rVert}\right)^{\frac{2}{F-1}}}\\
    &= \frac{\lVert x_i - c_j \rVert^{\frac{-2}{F-1}}}{ \sum_{k=1}^{n} \left(\lVert x_i - c_k \rVert\right)^{\frac{-2}{F-1}}} \label{eq_fuzzification}
\end{align}

\subsection{Model Specialization through Loss Weighting} \label{section_model_specialization}
\begin{figure}[ht]
    \centering
    \centerline{\includegraphics[width=0.6\columnwidth]{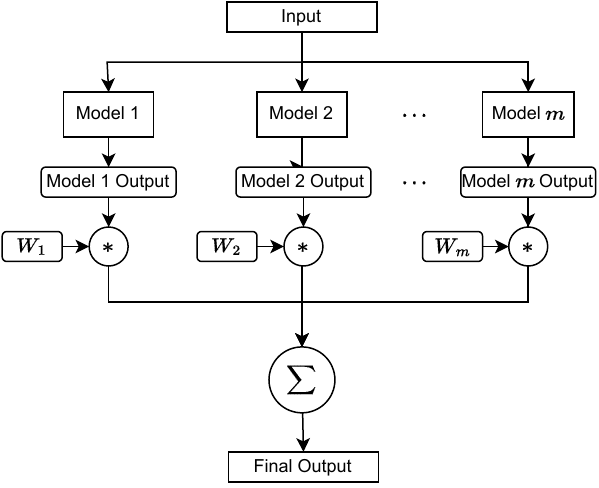}}
    \caption{The procedure to use our specialized trained models to infer the results}
    \label{fig_infer}
\end{figure}

In order to assess the effectiveness of our task grouping results, we use loss weighting as a method of model specialization. We construct MTL models that are tailored to specialize in specific sets of tasks based on the membership weights obtained from soft clustering results. For each cluster, we build an individual MTL model that focuses on the tasks assigned to that cluster according to their corresponding weights (Equation \ref{eq_weighted_loss}). To evaluate the performance of our solution, we apply the weighted average of the models' outputs according to the membership weights as in Equation \ref{eq_weighted_average}, where $O_j$ is the output of the $j^{th}$ model and $\circ$ represents Hadamard, element-wise, product. Figure \ref{fig_infer} provides an overview of these operations while inferring the output. By comparing the resulting values with both STL and traditional MTL schemes in Section \ref{section_evaluation_results}, we can gain insights into the benefits and improvements brought by our task grouping approach.
\begin{align}
    WL_j &= W_j^{T} L_j\label{eq_weighted_loss}\\
    \text{Final Output} &= \sum_{j=1}^{m} W_j \circ  O_j\label{eq_weighted_average}
\end{align}

\subsection{Theoretical Scalability Comparison}\label{section_theoretical_comparison}

In the theoretical scalability comparison, we evaluate our method against existing literature, focusing on the required number of trained models to get the clustering results. Table \ref{table_asymptotic} presents the comparison, where lower numbers indicate better scalability. Our approach stands out with excellent scalability, as it only necessitates training a single MTL model to extract data maps and perform clustering, or even $\mathcal{O}(n)$ if we consider extracting data maps from $n$ STL models. This offers the most promising scalability potential for a larger number of tasks. That is why we can scale our experiments to a very large number of tasks as in Section \ref{experiments}. We also highlight the note of Table \ref{table_asymptotic} that although TAG requires a single model to extract the groupings, we compute the gradients $\Theta(n^2)$, $\binom{n}{2}$ in particular, times than usual to compute the inter-task affinities pairs. Therefore, one model training of TAG utilizes the same asymptotic time of $\Theta(n^2)$ normally trained models. 

Furthermore, our method's data map computation is performed on the fly, making it both model and task agnostic. This feature enhances the modularity of our approach, enabling effortless adaptation to different model architectures and tasks without manual intervention. Refer to Appendix \ref{appendix_code_structure} for further discussion on the modularity aspect of our method.

\begin{table}[t]
\centering
\caption{Comparison of asymptotic growth of the required number of trained MTL models to get task grouping of various methods and ours}
\label{table_asymptotic}
\begin{tabular}{lcccr}
\toprule
Method & Number of Models $(\downarrow)$\\
\midrule
Exhaustive Search    & $\Theta\left(2^{n}\right)$ \\
HOA    & $\Theta\left(\binom{n}{2}\right)=\Theta\left(n^2\right)$ \\
TAG$^{*}$    & $\Theta\left(n^2\right)$ \\
MTG-Net    & $\Theta\left(n \cdot K\right)$ \\
\midrule
STG-MTL    & $\Theta\left(1\right)$ \\
\bottomrule
\end{tabular}\\
\begin{minipage}[b]{0.45\columnwidth}
    \small $*\quad$ TAG originally requires one model, yet we do $\binom{n}{2}$ updates to compute the inter-task affinities. For that reason a single training model of TAG is asymptotically equivalent to training $\Theta(n^2)$ models from other approaches. 
\end{minipage}
\end{table}

\section{Experiments} \label{experiments}
In this section, we present a comprehensive overview of our experiments, focusing on assessing the effectiveness of our method and presenting the corresponding results. As outlined in Section \ref{section_theoretical_comparison}, the methods in literature has severe scalability issues. This makes a direct comparison of performance across all methods unfeasible, considering their poor scalability, the substantial number of tasks involved in our evaluation, and our limited computational power. To illustrate, the clustering results of just $15$ tasks for HOA necessitate the training of $\binom{15}{2} = 105$ models. Therefore, we evaluate our method against both the MTL and STL results to showcase the added information gained from our clustering results. 

Section \ref{section_datasets_tasks} outlines the specifics of the datasets utilized in our experimentation, as well as the tasks employed. In Section \ref{section_model_architecture_hyper_parameters}, we delve into the details of the model architecture and the hyperparameters used during experimentation. The outcomes of the soft clustering of tasks are presented in Section \ref{section_clustering_results}, where we highlight the effectiveness of our approach in grouping related tasks. Finally, in Section \ref{section_evaluation_results}, we evaluate the quality of the obtained clustering results comparing them to STL and MTL results. Appendix \ref{appendix_code_structure} illustrates our modular pipeline structure for all the experiments for easier reproducibility.

\subsection{Datasets and Tasks}\label{section_datasets_tasks}
Our task generation is based on the CIFAR10 and CIFAR100 datasets \cite{krizhevsky2009learning_cifar} along with CelebA dataset \cite{liu2015faceattributes}. We mainly use them due to their public availability and the available high number of tasks we could generate from them. We show the results of four groups of tasks in our experiments. For all of these datasets, we define binary tasks to detect whether the input image belongs to a particular label or not. In Group 1 (\textbf{G1}), for example, we include binary classification tasks that determine whether an image belongs to CIFAR10 labels. G1 consists of 10 tasks: $\{$airplane, automobile, bird, cat, deer, dog, frog, horse, ship, truck$\}$. That is the ``airplane'' task is to detect whether the image is an airplane or not.

Group 2 (\textbf{G2}) expands on G1 by introducing additional tasks on CIFAR10, designed for testing qualitatively the method behavior. These tasks include $\{$Living being, Odd-numbered, Downside, Not living being, random$\}$. The ``Living being'' task aims to detect whether an image contains a living being, which includes images with the labels $\{$bird, cat, deer, dog, frog, horse$\}$. Similarly, the ``Not living being'' task focuses on identifying non-living beings; these are $\{$ airplane, automobile, ship, truck$\}$ classes in CIFAR10. Notably, ``Living being'' and ``Not living being'' are intentionally designed to be similar tasks to check if the method can detect such similarity. The ``Odd-numbered'' task identifies whether the label of a CIFAR10 image is odd or not, encompassing $\{$automobile, cat, dog, horse, truck$\}$ classes. Additionally, we flip half of CIFAR10 images and create a task to train the model to recognize vertically flipped images, the ``Downside'' task. Lastly, the ``random'' task assigns random binary labels to the entire dataset with a predefined seed for consistency and reproducibility; the ``random'' task is added to represent extremely difficult task, for example. It is worth mentioning that while the original tasks in G1 are imbalanced, the extra tasks in G2 are balanced.

Group 3 (\textbf{G3}), similar to G1, consists of $100$ binary classification tasks using the CIFAR100 labels. We also utilize the $20$ \textbf{super labels} of CIFAR100 as our ground truth for task clustering evaluation. It is worth mentioning that CIFAR100 super labels are not intended for task grouping, so they are not grouped based on visual similarities like our method's objective. Instead, they are mostly clustered semantically, even though there are some exceptions like mushrooms and the classes of vehicles 1 and 2. Still, we think they serve as an informative indicator of the effectiveness of our approach, especially in the visually coherent superclasses. Similarly, we define group 4 (\textbf{G4}) to include the $40$ binary classification tasks included in CelebA.

\subsection{Model Architectures and Hyper-Parameters} \label{section_model_architecture_hyper_parameters}
For all our experiments, we adopt the RESNET18 architecture \cite{resnet_he2016deep} as our base model. Our method is model-agnostic, so we have also experimented with simpler models, results \& archtecture discussed in Appendix \ref{appendix_custom_cnn_exp}, yet we use RESNET18 considering its moderate model capacity. Furthermore, we utilize it without any pre-training, ensuring that the model starts from scratch for each task grouping scenario. The last fully connected layer of RESNET18 serves as the task heads, with the number of output neurons corresponding to the number of tasks. Each neuron in the task heads represents a specific classification task. Throughout our experiments, the rest of the network, excluding the task heads, is shared among all tasks. Also, we train the model for $50$ epochs in all our experiments: to extract data maps and to evaluate the models. Additionally, in our clustering process, we primarily set the fuzzification index ($F$) to $2$. The fuzzification index controls the level of fuzziness in the soft clustering algorithm, so increasing it produce softer decisions.

In terms of the loss function, we utilize Binary Cross Entropy as the binary classification loss for our tasks. However, to address the issue of task imbalance, we incorporate a penalty on positive instances for each task. By applying this penalty, we ensure that the model pays more attention to the minority label during training, thereby mitigating the impact of the imbalance and promoting better overall performance. Finally, it worth mentioning that we do not perform any kind of tuning to any model. We use the same basic settings in all of our experiments.

\subsection{Task Clustering Results} \label{section_clustering_results}
\begin{figure}[ht]
    \centering
    \begin{minipage}[b]{\columnwidth}
        \centerline{\includegraphics[width=0.9\columnwidth]{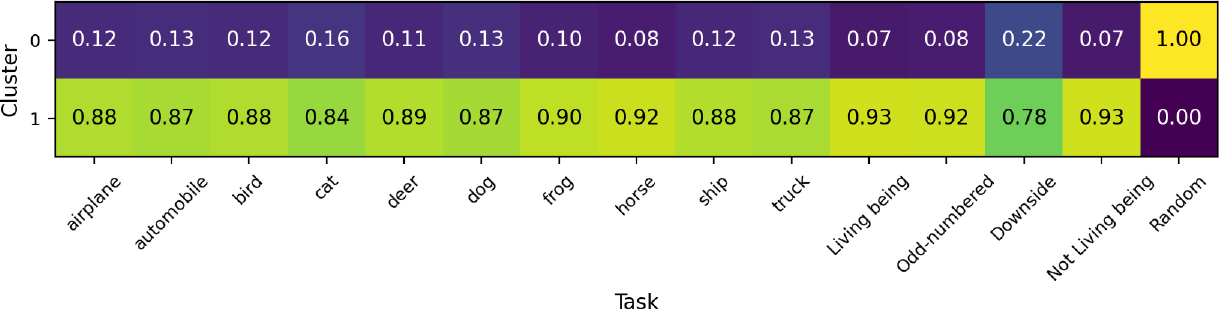}}
        \subcaption{G2: $2$ clusters}\label{fig_15T_2C}
    \end{minipage}%
    \\
    \begin{minipage}[b]{\columnwidth}
        \centerline{\includegraphics[width=0.9\columnwidth]{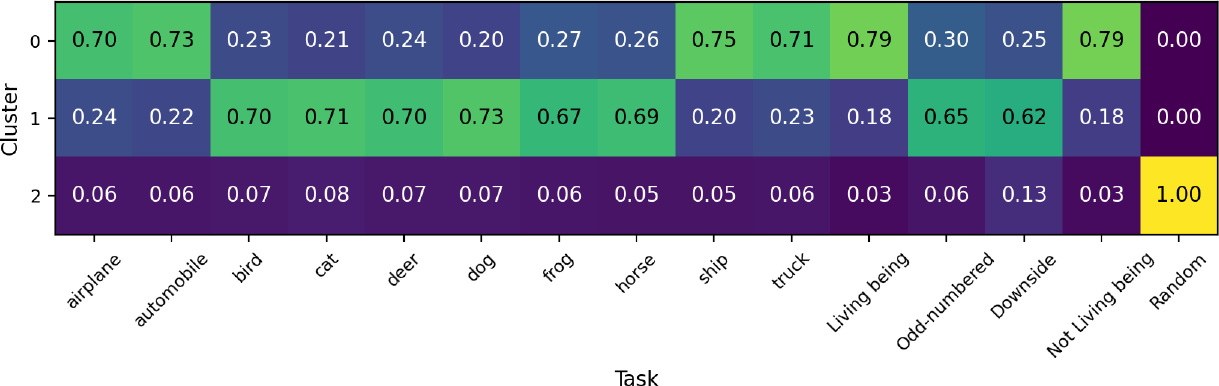}}
        \subcaption{G2: $3$ clusters}\label{fig_15T_3C}
    \end{minipage}
    \\
    \begin{minipage}[b]{0.48\columnwidth}
        \centerline{\includegraphics[width=\columnwidth]{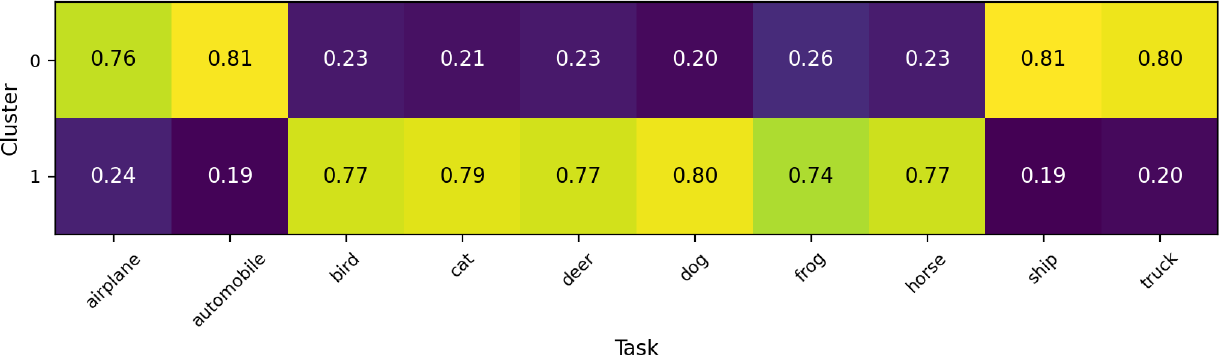}}
        \subcaption{G1: $2$ clusters}\label{fig_10T_2C}
    \end{minipage}%
    \hfill
    \begin{minipage}[b]{0.48\columnwidth}
        \centerline{\includegraphics[width=\columnwidth]{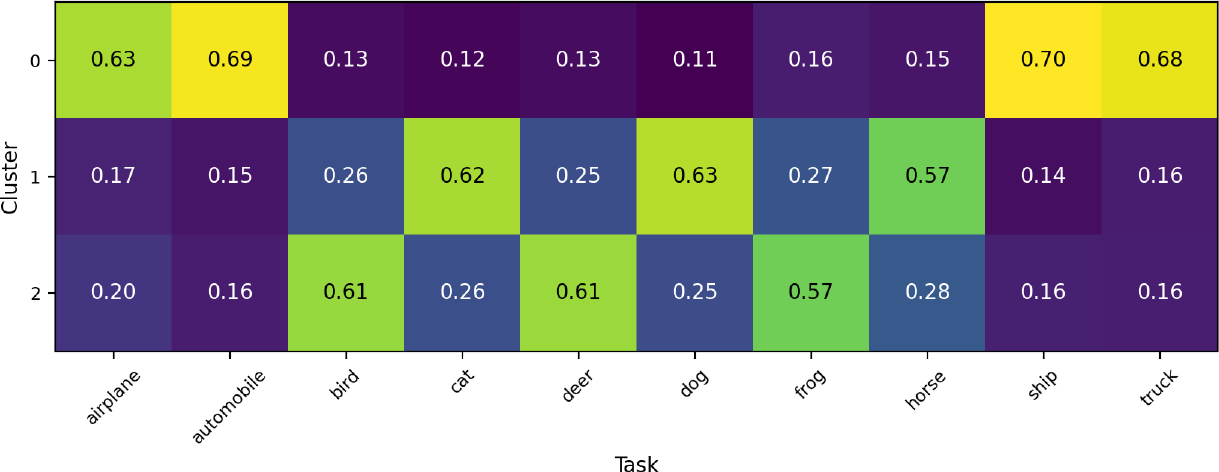}}
        \subcaption{G1: $3$ clusters}\label{fig_10T_3C}
    \end{minipage}
    \caption{Task grouping of G1 \& G2 with $F=2$}\label{fig_cifar10_tasks}
\end{figure}

Results of our task clustering experiments are presented for all groups. We initially experimented on G2, generating their data maps as described in Section \ref{section_data_maps} and Clustering them as in Section \ref{section_task_clustering}, as depicted in Figure \ref{fig_cifar10_tasks}. Notably, our method successfully clustered the ``random'' task separately, indicating its dissimilarity to the other tasks. Furthermore, throughout all our experiments, the tasks ``Living being'' and ``Not living being'' consistently exhibited the \textbf{same membership distribution}, which is reasonable considering their equivalence.

Moreover, when focusing solely on the first 10 tasks from G2 without any additional tasks, our clustering algorithms demonstrated some semantic clustering capabilities, as shown in Figure \ref{fig_15T_3C}. The algorithm successfully grouped images of living beings, including $\{$bird, cat, deer, dog, frog, horse$\}$, while another group consisted of images of non-living beings such as $\{$airplane, automobile, ship, truck$\}$. Nevertheless, this might be due to the impact of the ``Living being'' and ``Not living being'' tasks; we therefore conducted a similar experiment on G1, generating their data maps and clustering the tasks, without any extra tasks.

As illustrated in Figure \ref{fig_10T_2C}, even without additional tasks, our method performed the same reasonable clustering for G1, grouping living beings together and non-living things together. Additionally, Figure \ref{fig_10T_3C} demonstrates the clustering using three clusters, revealing that the living being cluster was divided into two groups: cluster $1$ and cluster $2$. Cluster $1$ predominantly contained quadruped animals $\{$cat, dog, horse$\}$, while cluster $2$ included $\{$bird, frog, deer$\}$ that represented the other living creatures except for the deer. These results showcase the effectiveness of our clustering algorithm in capturing semantic, visual similarities among tasks based on the visual data leading to meaningful task groupings.

\begin{figure}[h]
    \begin{minipage}[c]{\columnwidth}
        \centerline{\includegraphics[width=0.9\columnwidth]{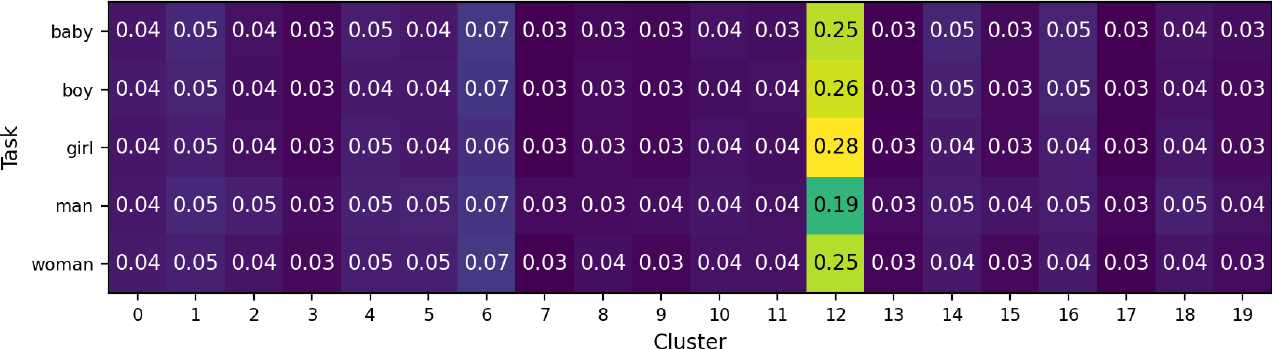}}
        \subcaption{Membership clustering of People superclass}\label{fig_100T_20C_people}
    \end{minipage}%
    \\
    \begin{minipage}[c]{\columnwidth}
        \centerline{\includegraphics[width=0.9\columnwidth]{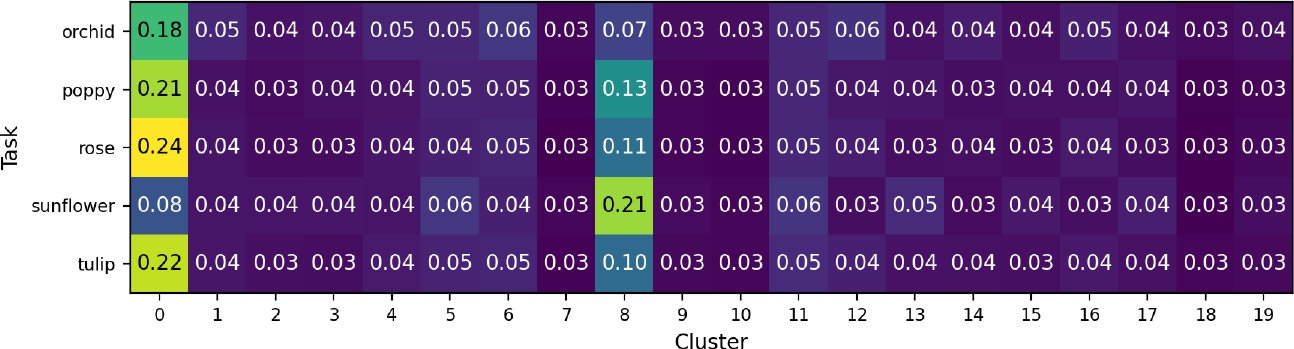}}
        \subcaption{Membership clustering of Flowers superclass}\label{fig_100T_20C_flowers}
    \end{minipage}
    \caption{Task grouping of G3 (100 Tasks of CIFAR100) into $20$ clusters with $F=2$}\label{fig_100T_20C}
\end{figure}

In addition to our experiments on G1 and G2, we conducted a comprehensive evaluation of an unprecedented number of tasks, specifically $100$ tasks from CIFAR100, in G3. As part of this evaluation, we compared our task clustering results against the predefined superclasses provided by CIFAR100. It is important to note that the superclasses in CIFAR100 primarily rely on semantic relationships as illustrated in Section \ref{section_datasets_tasks}, so we show the results of people and flowers, as examples.

In Figure \ref{fig_100T_20C}, we showcase an example of the clustering results for a group of super tasks. It is noteworthy that our method successfully clusters certain groups of tasks in alignment with the predefined CIFAR100 superclasses, as illustrated in Figure \ref{fig_100T_20C_people}. However, it is important to acknowledge that there are cases where the clustering may not be perfect, as depicted in Figure \ref{fig_100T_20C_flowers}; we think this is primarily because our method focus one visual similarities, which is exploited during training rather than semantics. Nevertheless, even in such instances, our clustering algorithm manages to allocate significant weights of all tasks into distinctive clusters, such as clusters $0$ and $8$ in Figure \ref{fig_100T_20C_flowers}. Notably, in cluster $8$, the participation percentages of the tasks $\{$orchid, poppy, rose, tulip$\}$ are the $2^{nd}$ highest across all clusters, indicating a close relationship with the missclassified task sunflower, yet our method suggests that the other four tasks are more visually related. We further discuss all the clustering results of the 100 Tasks in Appendix \ref{appendix_task_clustering} for RESNET18; we also show the clustering results of simpler CNN model in Appendix \ref{appendix_custom_cnn_exp}, which shows the possible potential of transferring the results generated from simpler models to other models. 

\subsection{Evaluation Analysis} \label{section_evaluation_results}
\begin{figure}[t]
    \centering
    \begin{minipage}[c]{0.24\columnwidth}
        \begin{minipage}[c]{\columnwidth}
            \centerline{\includegraphics[width=\columnwidth]{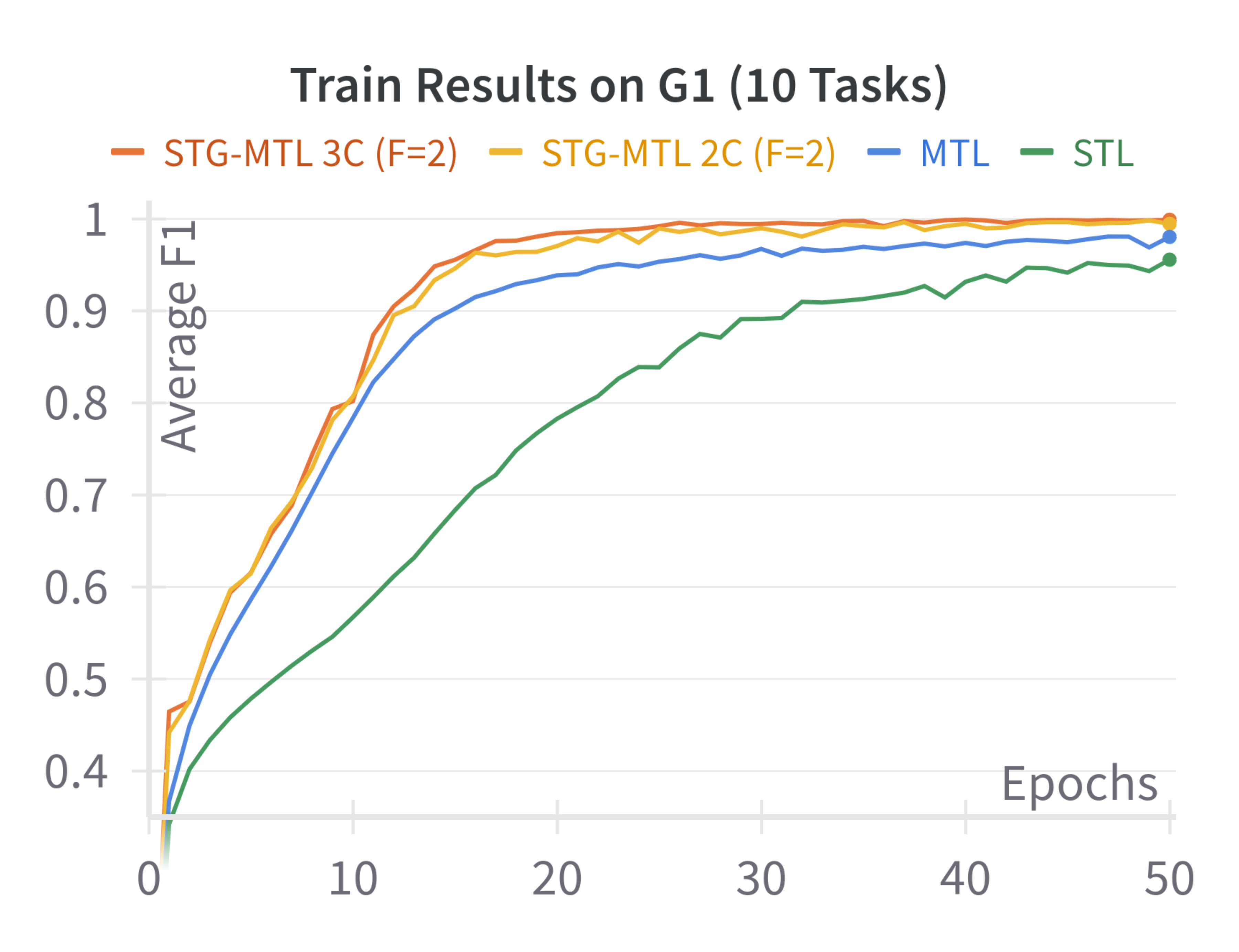}}
            \subcaption{Training Results (G1)}\label{fig_curve_G1_train}
        \end{minipage}
        \\
        \begin{minipage}[c]{\columnwidth}
            \centerline{\includegraphics[width=\columnwidth]{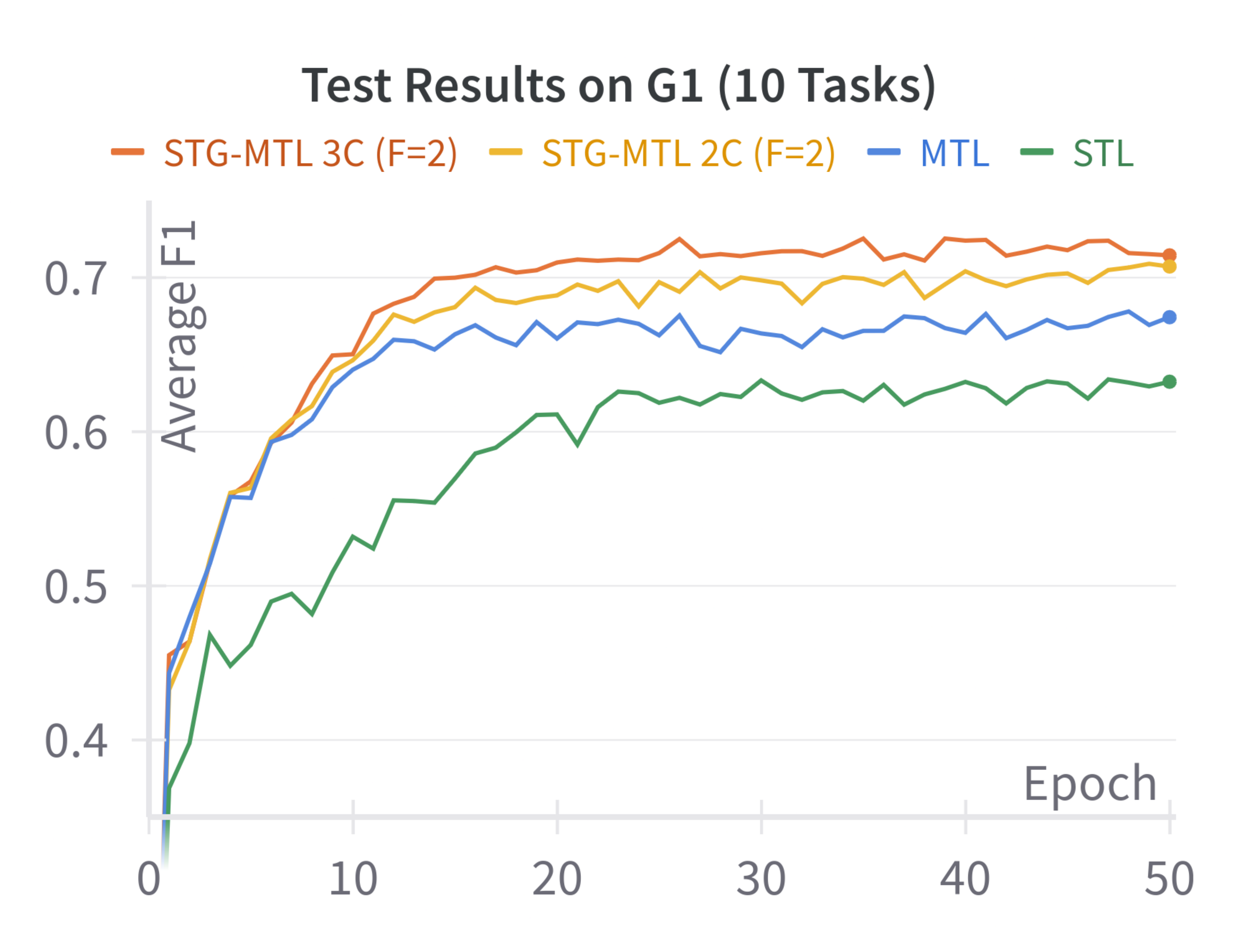}}
            \subcaption{Test Results (G1)}\label{fig_curve_G1_test}
        \end{minipage}
    \end{minipage}
    \hfill
    \begin{minipage}[c]{0.24\columnwidth}
        \begin{minipage}[c]{\columnwidth}
            \centerline{\includegraphics[width=\columnwidth]{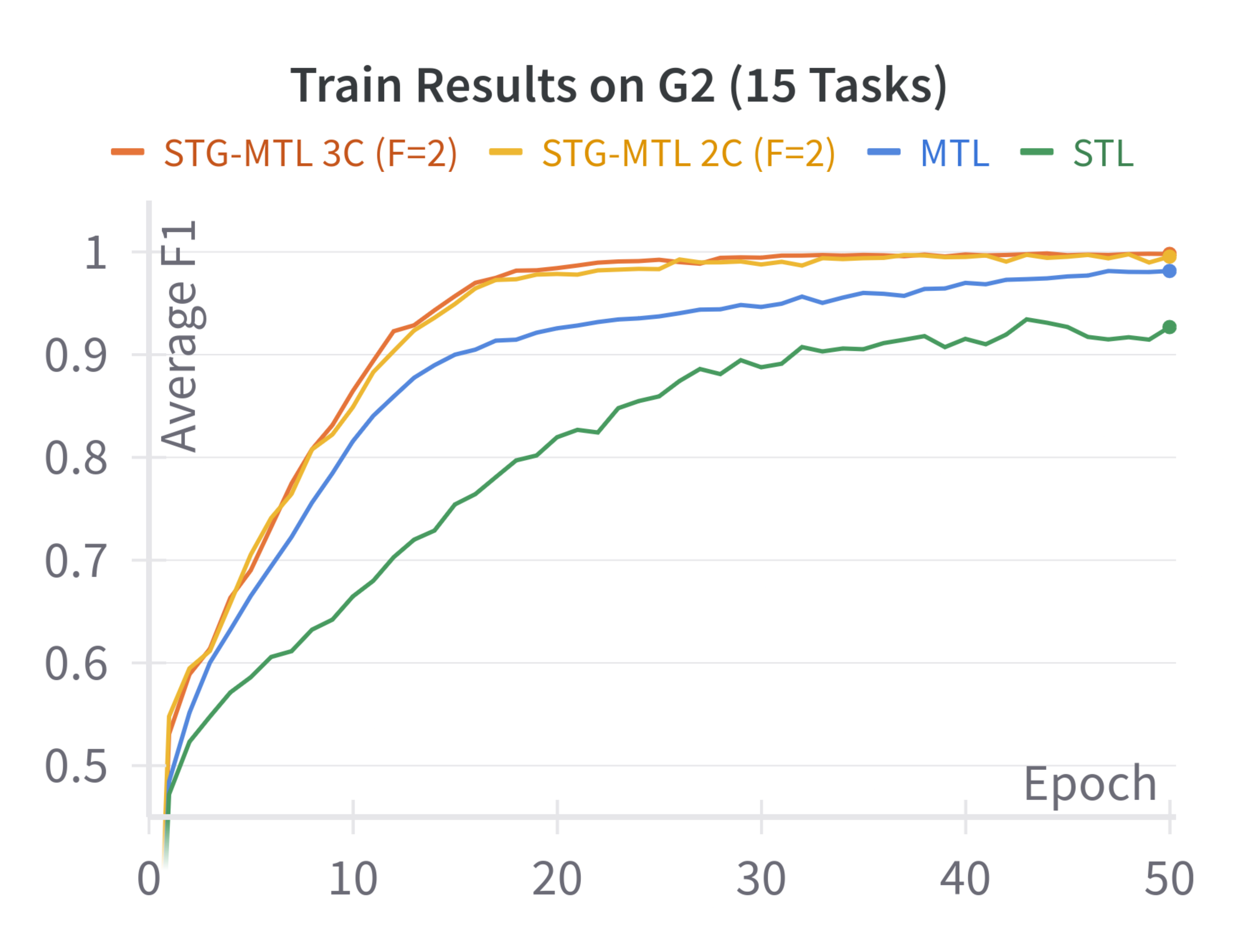}}
            \subcaption{Training Results (G2)}\label{fig_curve_G2_train}
        \end{minipage}
        \\
        \begin{minipage}[c]{\columnwidth}
            \centerline{\includegraphics[width=\columnwidth]{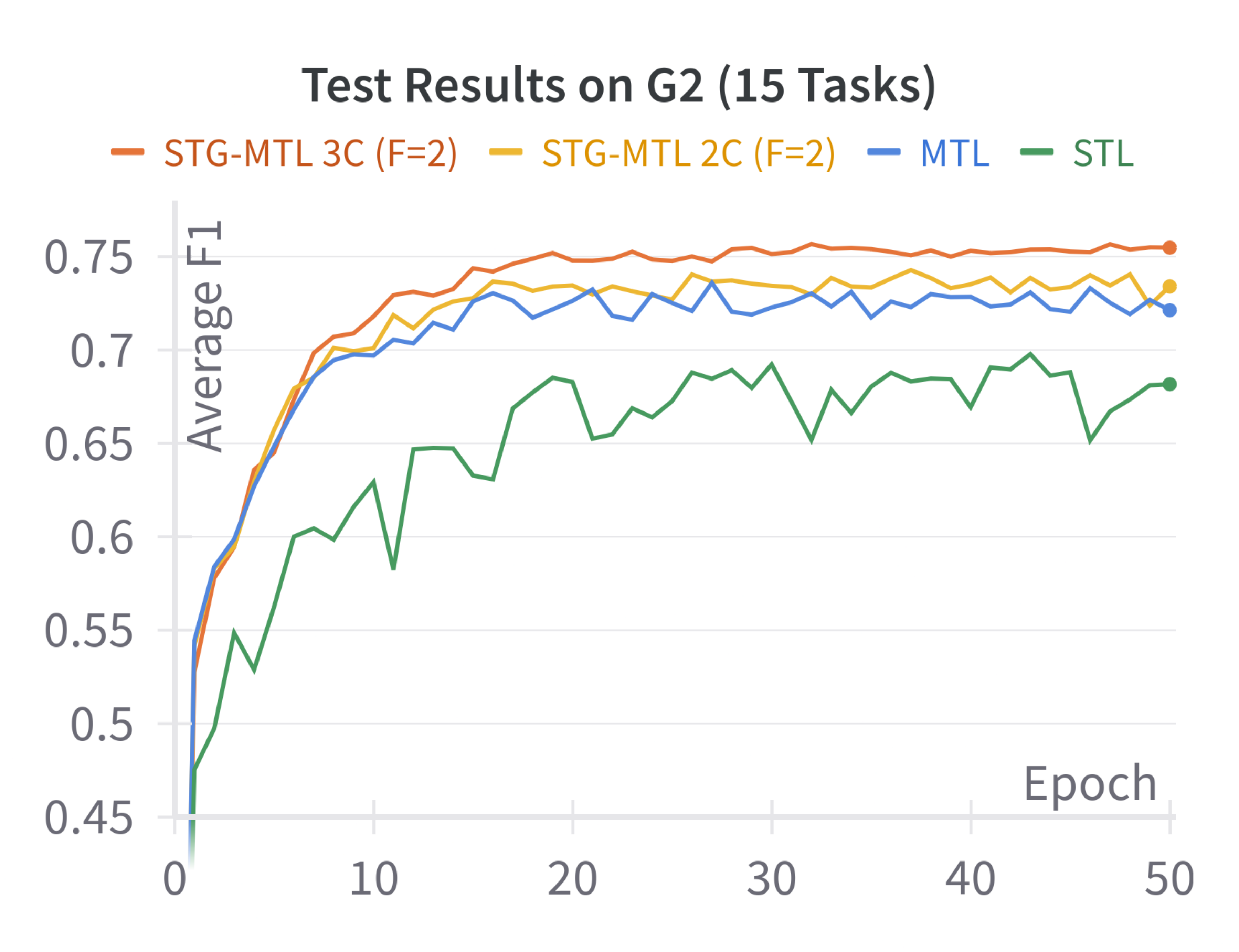}}
            \subcaption{Test Results (G2)}\label{fig_curve_G2_test}
        \end{minipage}
    \end{minipage}
    \hfill
    \begin{minipage}[c]{0.24\columnwidth}
        \begin{minipage}[c]{\columnwidth}
            \centerline{\includegraphics[width=\columnwidth]{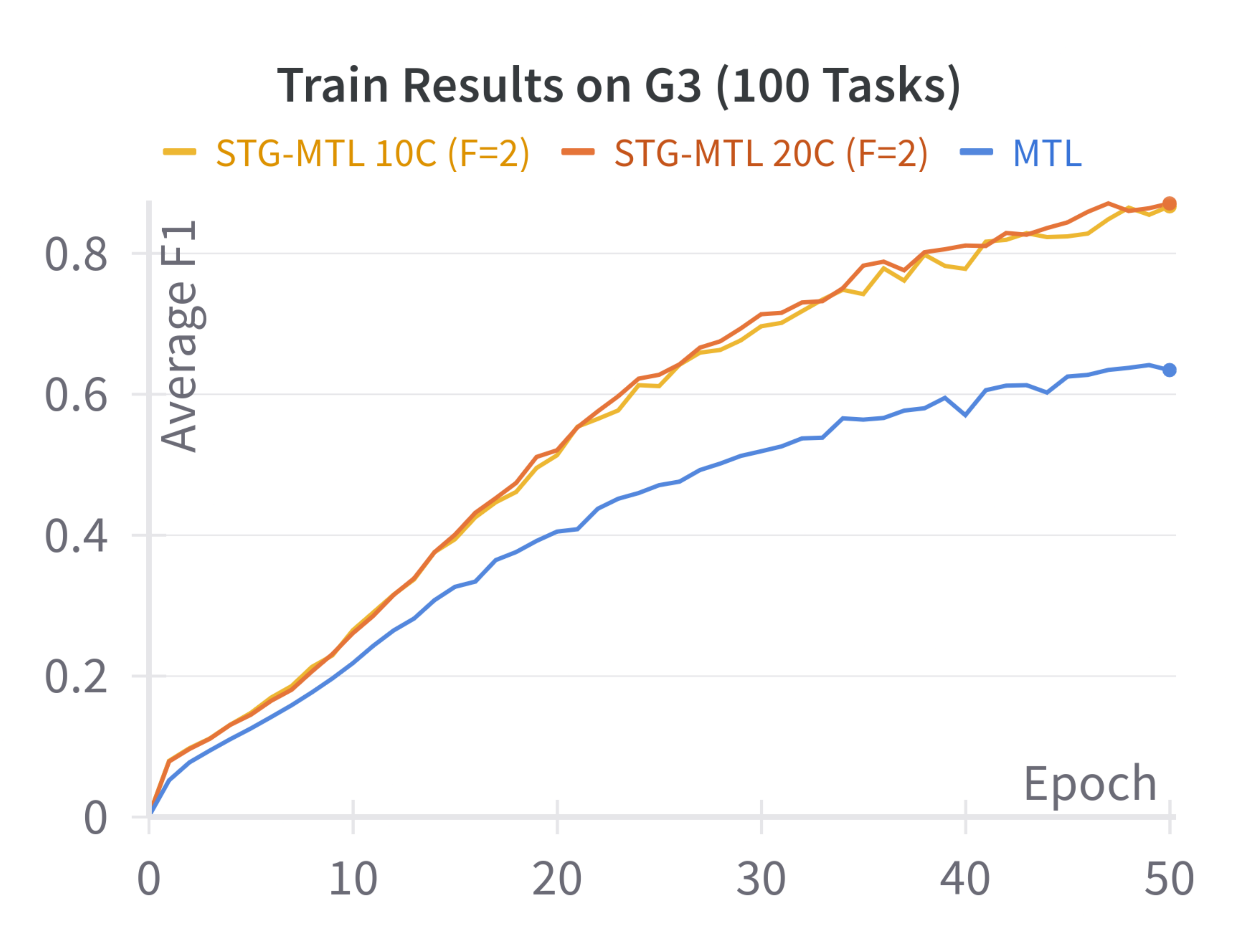}}
            \subcaption{Training Results (G3)}\label{fig_curve_G3_train}
        \end{minipage}
        \\
        \begin{minipage}[c]{\columnwidth}
            \centerline{\includegraphics[width=\columnwidth]{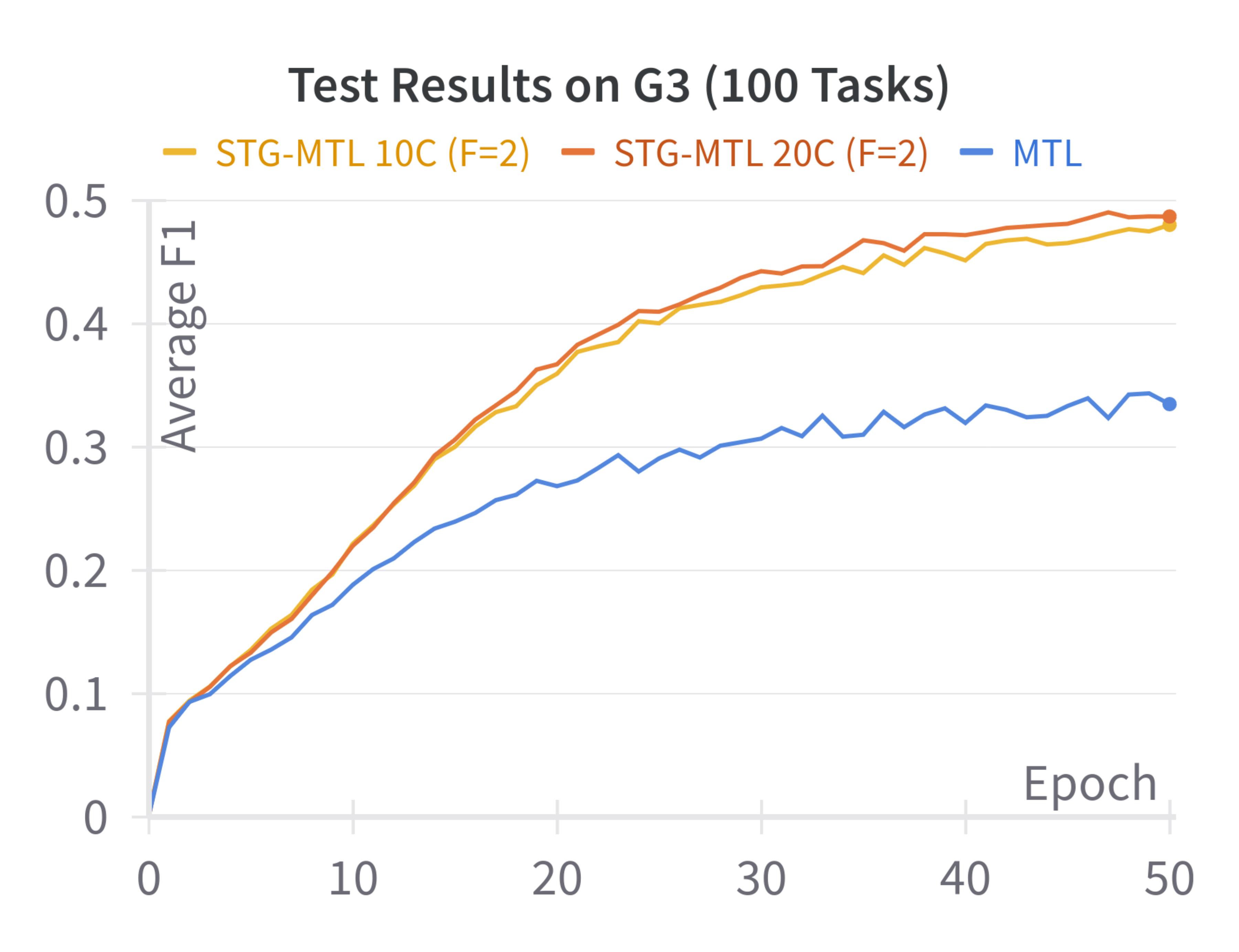}}
            \subcaption{Test Results (G3)}\label{fig_curve_G3_test}
        \end{minipage}
    \end{minipage}
    \hfill
    \begin{minipage}[c]{0.24\columnwidth}
        \begin{minipage}[c]{\columnwidth}
            \centerline{\includegraphics[width=\columnwidth]{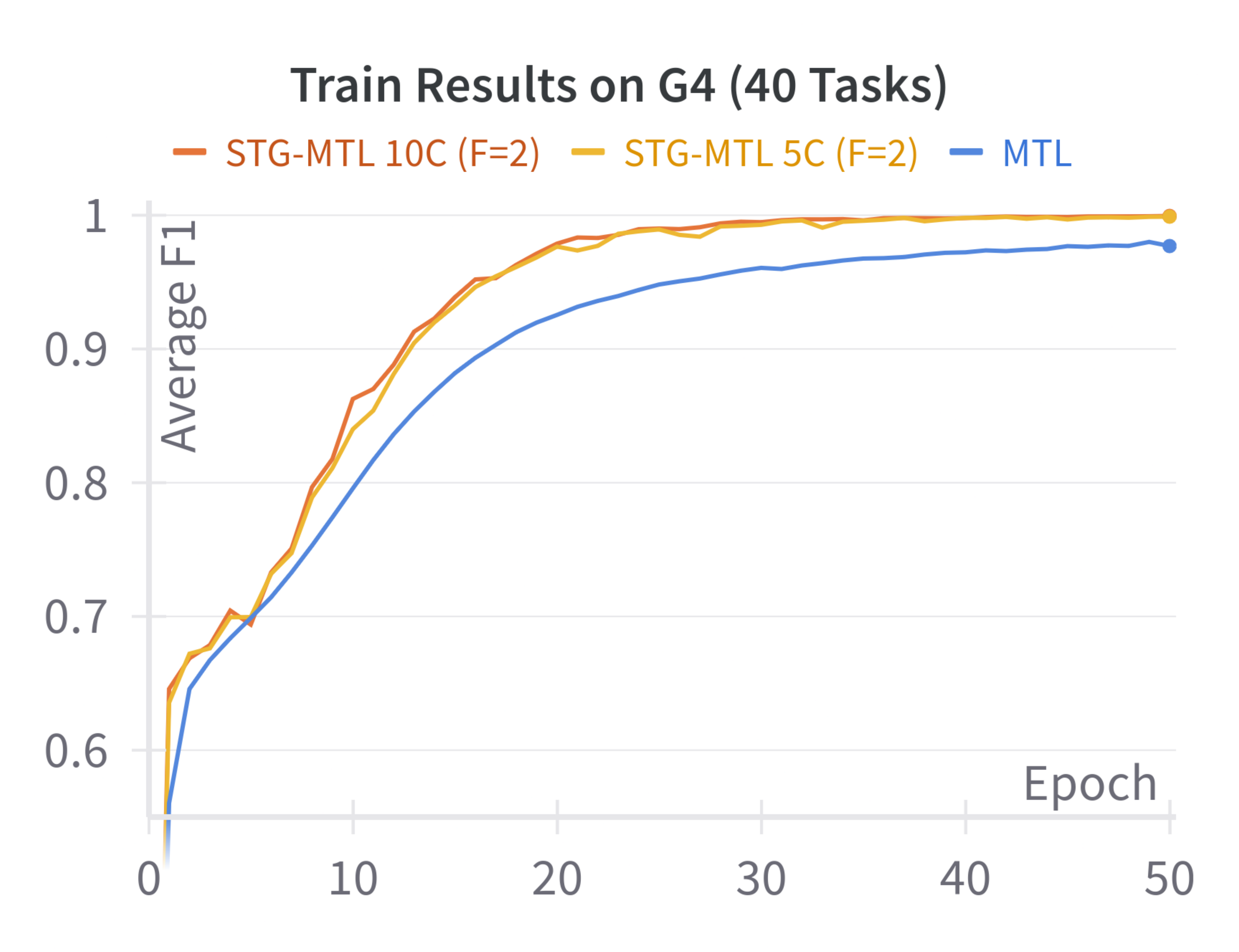}}
            \subcaption{Training Results (G4)}\label{fig_curve_G4_train}
        \end{minipage}
        \\
        \begin{minipage}[c]{\columnwidth}
            \centerline{\includegraphics[width=\columnwidth]{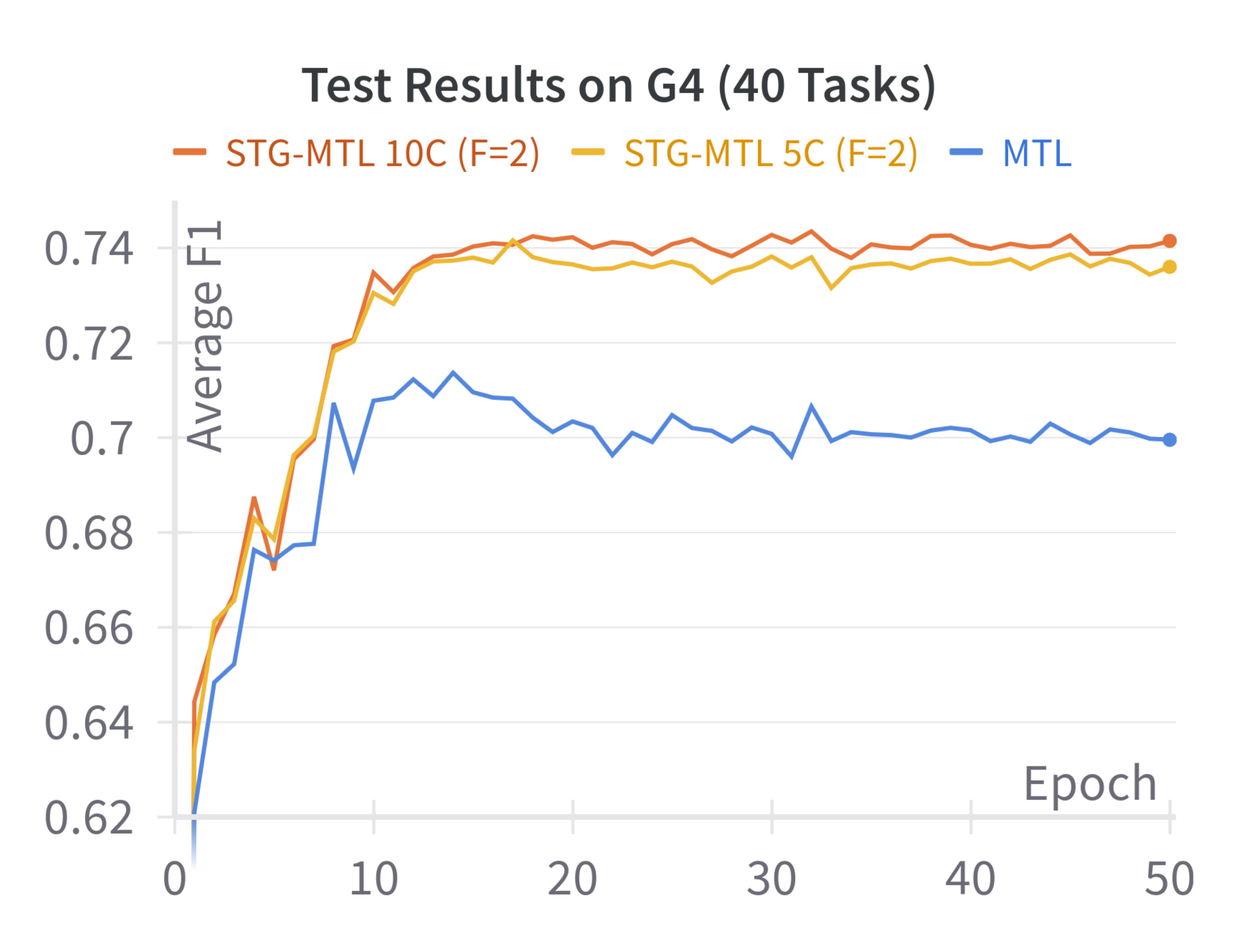}}
            \subcaption{Test Results (G4)}\label{fig_curve_G4_test}
        \end{minipage}
    \end{minipage}
    \caption{Average F1 Scores of all the Groups on both the training and test sets}\label{fig_f1_eval}
\end{figure}

To further validate the effectiveness of our method, we conducted a comprehensive evaluation as described in Section \ref{section_model_specialization} on all task groups. Figure \ref{fig_f1_eval} presents the average F1 score for both the training and test sets of all the three sets of tasks; we use F1 as our evaluation metric becuase many tasks are imbalanced, so we use F1 score to adhere to such issue. Our method is denoted by \texttt{STG-MTL xxC (F=2)} where \texttt{xx} represents the number of clusters. The MTL curve represents the results obtained from training an MTL model on all tasks without any grouping, while the STL curve represents the results obtained by training separate STL models for each task and merging their outputs. We compare the performance of our method against the MTL and STL approaches in both G1 \& G2 and against the MTL approach only in G3 \& G4 because the STL performance is poorer than the MTL, as it overfits.

\begin{figure}[t]
    \centering
    \captionsetup{justification=centering}
    \begin{minipage}[c]{0.24\columnwidth}
        \begin{minipage}[c]{\columnwidth}
            \centerline{\includegraphics[width=\columnwidth]{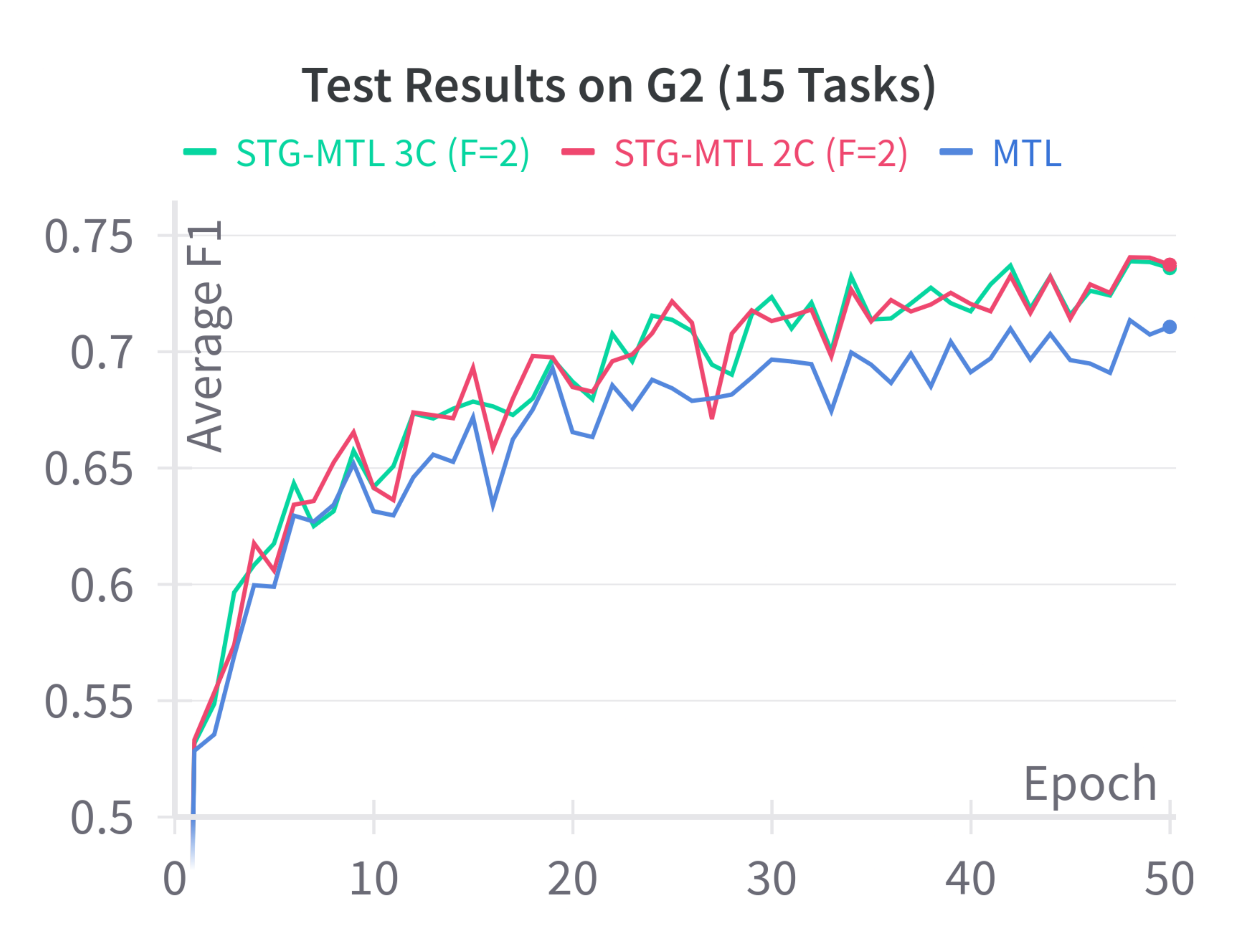}}
            \subcaption{MTL vs STG Results (G2)}\label{fig_custom_curve_mtl_stg_G2}
        \end{minipage}
        \\
        \begin{minipage}[c]{\columnwidth}
            \centerline{\includegraphics[width=\columnwidth]{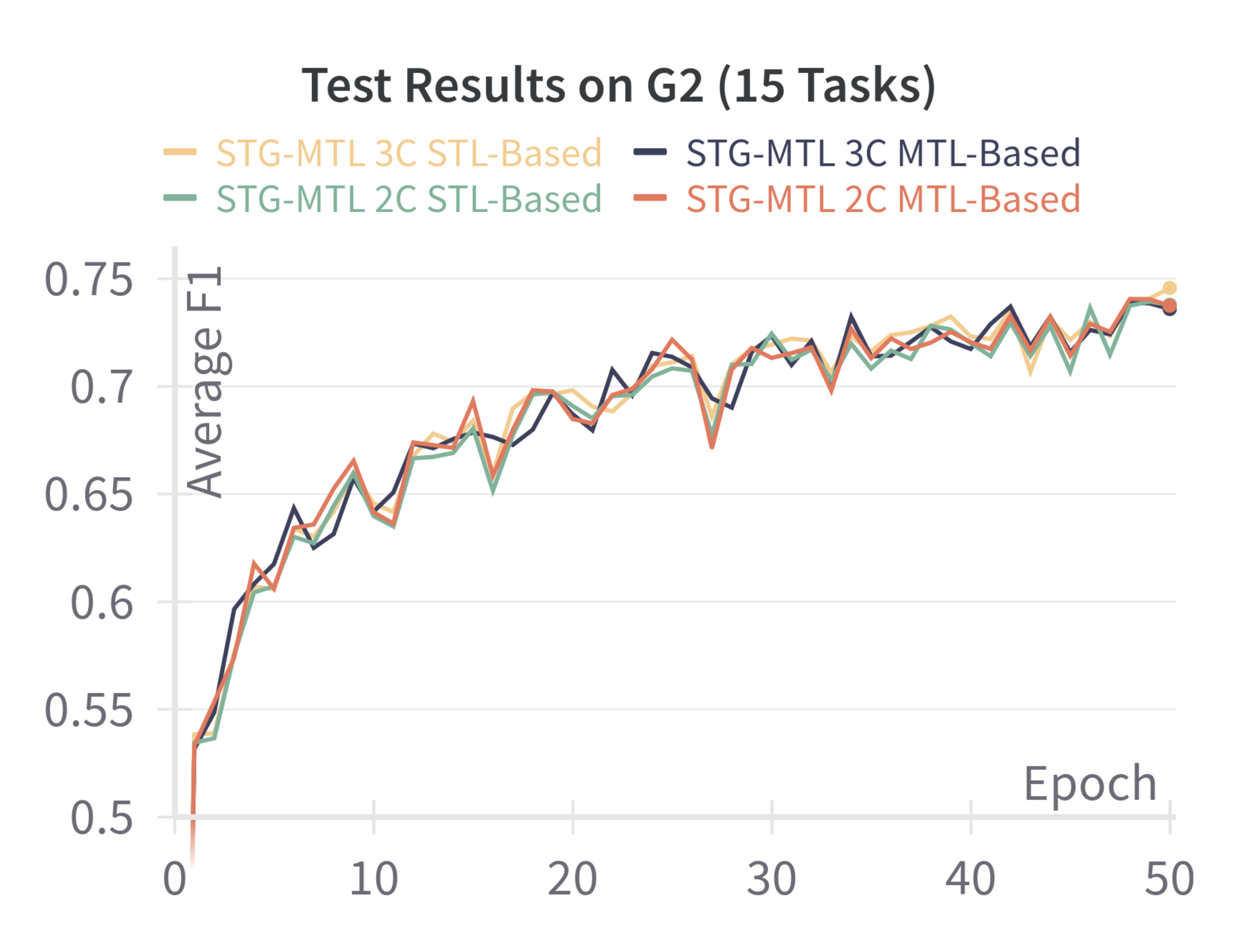}}
            \subcaption{Comparing Data Map Sources (G2)}\label{fig_custom_curve_mtl_stl_G2}
        \end{minipage}
    \end{minipage}
    \hfill
    \begin{minipage}[c]{0.24\columnwidth}
        \begin{minipage}[c]{\columnwidth}
            \centerline{\includegraphics[width=\columnwidth]{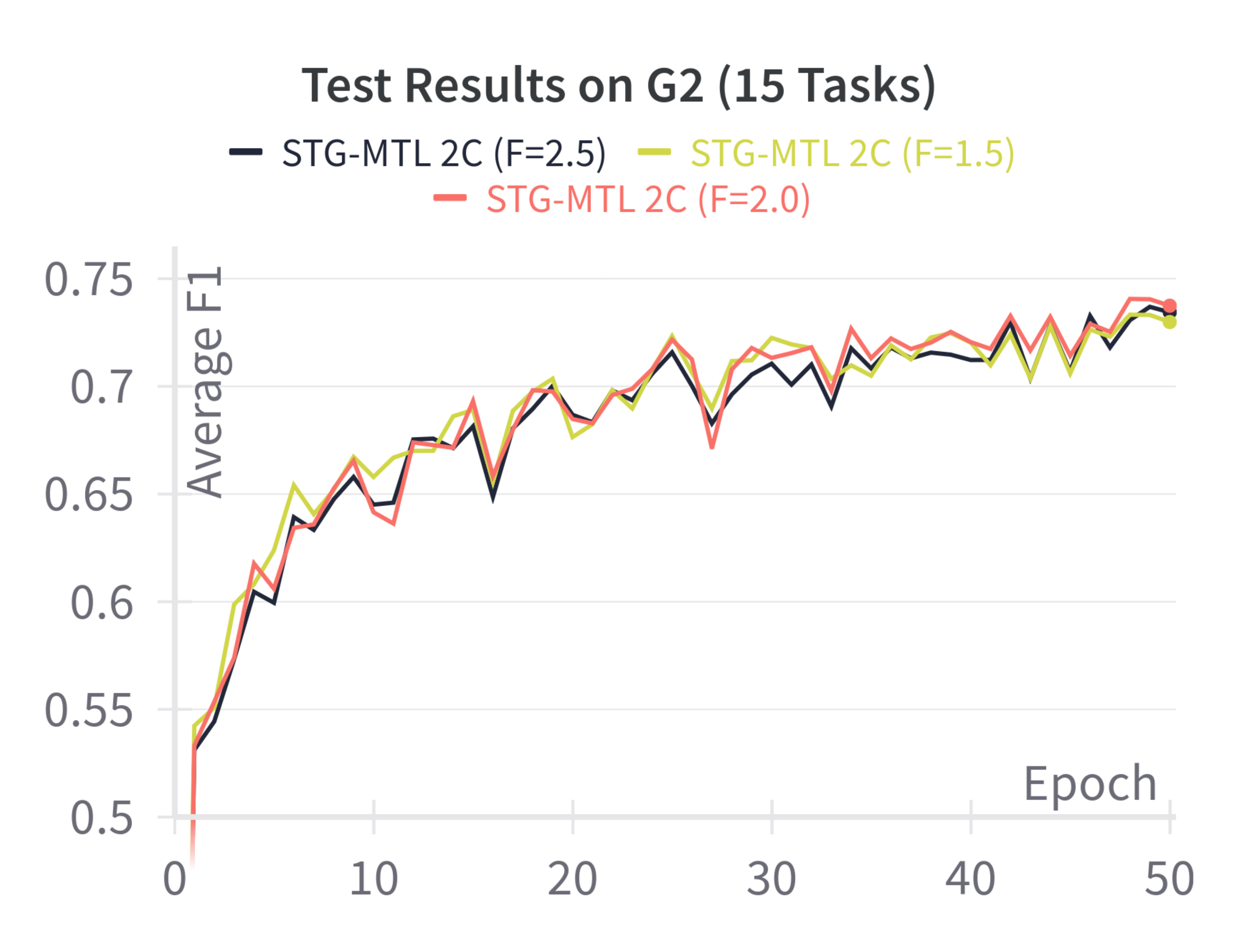}}
            \subcaption{Different $F$ using $2$ clusters (G2)}\label{fig_custom_curve_2c_G2}
        \end{minipage}
        \\
        \begin{minipage}[c]{\columnwidth}
            \centerline{\includegraphics[width=\columnwidth]{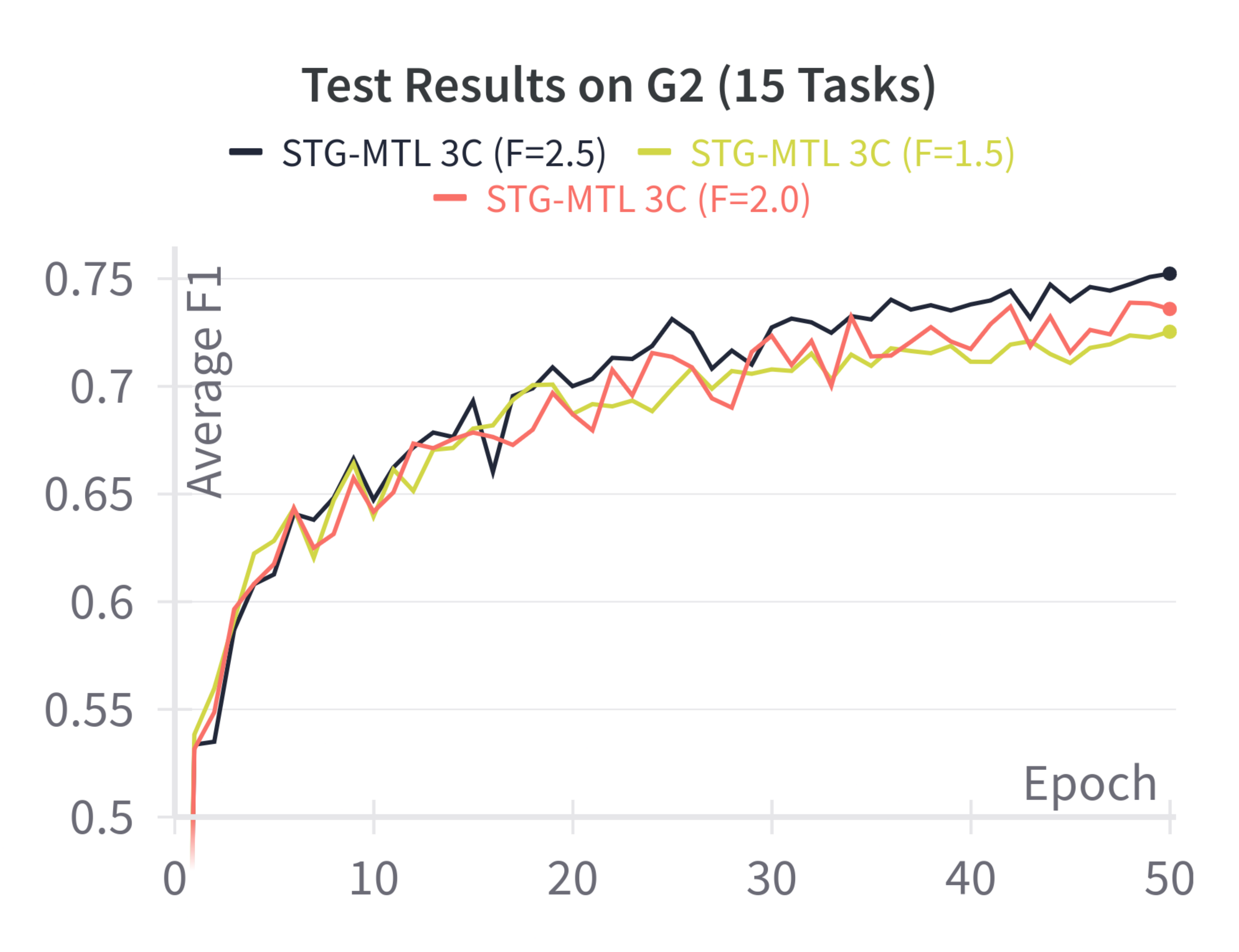}}
            \subcaption{Different $F$ using $3$ clusters (G2)}\label{fig_custom_curve_3c_G2}
        \end{minipage}
    \end{minipage}
    \hfill
    \begin{minipage}[c]{0.24\columnwidth}
        \begin{minipage}[c]{\columnwidth}
            \centerline{\includegraphics[width=\columnwidth]{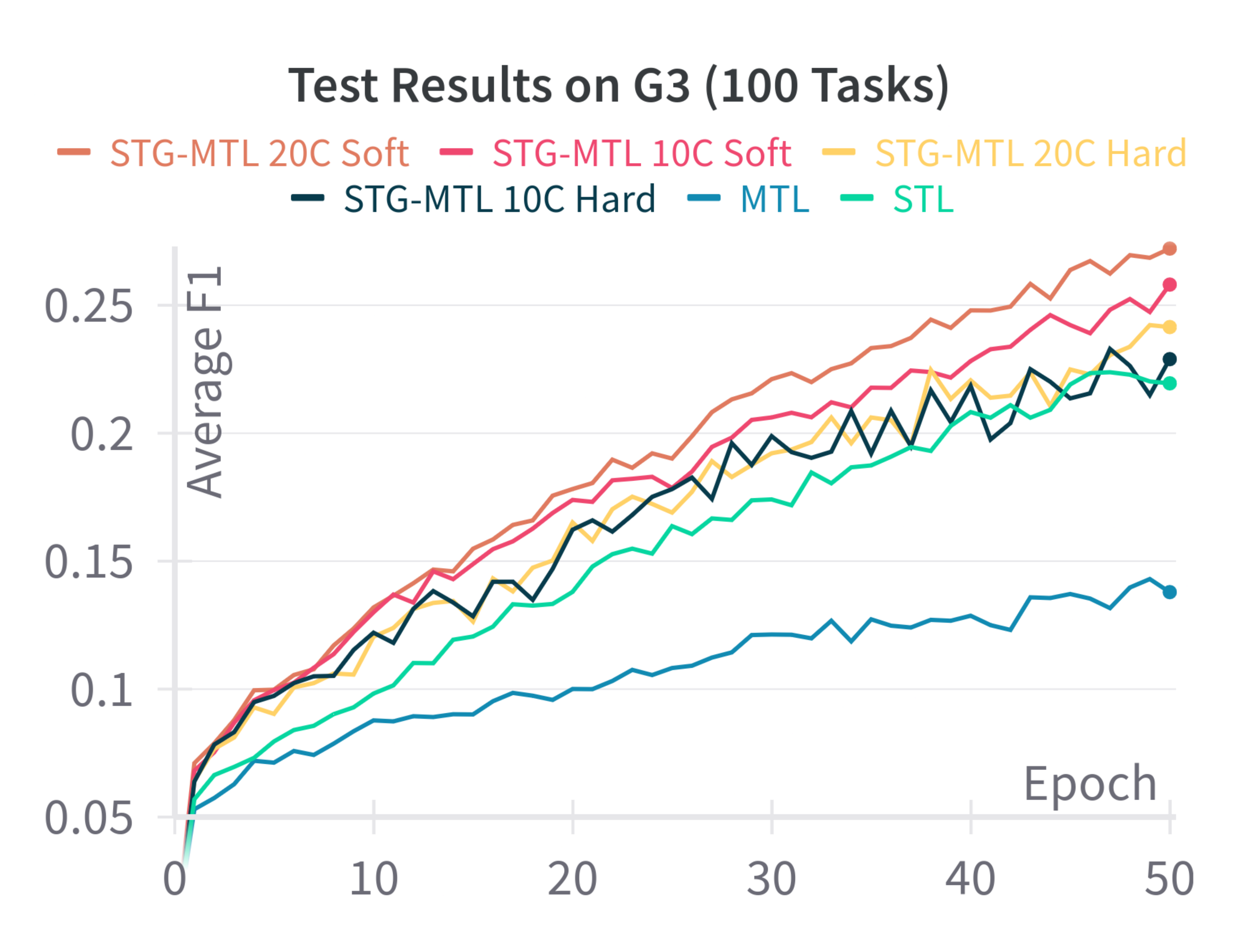}}
            \subcaption{MTL vs STG vs STL Results (G3)}\label{fig_custom_curve_mtl_stg_stl_G3}
        \end{minipage}
        \\
        \begin{minipage}[c]{\columnwidth}
            \centerline{\includegraphics[width=\columnwidth]{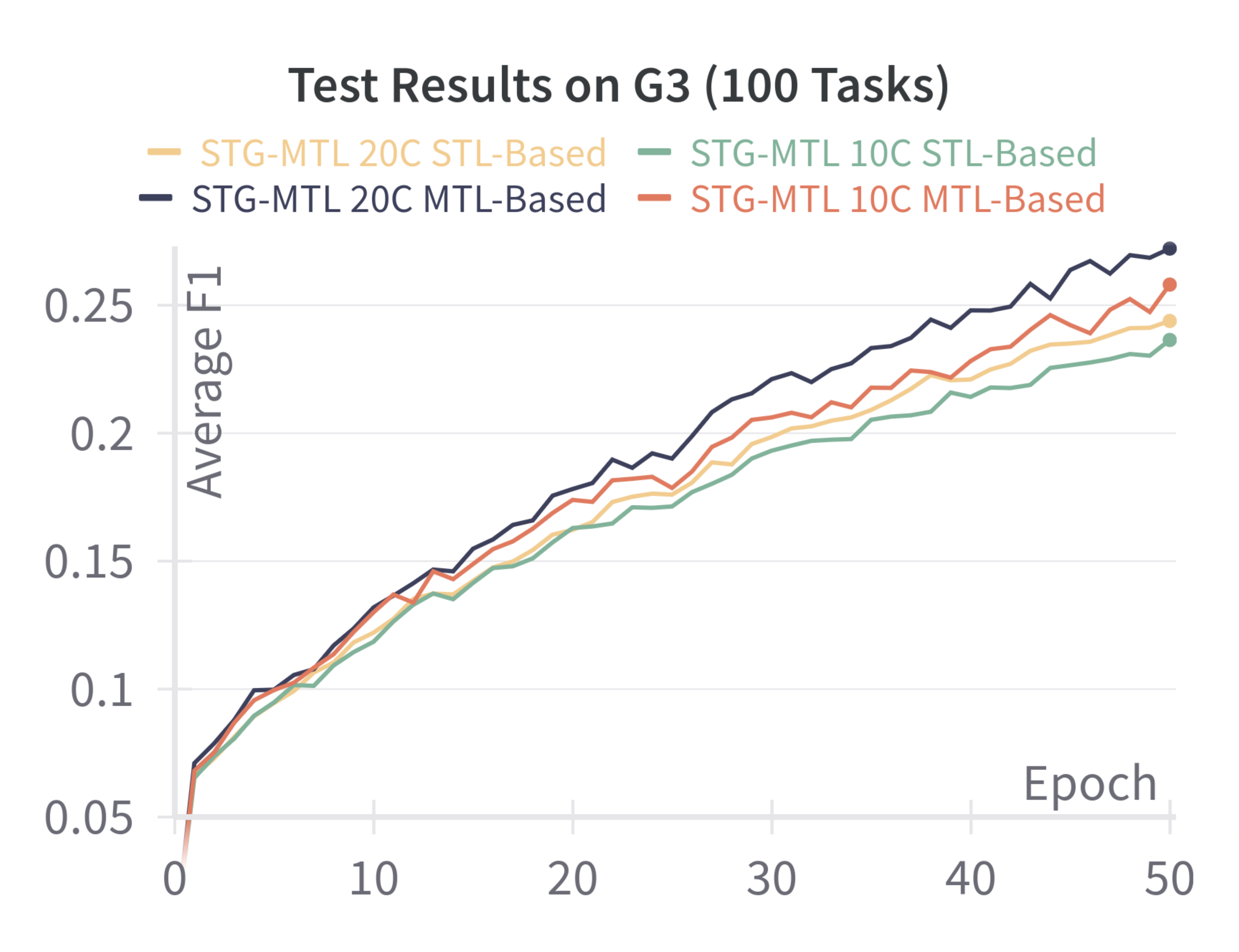}}
            \subcaption{Comparing Data Map Sources (G3)}\label{fig_custom_curve_mtl_stl_G3}
        \end{minipage}
    \end{minipage}
    \hfill
    \begin{minipage}[c]{0.24\columnwidth}
        \centerline{\includegraphics[width=\columnwidth]{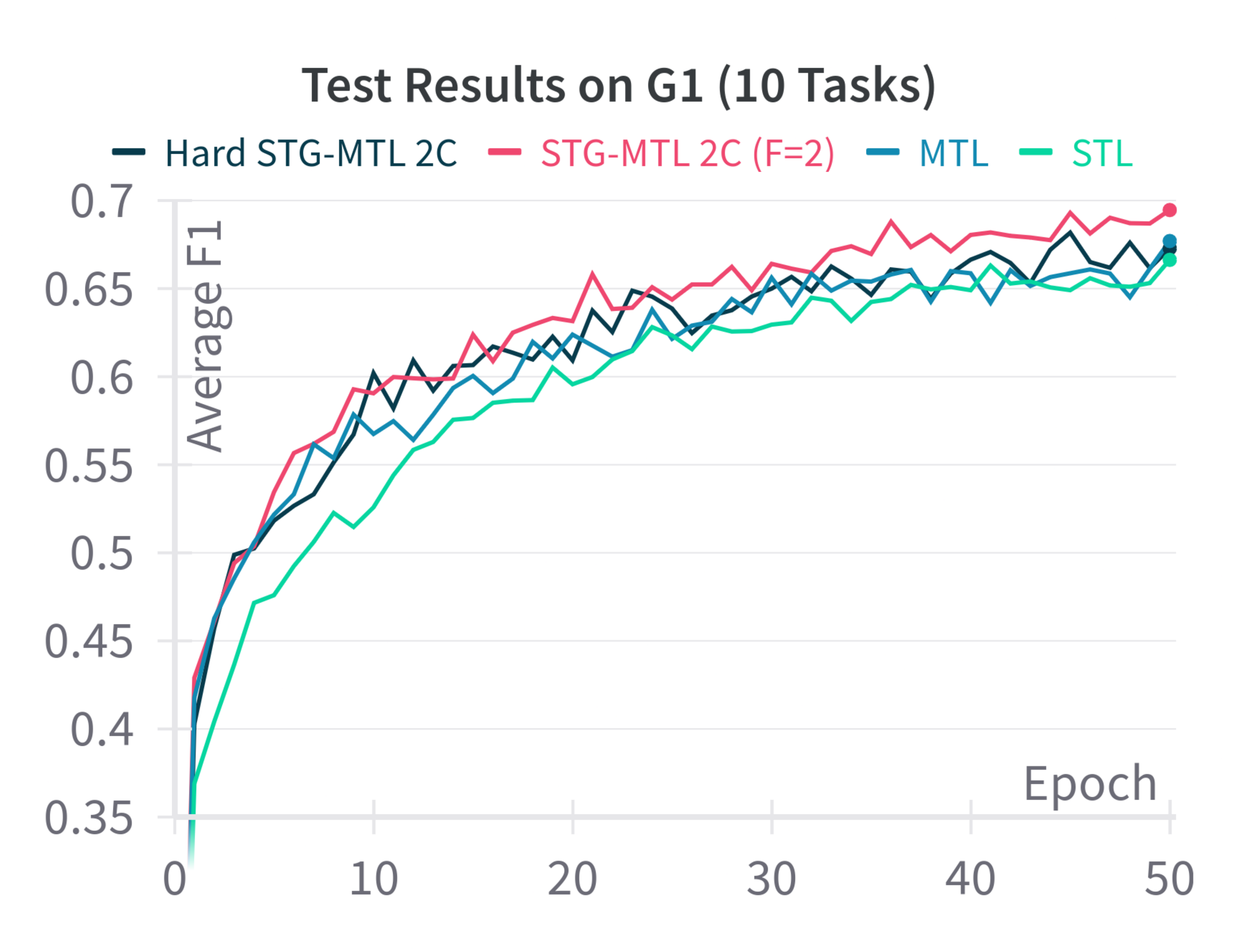}}
        \subcaption{MTL vs STG vs STL Results (G1)}\label{fig_custom_curve_G1_test}
    \end{minipage}
    \caption{Average F1 score curves on test set generated using the custom CNN}\label{fig_f1_custom_eval}
\end{figure}

We also present additional experiments, figure \ref{fig_f1_custom_eval} using a simpler model with much less number of parameters, of a custom CNN architecture depicted in Table \ref{table_custom_model}, in Appendix \ref{appendix_custom_cnn_exp}. These experiments demonstrate the robustness and adaptability of our method across different model architectures. It is worth mentioning that even with \textit{hard clustering}, our method outperforms standard MTL, showcasing its effectiveness in leveraging task grouping for enhanced performance as shown in Figures \ref{fig_custom_curve_mtl_stg_stl_G3} and \ref{fig_custom_curve_G1_test}. Moreover, the results obtained using our approach with data maps extracted from the MTL model are comparable to those from the STL models as in Figures \ref{fig_custom_curve_mtl_stl_G2} and \ref{fig_custom_curve_mtl_stl_G3}.

Overall, our method consistently outperforms both the MTL and STL approaches, indicating that the task grouping provides valuable information for improving task performance. Notably, although our method tends to overfit, in RESNET18 experiments, and achieves excellent training performance, it also achieves the best performance on the test set. This suggests that if the models were further fine-tuned, even greater gains could be achieved, yet we refrain from tuning any of the models in this study to guarantee fairness in comparison.

\section{Conclusion and Future Work}

In conclusion, we have presented STG-MTL, which is a novel scalable approach for task grouping in multi-task learning (MTL) settings. Our method utilizes data maps \cite{data_maps} to identify task similarities and group them accordingly. We showed its superior scalability theoretically in comparison to TAG \cite{TAG}, HOA \cite{which_tasks_to_learn}, and MTG-Net \cite{MTG_task_grouping_meta_learner}. We have also demonstrated the effectiveness of our method through our experiments on CIFAR10 \& CIFAR100 \cite{krizhevsky2009learning_cifar} and CelebA \cite{liu2015faceattributes} datasets , where we pushed the boundaries by experimenting with \textbf{100 tasks}, which has never been done before in the literature proving its scalability. We have also compared our clustering results against the predefined superclasses in CIFAR100, further validating the effectiveness of our approach. Nevertheless due to our limited computational power and the poor scalability of the current methods, we were not able to evaluate the performance of our results against the other methods. Instead, we showed that our method outperformed both traditional MTL and single-task learning (STL) approaches, showcasing the quality of task grouping and its ability to improve multi-task learning performance.

For future work, we aim to explore Data Maps' generalization to other task types, such as regression, because they are currently limited to classification tasks only. Additionally, we hope our research could open a new research direction in the MTL community to explore the development of new features that can capture the training dynamics efficiently, other than data maps. By advancing this research direction, we can unlock new possibilities for enhancing performance and driving further advancements in the field of MTL.

\textbf{Acknowledgments} We gratefully thank the Fatima Fellowship\footnote{\url{https://www.fatimafellowship.com/}} for supporting this research especially during its early stages.

\bibliography{ref}

\newpage

\appendix
\begin{figure*}[ht!]
    \centering
    \begin{minipage}[c]{0.32\columnwidth}
        \centerline{\includegraphics[width=\columnwidth]{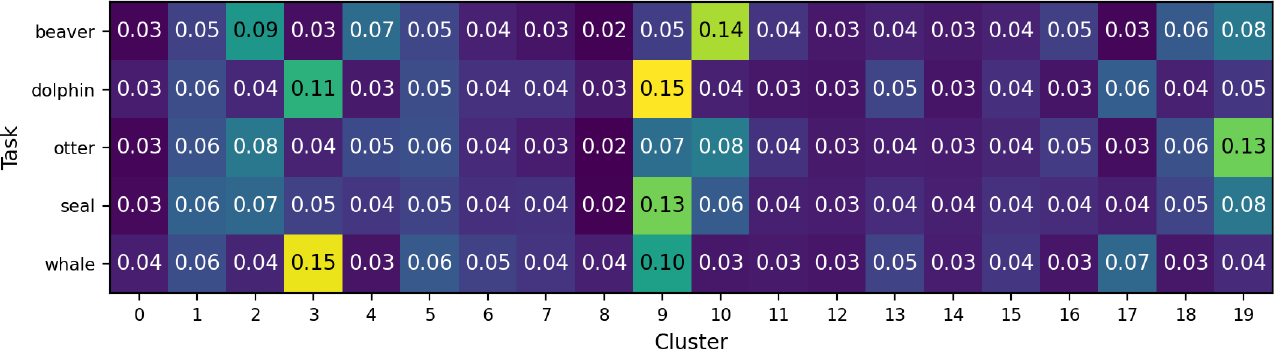}}
        \subcaption{aquatic mammals }\label{fig_G3_20C_aquatic_mammals}
    \end{minipage}\
    \begin{minipage}[c]{0.32\columnwidth}
        \centerline{\includegraphics[width=\columnwidth]{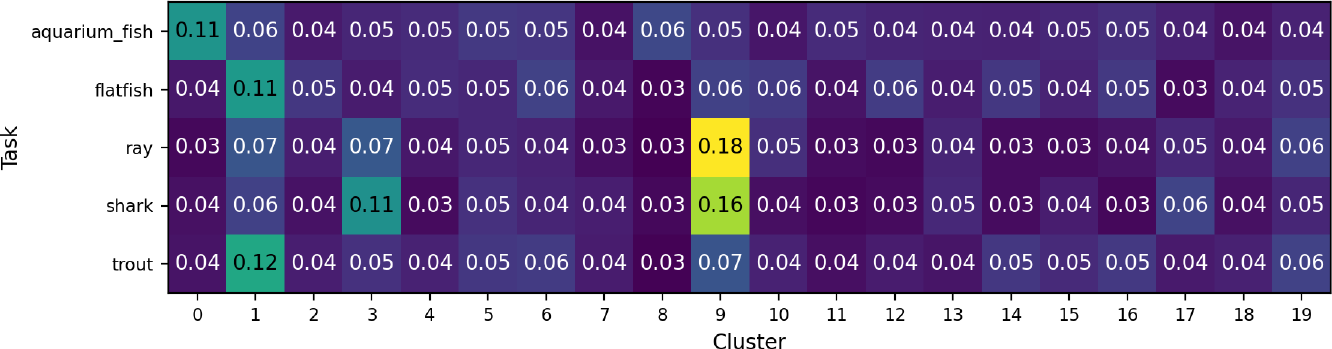}}
        \subcaption{fish}\label{fig_G3_20C_fish}
    \end{minipage}\
    \begin{minipage}[c]{0.32\columnwidth}
        \centerline{\includegraphics[width=\columnwidth]{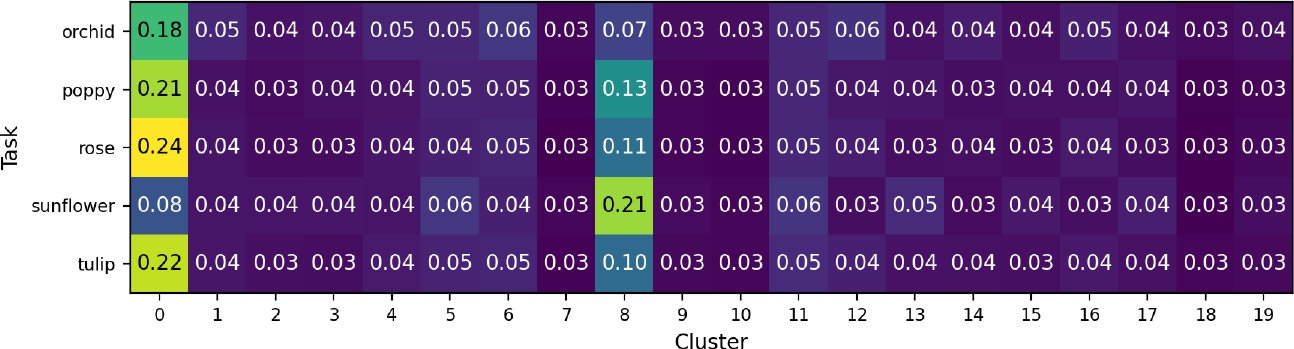}}
        \subcaption{flowers}\label{fig_G3_20C_flowers}
    \end{minipage}\\
    \begin{minipage}[c]{0.32\columnwidth}
        \centerline{\includegraphics[width=\columnwidth]{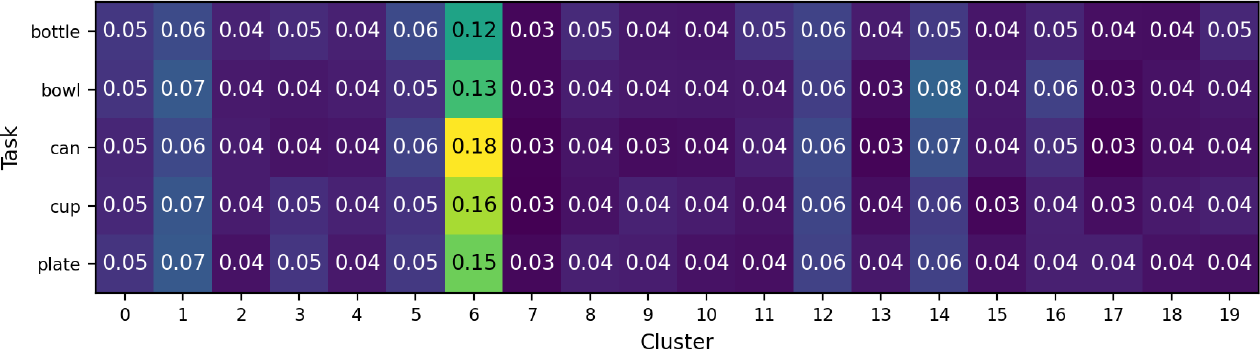}}
        \subcaption{food containers}\label{fig_G3_20C_food_containers}
    \end{minipage}\
    \begin{minipage}[c]{0.32\columnwidth}
        \centerline{\includegraphics[width=\columnwidth]{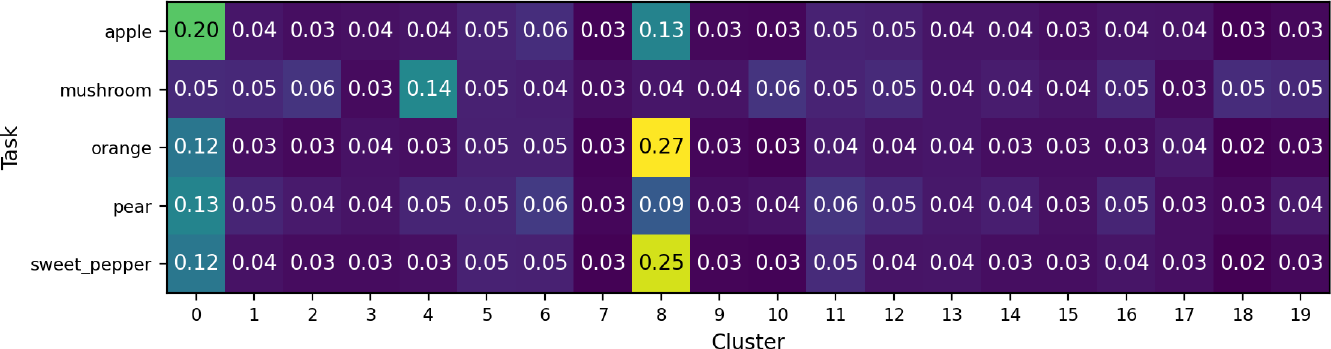}}
        \subcaption{fruit and vegetables}\label{fig_G3_20C_fruit_and_vegetables}
    \end{minipage}\
    \begin{minipage}[c]{0.32\columnwidth}
        \centerline{\includegraphics[width=\columnwidth]{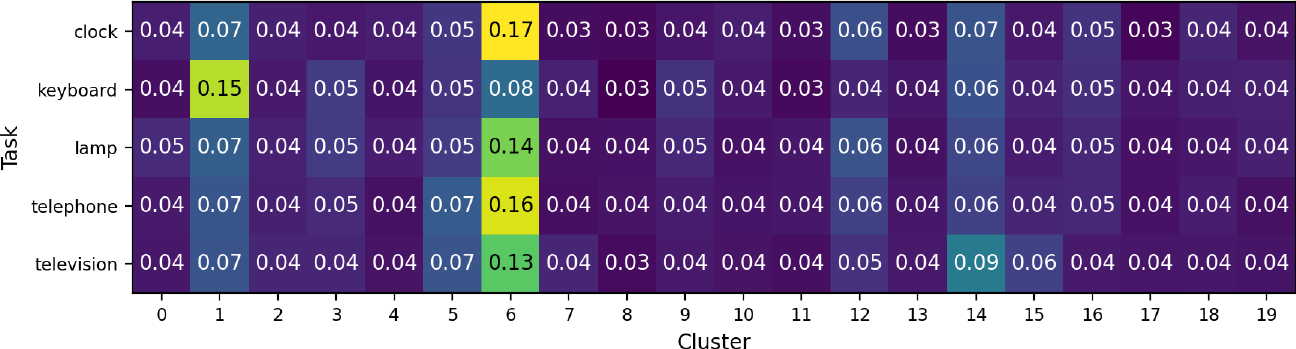}}
        \subcaption{household electrical devices}\label{fig_G3_20C_household_electrical_devices}
    \end{minipage}\\
    \begin{minipage}[c]{0.32\columnwidth}
        \centerline{\includegraphics[width=\columnwidth]{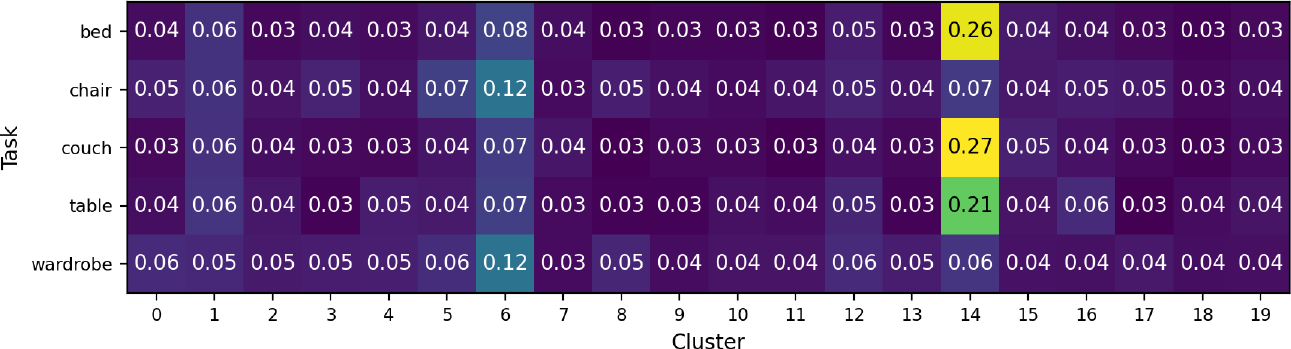}}
        \subcaption{household furniture}\label{fig_G3_20C_household_furniture}
    \end{minipage}\
    \begin{minipage}[c]{0.32\columnwidth}
        \centerline{\includegraphics[width=\columnwidth]{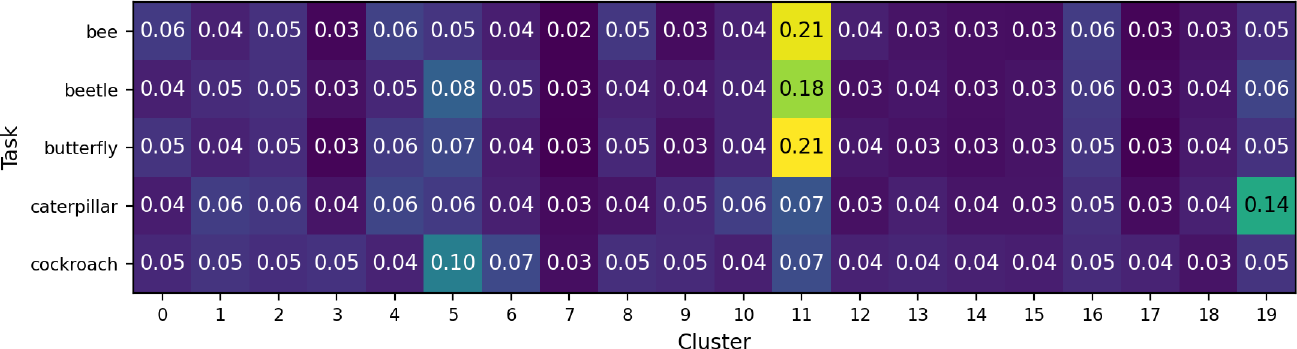}}
        \subcaption{insects}\label{fig_G3_20C_insects}
    \end{minipage}\
    \begin{minipage}[c]{0.32\columnwidth}
        \centerline{\includegraphics[width=\columnwidth]{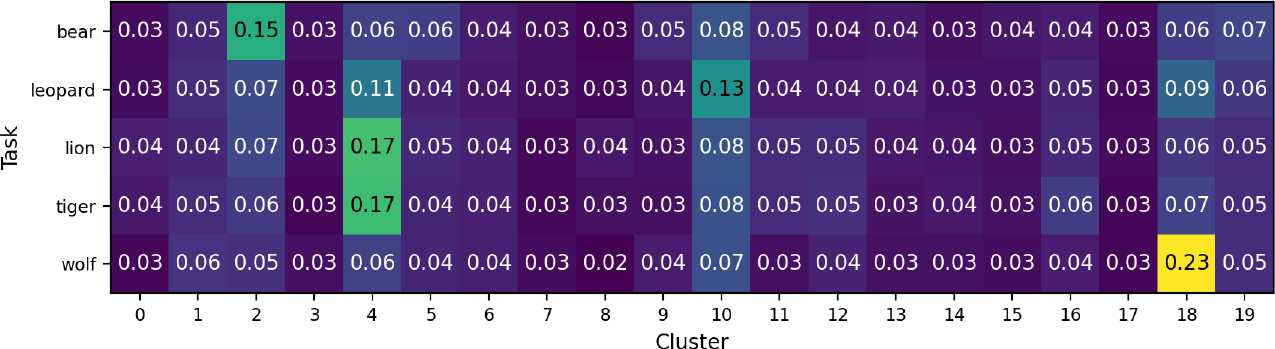}}
        \subcaption{large carnivores}\label{fig_G3_20C_large_carnivores}
    \end{minipage}\\
    \begin{minipage}[c]{0.32\columnwidth}
        \centerline{\includegraphics[width=\columnwidth]{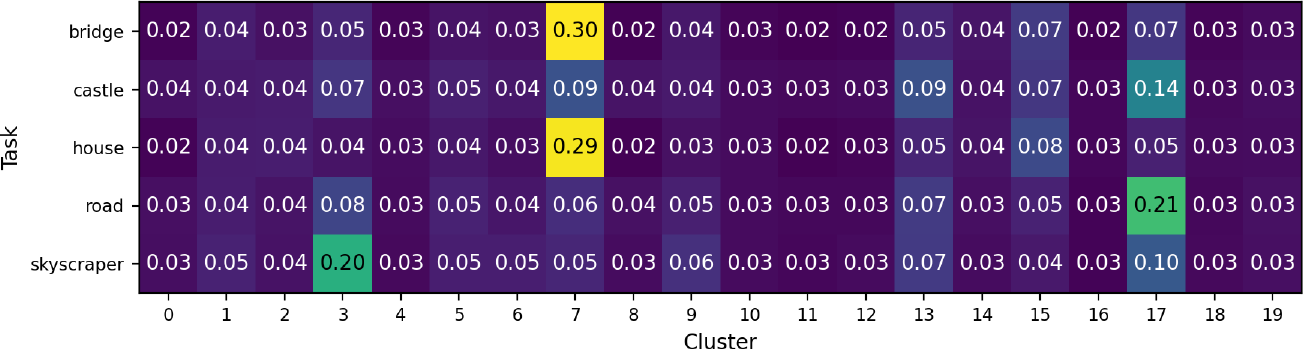}}
        \subcaption{large man-made outdoor things}\label{fig_G3_20C_large_man_made_outdoor_things}
    \end{minipage}\
    \begin{minipage}[c]{0.32\columnwidth}
        \centerline{\includegraphics[width=\columnwidth]{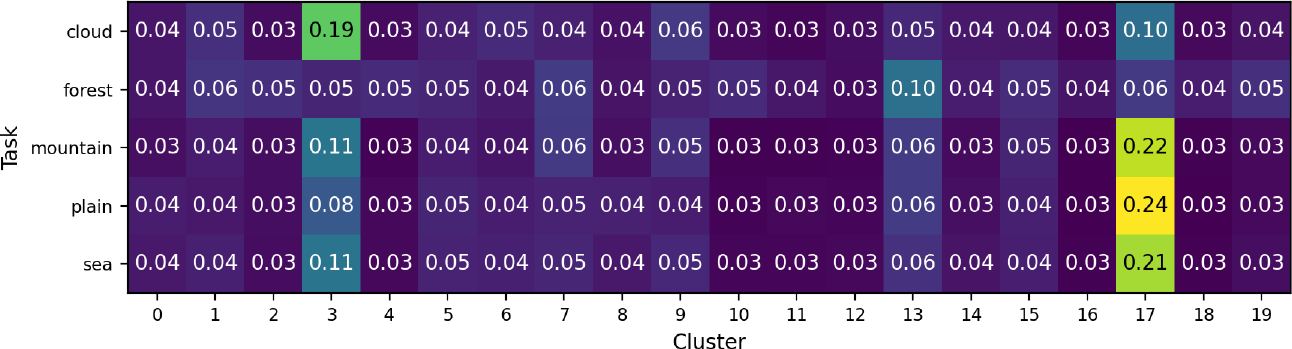}}
        \subcaption{large natural outdoor scenes}\label{fig_G3_20C_large_natural_outdoor_scenes}
    \end{minipage}\
    \begin{minipage}[c]{0.32\columnwidth}
        \centerline{\includegraphics[width=\columnwidth]{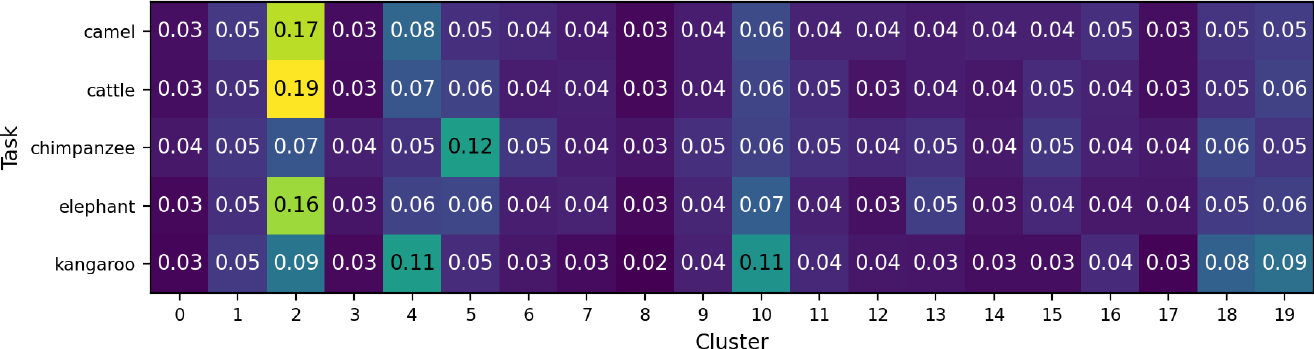}}
        \subcaption{large omnivores and herbivores}\label{fig_G3_20C_large_omnivores_and_herbivores}
    \end{minipage}\\
    \begin{minipage}[c]{0.32\columnwidth}
        \centerline{\includegraphics[width=\columnwidth]{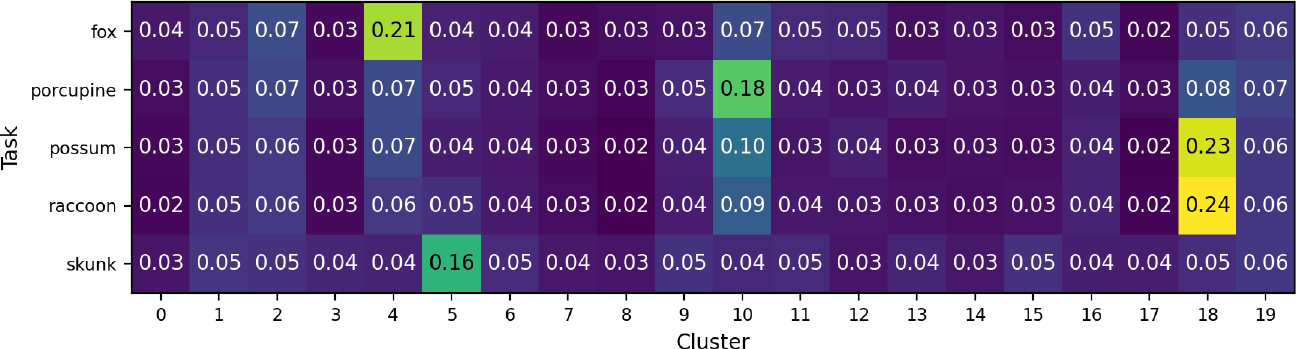}}
        \subcaption{medium-sized mammals}\label{fig_G3_20C_medium_sized_mammals}
    \end{minipage}\
    \begin{minipage}[c]{0.32\columnwidth}
        \centerline{\includegraphics[width=\columnwidth]{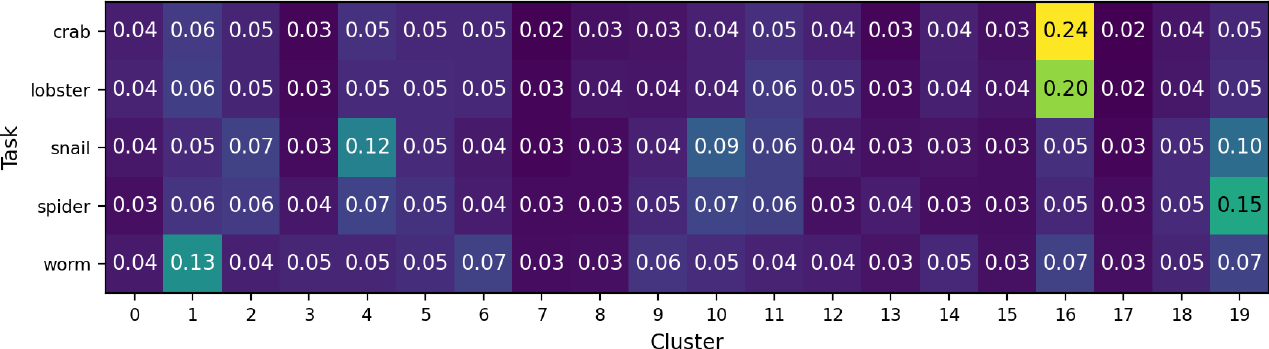}}
        \subcaption{non-insect invertebrates}\label{fig_G3_20C_non_insect_invertebrates}
    \end{minipage}\
    \begin{minipage}[c]{0.32\columnwidth}
        \centerline{\includegraphics[width=\columnwidth]{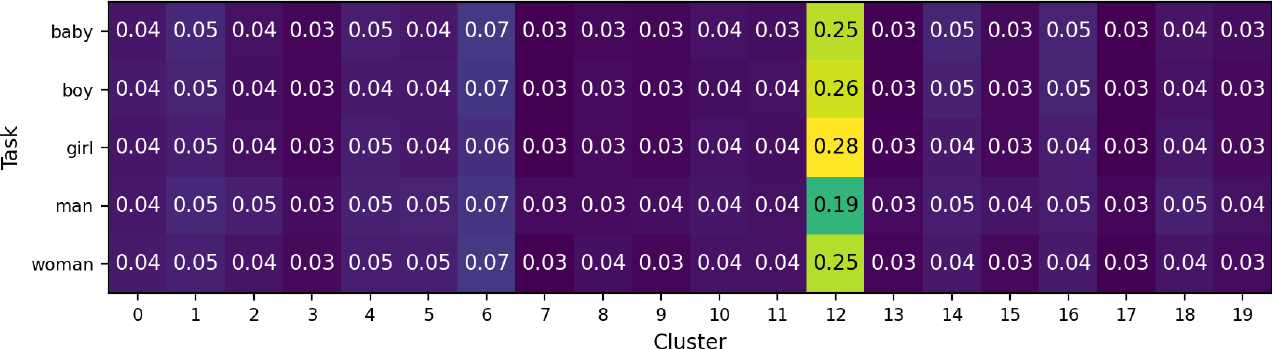}}
        \subcaption{people}\label{fig_G3_20C_people}
    \end{minipage}\\
    \begin{minipage}[c]{0.32\columnwidth}
        \centerline{\includegraphics[width=\columnwidth]{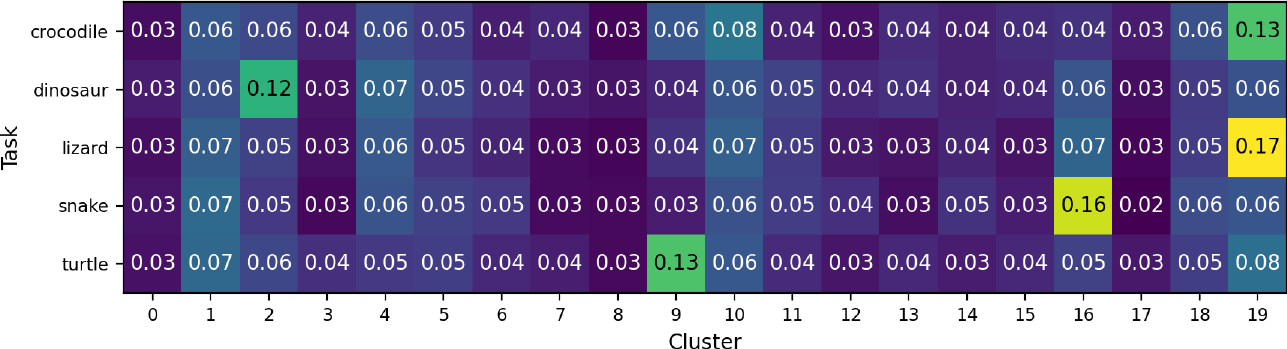}}
        \subcaption{reptiles}\label{fig_G3_20C_reptiles}
    \end{minipage}\
    \begin{minipage}[c]{0.32\columnwidth}
        \centerline{\includegraphics[width=\columnwidth]{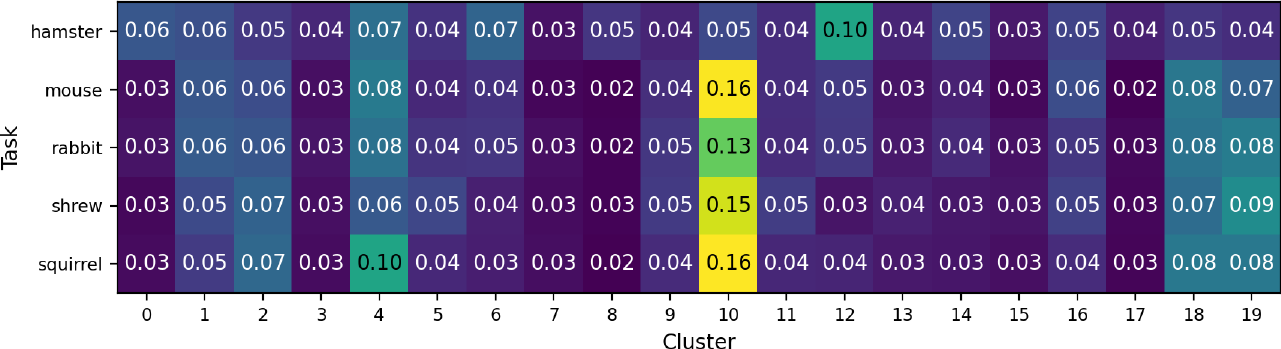}}
        \subcaption{small mammals}\label{fig_G3_20C_small_mammals}
    \end{minipage}\
    \begin{minipage}[c]{0.32\columnwidth}
        \centerline{\includegraphics[width=\columnwidth]{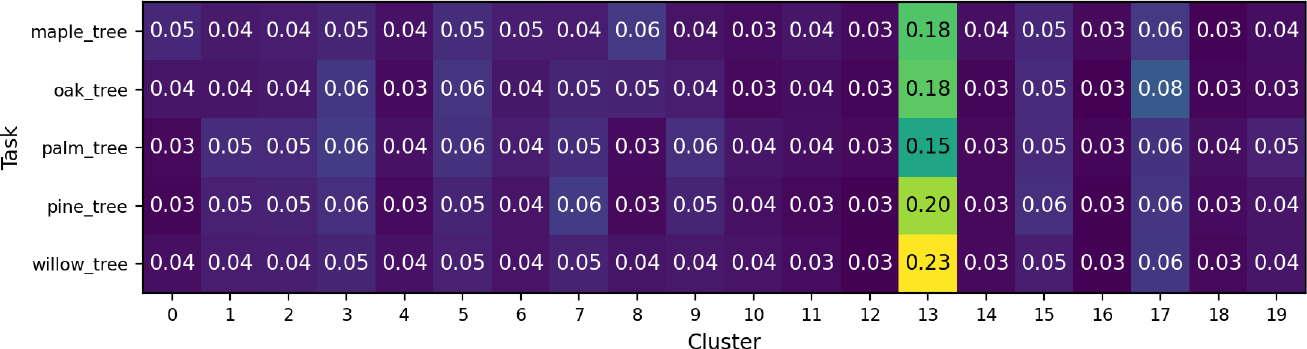}}
        \subcaption{trees}\label{fig_G3_20C_trees}
    \end{minipage}\\
    \begin{minipage}[c]{0.32\columnwidth}
        \centerline{\includegraphics[width=\columnwidth]{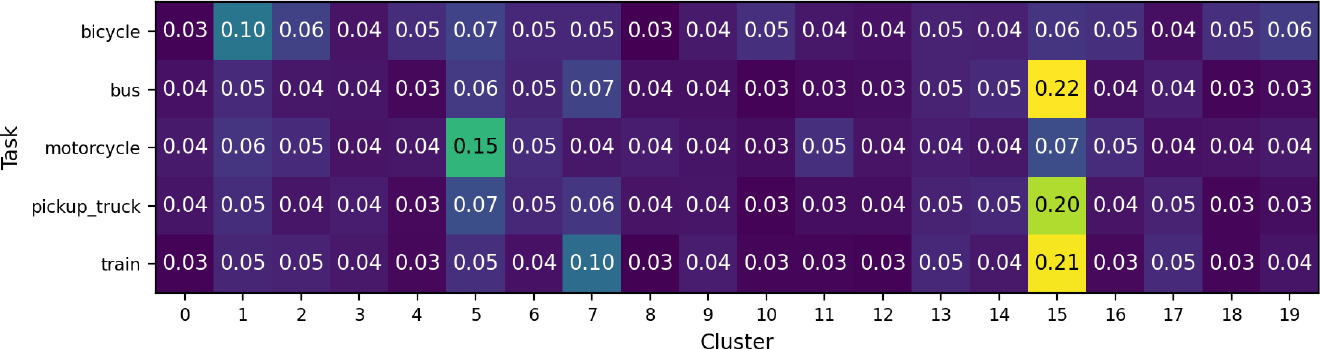}}
        \subcaption{vehicles 1}\label{fig_G3_20C_vehicles_1}
    \end{minipage}\
    \begin{minipage}[c]{0.32\columnwidth}
        \centerline{\includegraphics[width=\columnwidth]{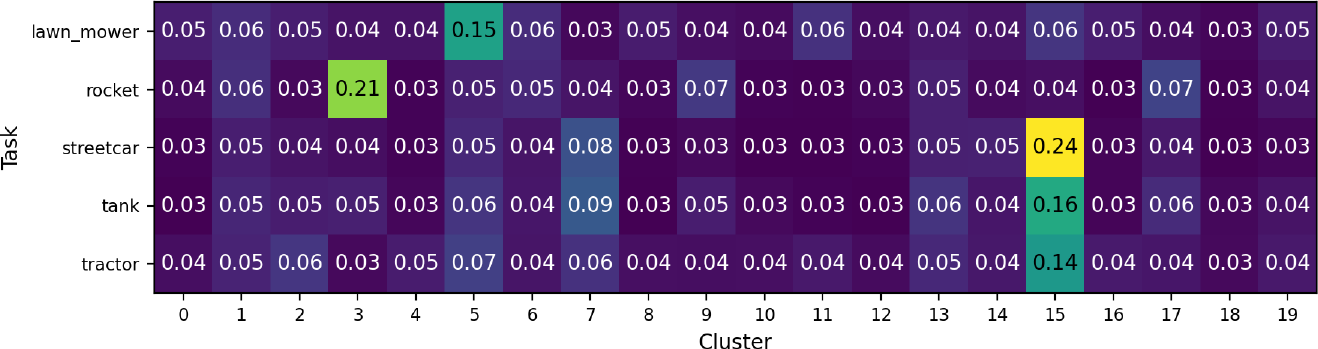}}
        \subcaption{vehicles 2}\label{fig_G3_20C_vehicles_2}
    \end{minipage}\\
    \caption{Task grouping of G3 (100 Tasks of CIFAR100) into $20$ clusters with $F=2$ using Data Maps from RESNET18}\label{fig_G3_20C}
\end{figure*}
\section{Task Clustering Results of 100 Tasks} \label{appendix_task_clustering}
\begin{figure*}[h!]
    \centering
    \begin{minipage}[c]{0.32\columnwidth}
        \centerline{\includegraphics[width=\columnwidth]{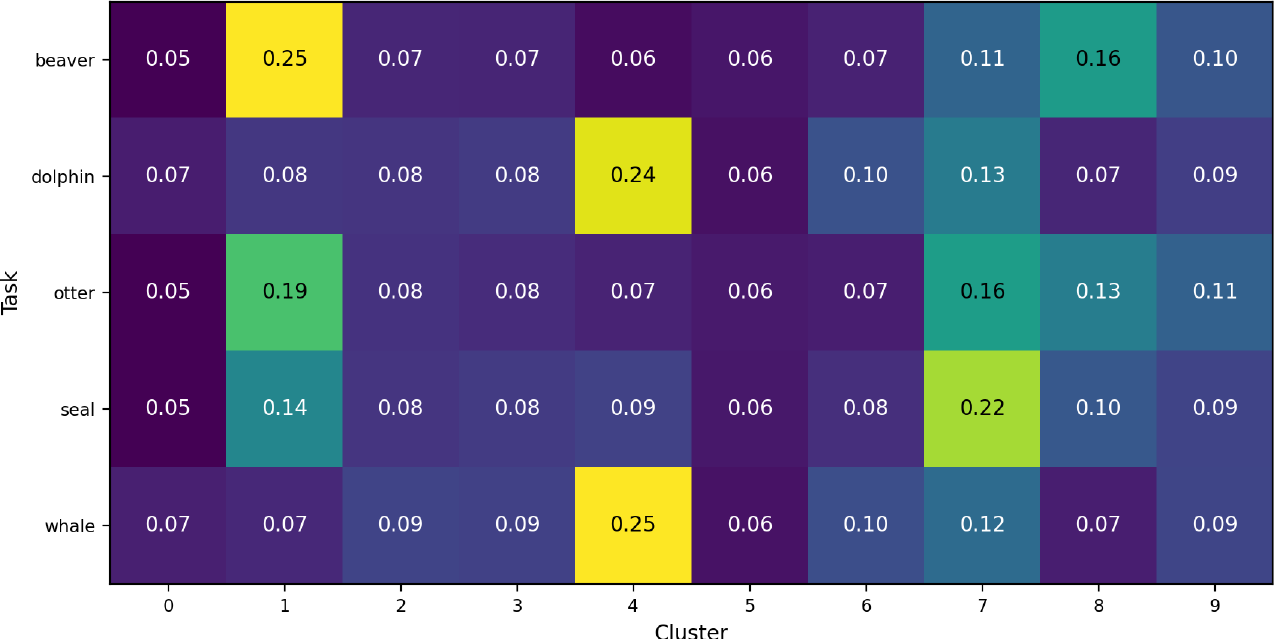}}
        \subcaption{aquatic mammals}\label{fig_G3_10C_aquatic_mammals}
    \end{minipage}\
    \begin{minipage}[c]{0.32\columnwidth}
        \centerline{\includegraphics[width=\columnwidth]{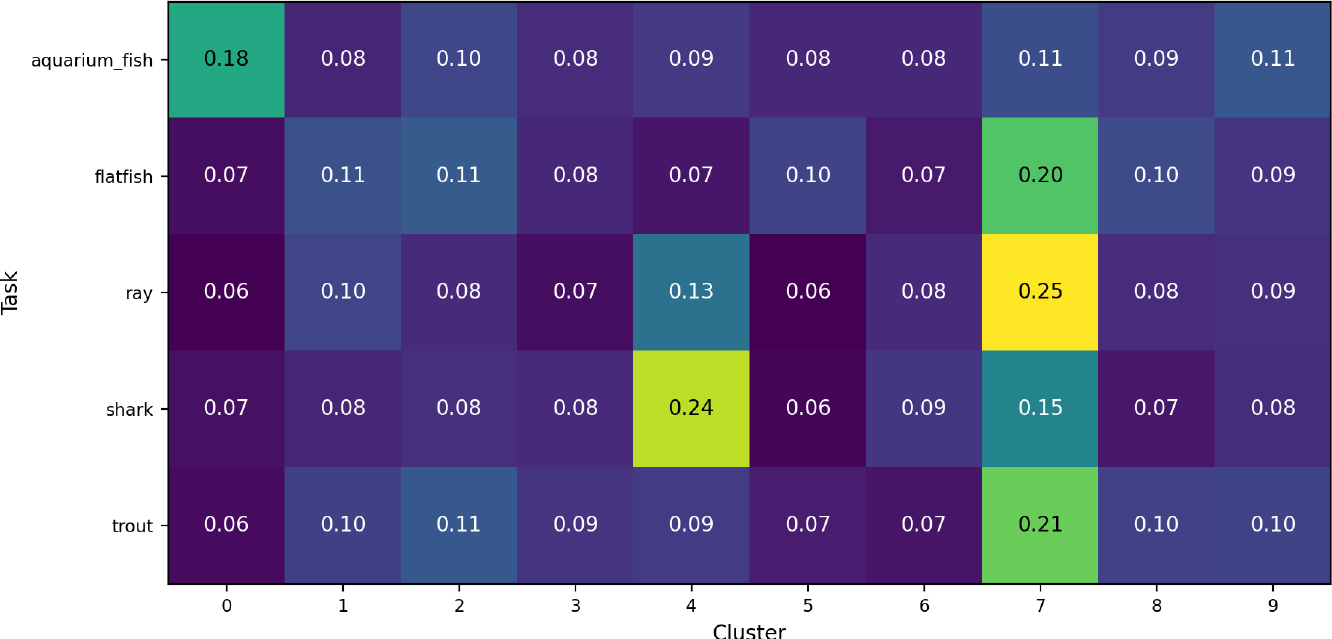}}
        \subcaption{fish}\label{fig_G3_10C_fish}
    \end{minipage}\
    \begin{minipage}[c]{0.32\columnwidth}
        \centerline{\includegraphics[width=\columnwidth]{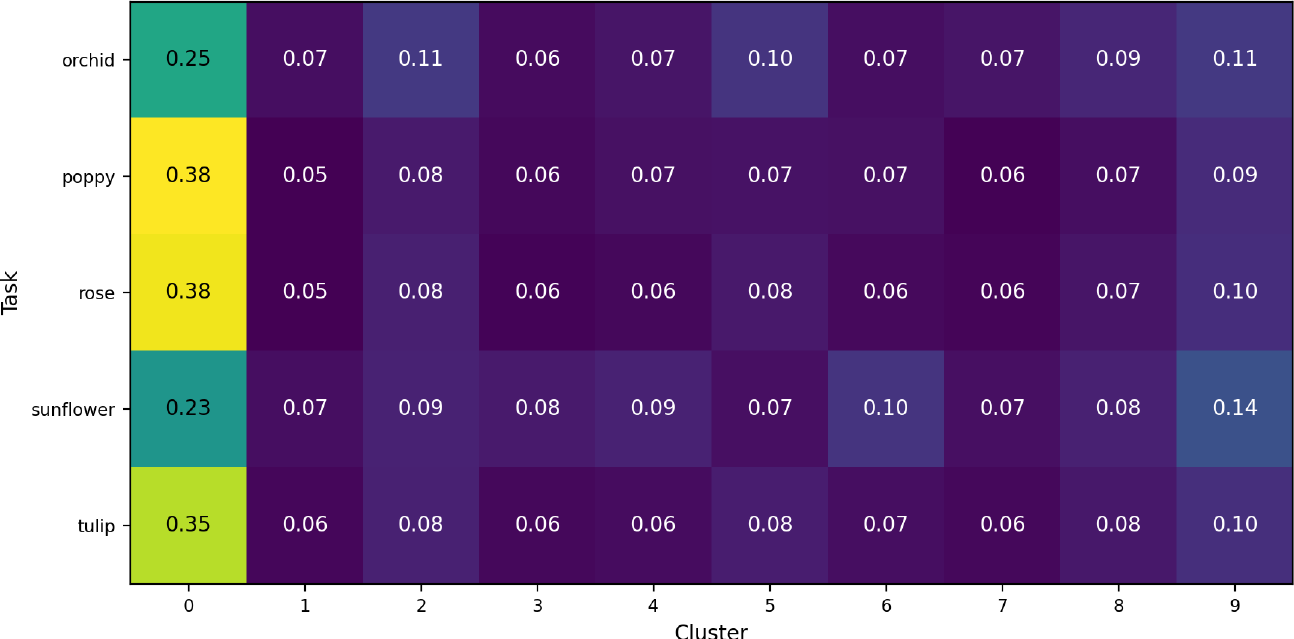}}
        \subcaption{flowers}\label{fig_G3_10C_flowers}
    \end{minipage}\\
    \begin{minipage}[c]{0.32\columnwidth}
        \centerline{\includegraphics[width=\columnwidth]{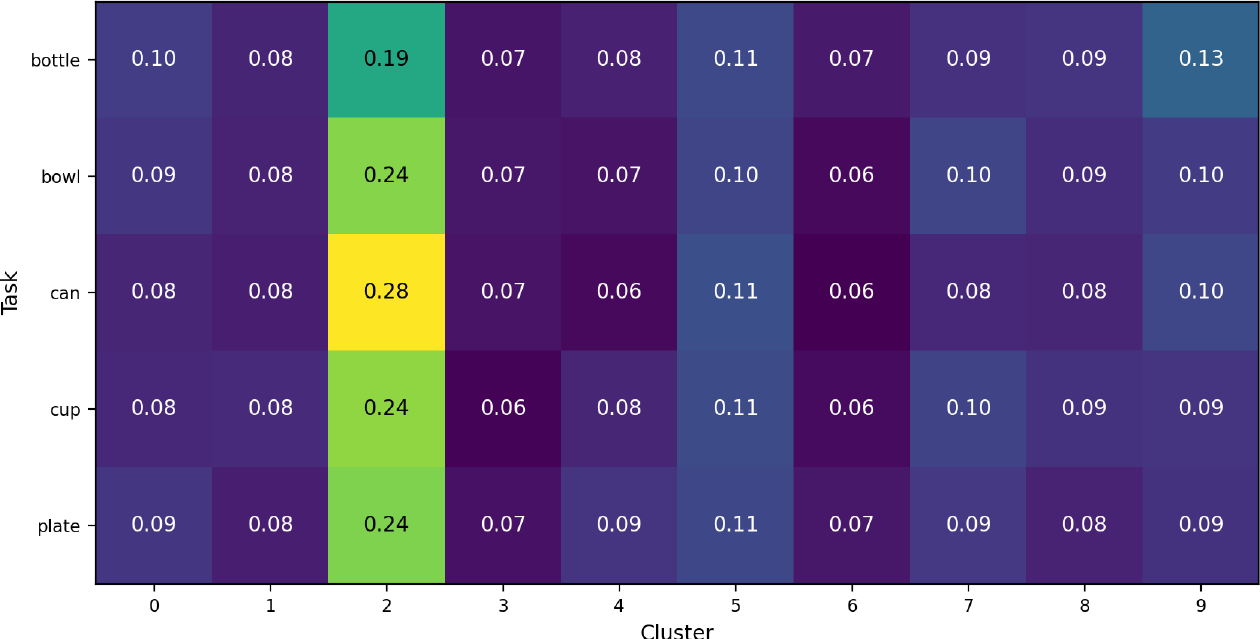}}
        \subcaption{food containers}\label{fig_G3_10C_food_containers}
    \end{minipage}\
    \begin{minipage}[c]{0.32\columnwidth}
        \centerline{\includegraphics[width=\columnwidth]{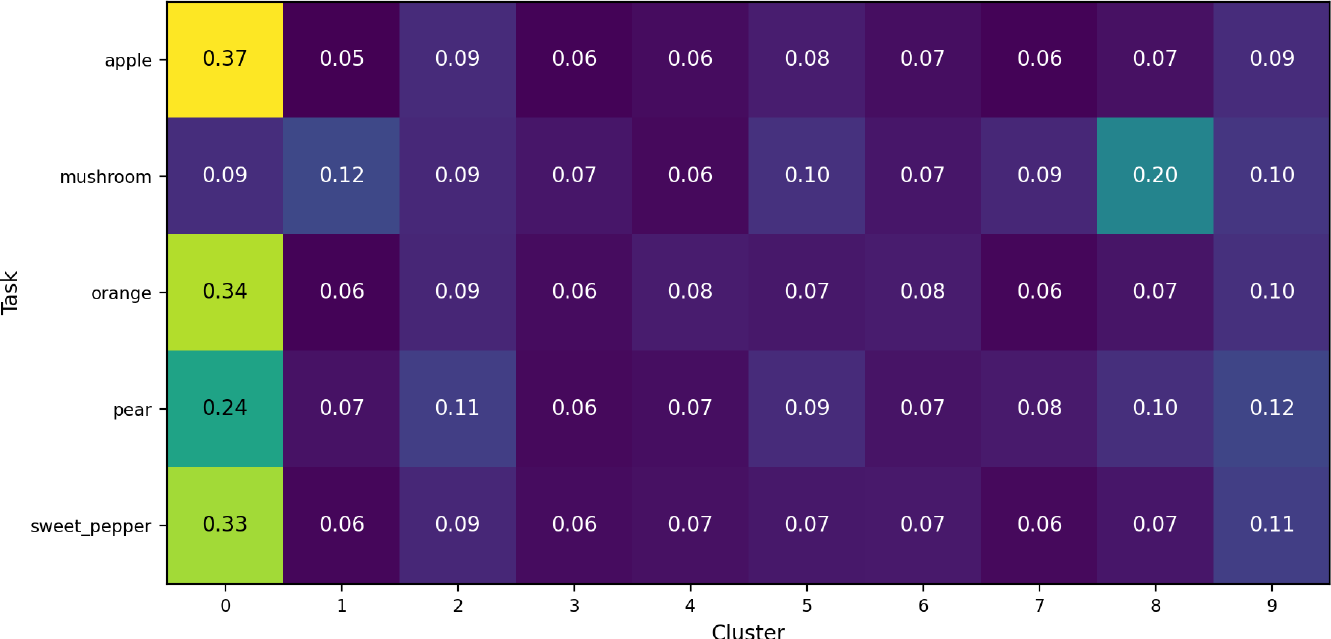}}
        \subcaption{fruit and vegetables}\label{fig_G3_10C_fruit_and_vegetables}
    \end{minipage}\
    \begin{minipage}[c]{0.32\columnwidth}
        \centerline{\includegraphics[width=\columnwidth]{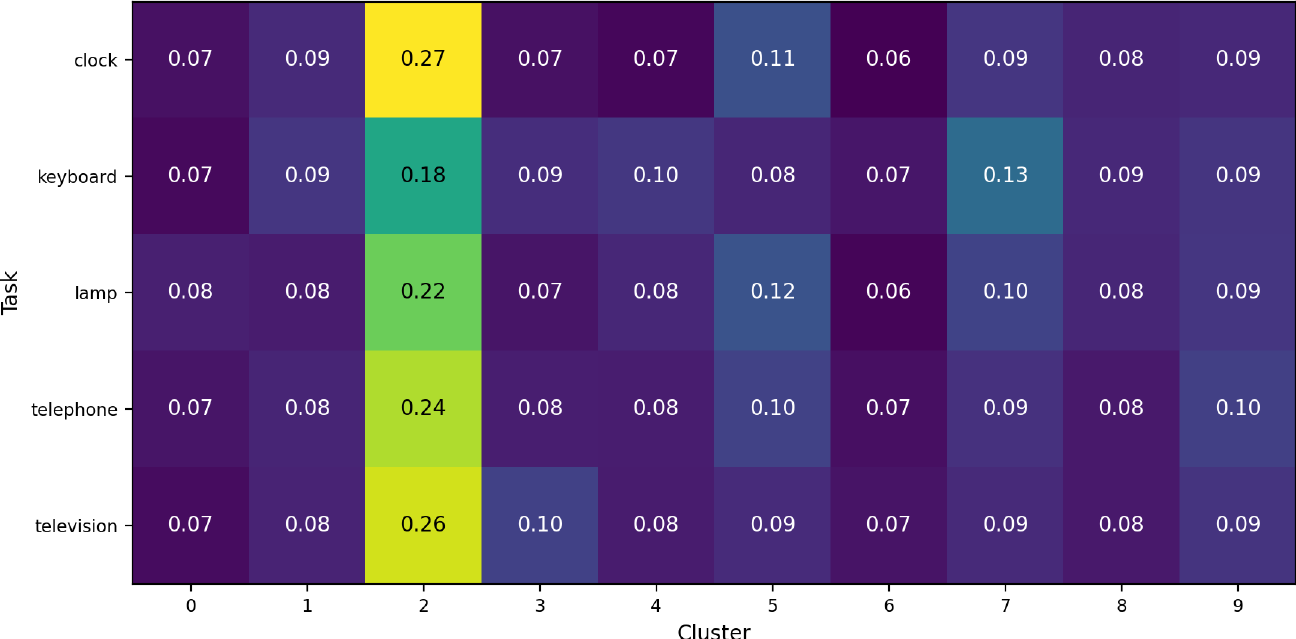}}
        \subcaption{household electrical devices}\label{fig_G3_10C_household_electrical_devices}
    \end{minipage}\\
    \begin{minipage}[c]{0.32\columnwidth}
        \centerline{\includegraphics[width=\columnwidth]{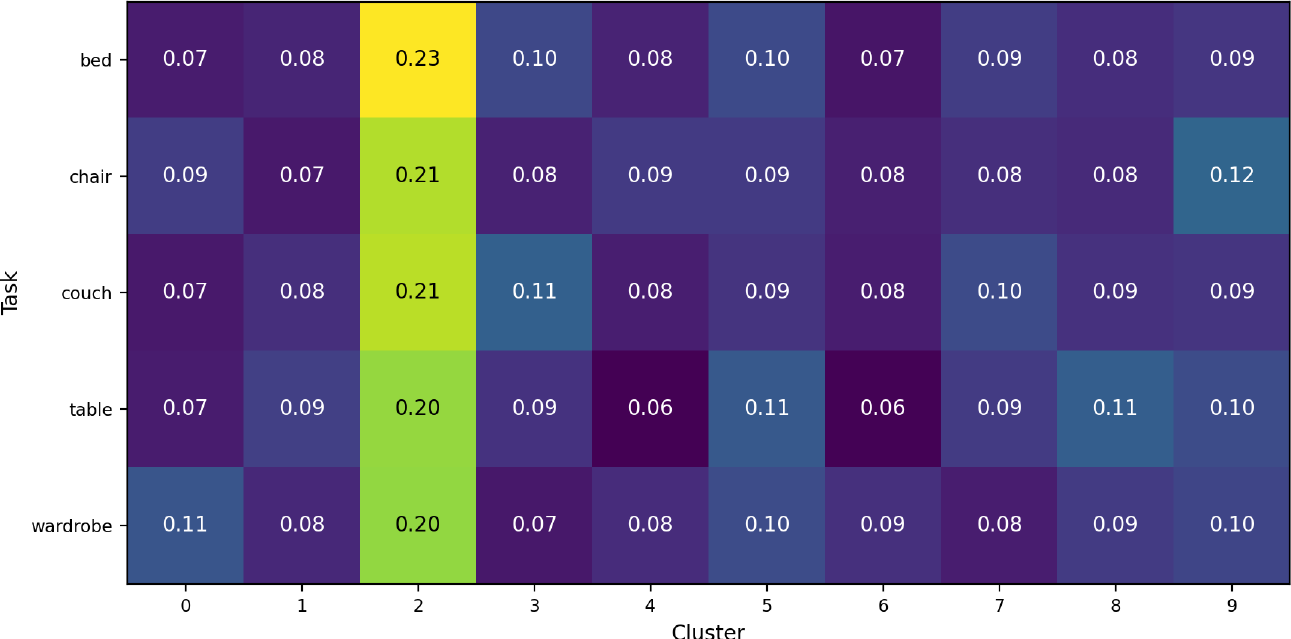}}
        \subcaption{household furniture}\label{fig_G3_10C_household_furniture}
    \end{minipage}\
    \begin{minipage}[c]{0.32\columnwidth}
        \centerline{\includegraphics[width=\columnwidth]{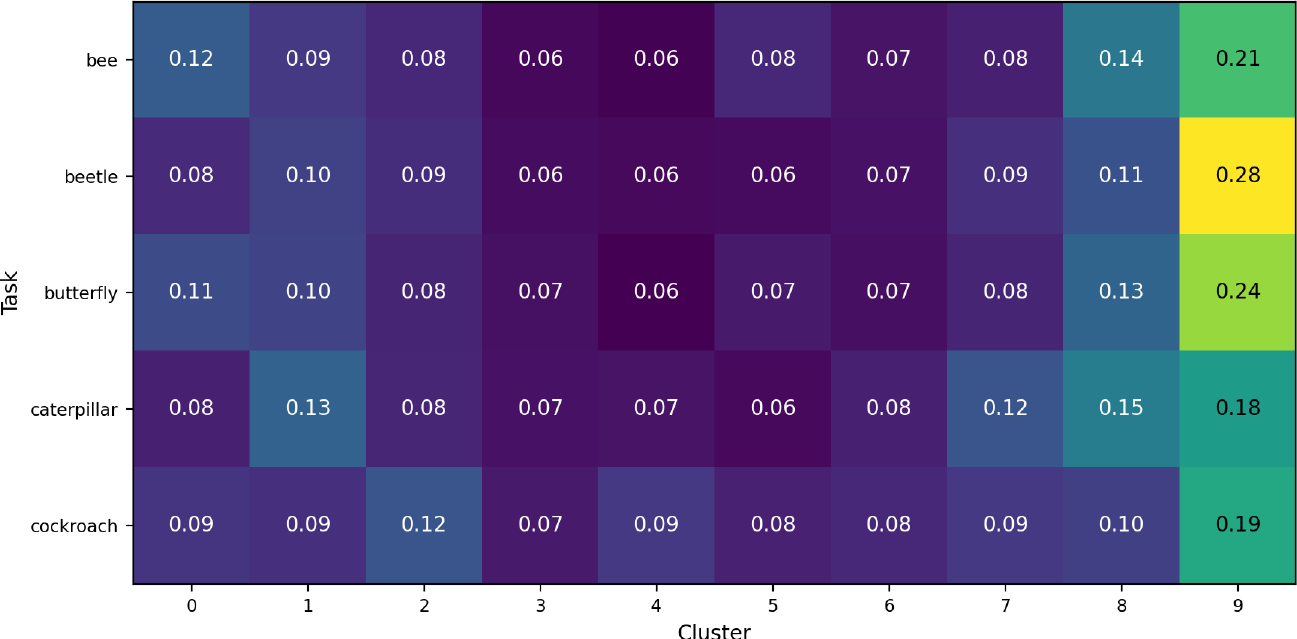}}
        \subcaption{insects}\label{fig_G3_10C_insects}
    \end{minipage}\
    \begin{minipage}[c]{0.32\columnwidth}
        \centerline{\includegraphics[width=\columnwidth]{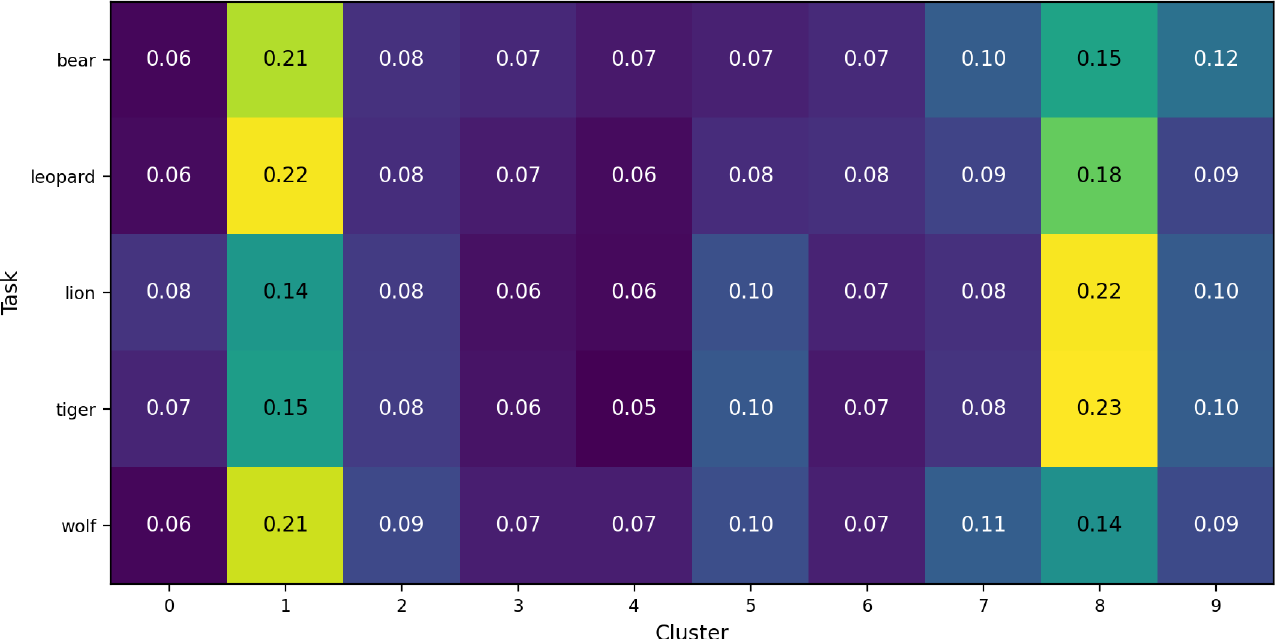}}
        \subcaption{large carnivores}\label{fig_G3_10C_large_carnivores}
    \end{minipage}\\
    \begin{minipage}[c]{0.32\columnwidth}
        \centerline{\includegraphics[width=\columnwidth]{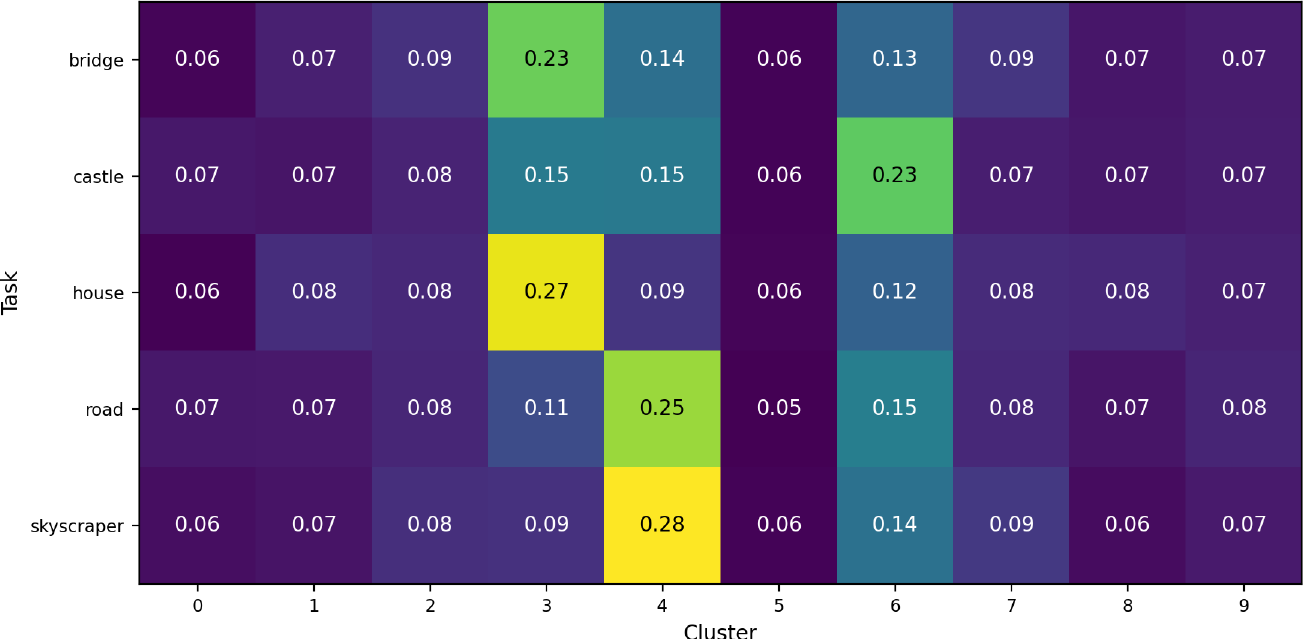}}
        \subcaption{large man-made outdoor things}\label{fig_G3_10C_large_man_made_outdoor_things}
    \end{minipage}\
    \begin{minipage}[c]{0.32\columnwidth}
        \centerline{\includegraphics[width=\columnwidth]{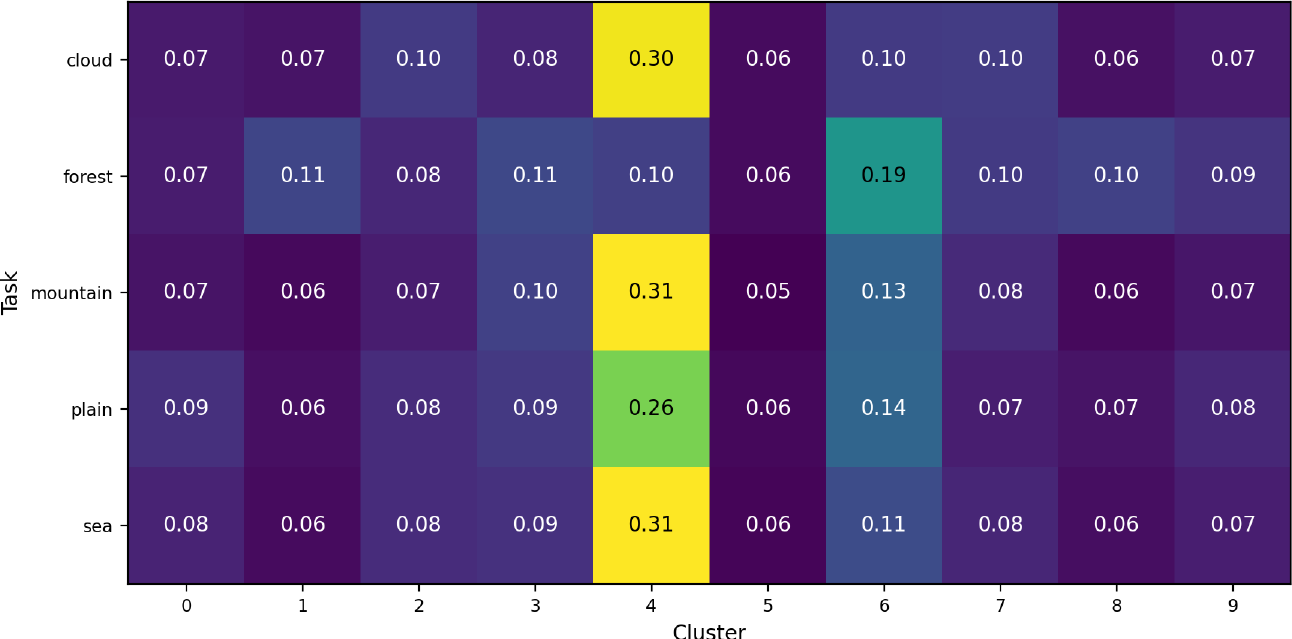}}
        \subcaption{large natural outdoor scenes}\label{fig_G3_10C_large_natural_outdoor_scenes}
    \end{minipage}\
    \begin{minipage}[c]{0.32\columnwidth}
        \centerline{\includegraphics[width=\columnwidth]{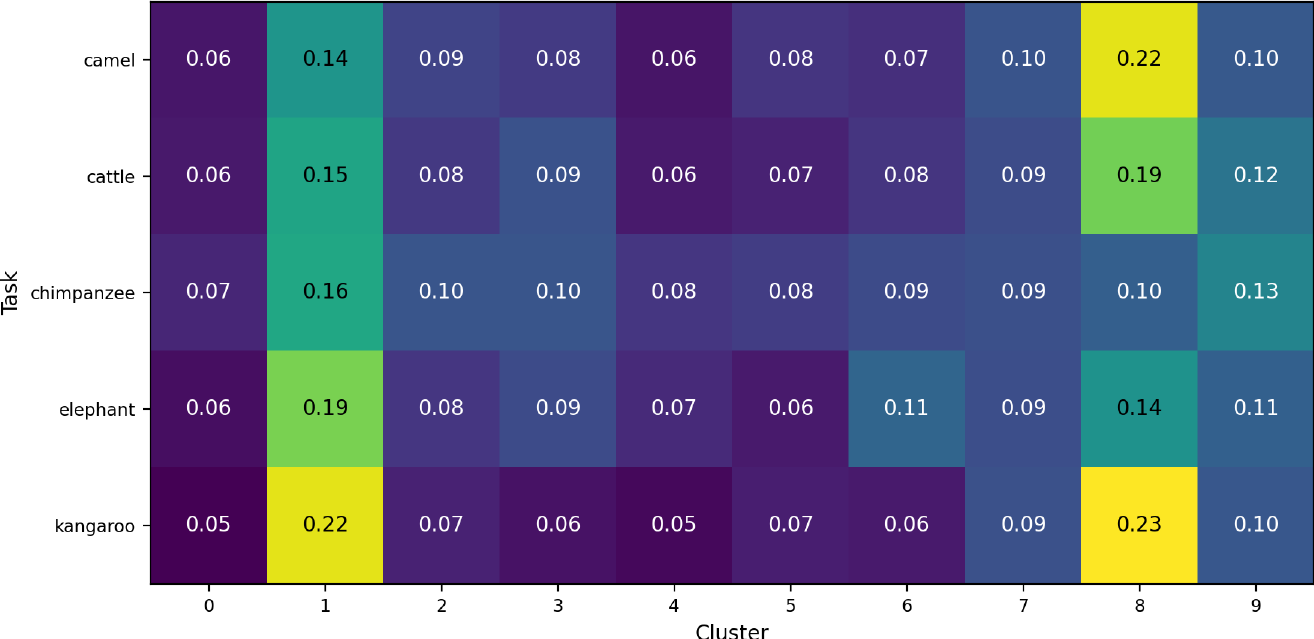}}
        \subcaption{large omnivores and herbivores}\label{fig_G3_10C_large_omnivores_and_herbivores}
    \end{minipage}\\
    \begin{minipage}[c]{0.32\columnwidth}
        \centerline{\includegraphics[width=\columnwidth]{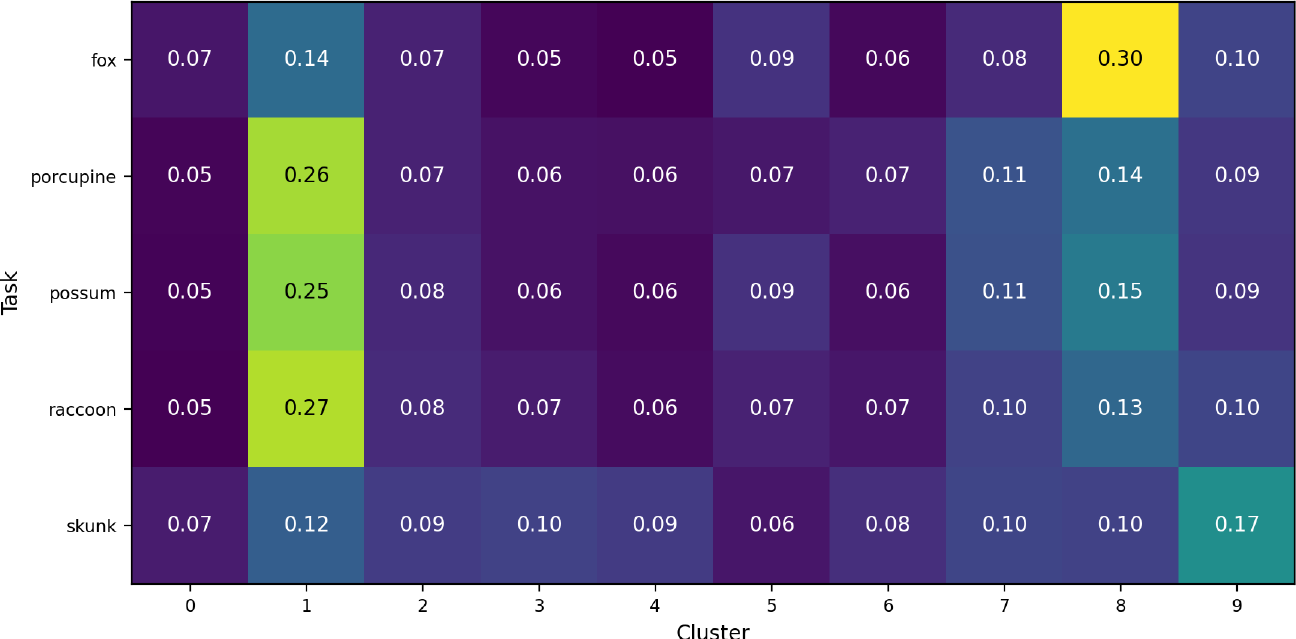}}
        \subcaption{medium-sized mammals}\label{fig_G3_10C_medium_sized_mammals}
    \end{minipage}\
    \begin{minipage}[c]{0.32\columnwidth}
        \centerline{\includegraphics[width=\columnwidth]{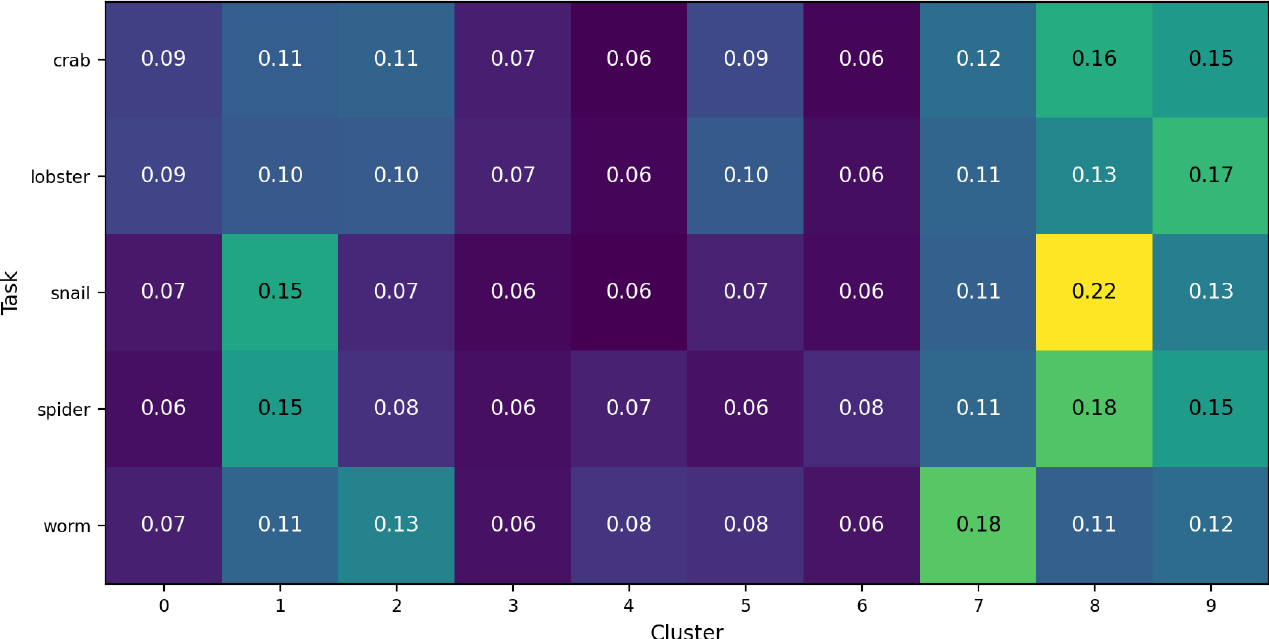}}
        \subcaption{non-insect invertebrates}\label{fig_G3_10C_non_insect_invertebrates}
    \end{minipage}\
    \begin{minipage}[c]{0.32\columnwidth}
        \centerline{\includegraphics[width=\columnwidth]{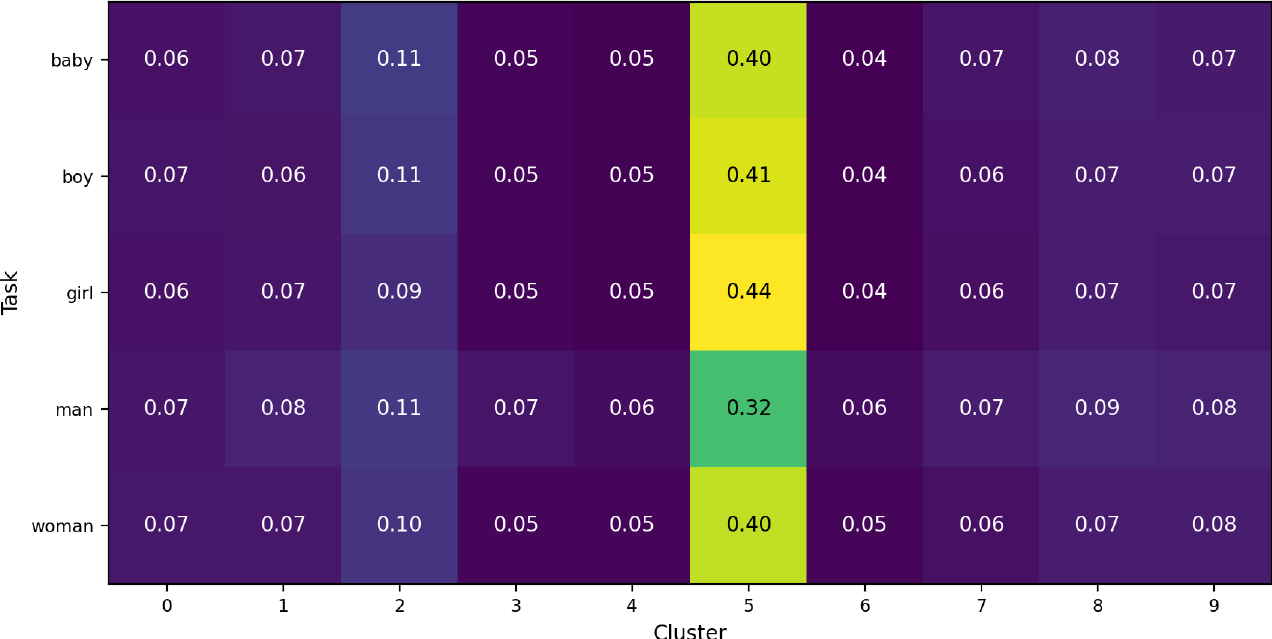}}
        \subcaption{people}\label{fig_G3_10C_people}
    \end{minipage}\\
    \begin{minipage}[c]{0.32\columnwidth}
        \centerline{\includegraphics[width=\columnwidth]{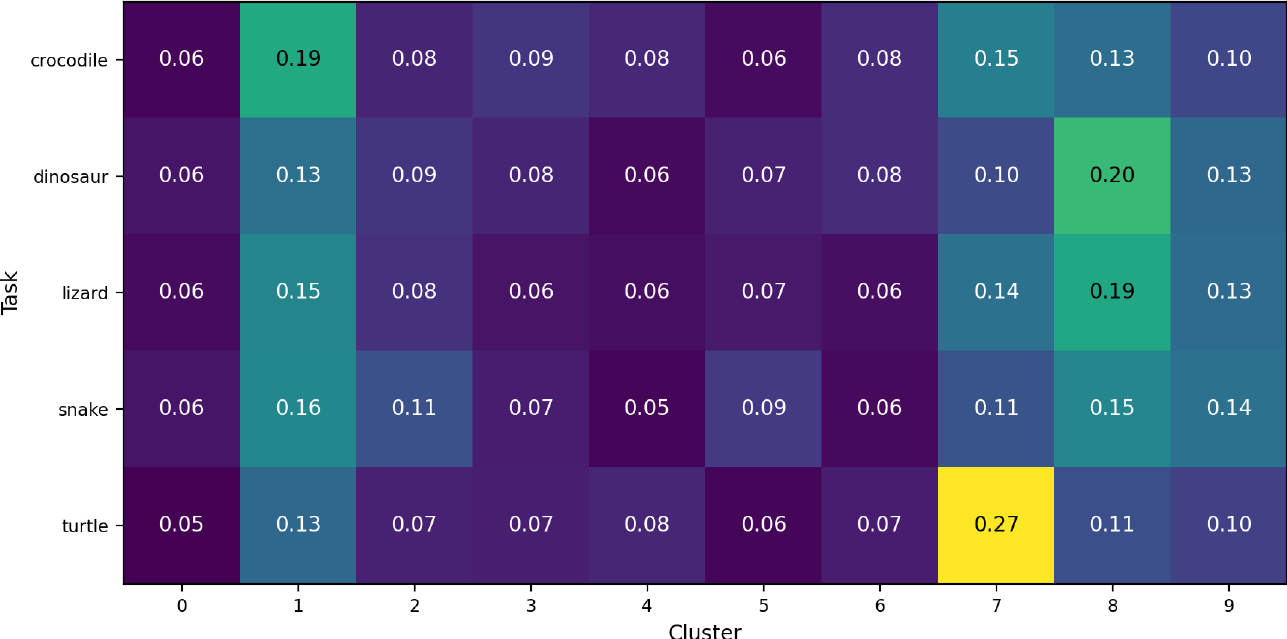}}
        \subcaption{reptiles}\label{fig_G3_10C_reptiles}
    \end{minipage}\
    \begin{minipage}[c]{0.32\columnwidth}
        \centerline{\includegraphics[width=\columnwidth]{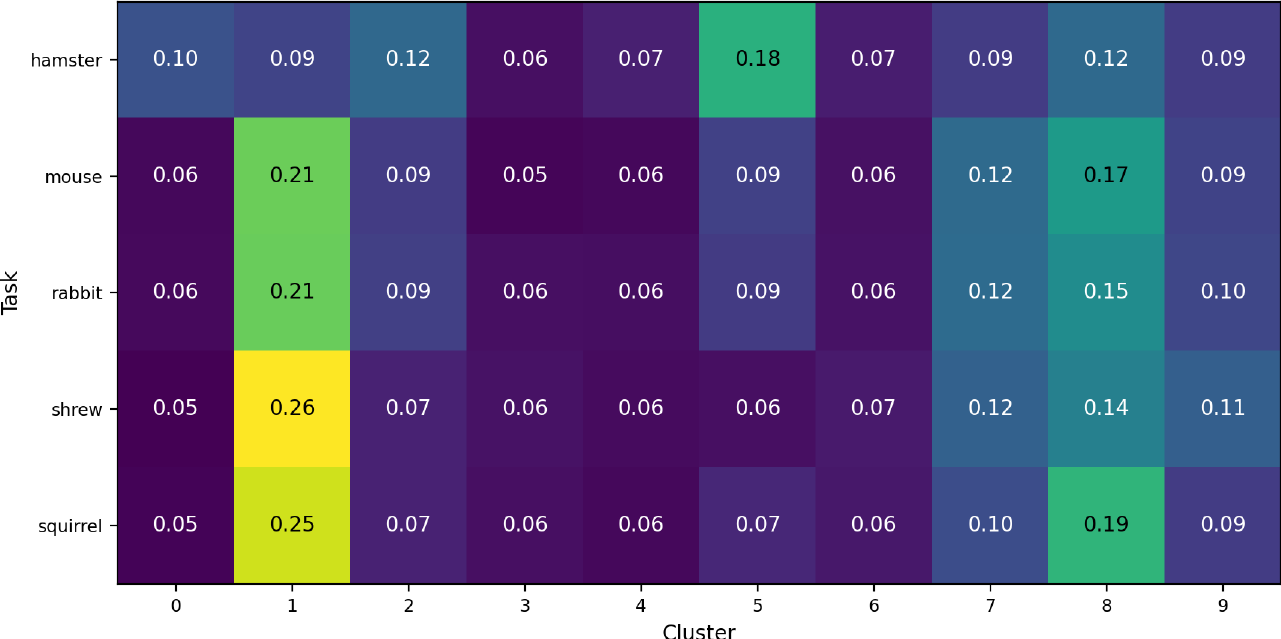}}
        \subcaption{small mammals}\label{fig_G3_10C_small_mammals}
    \end{minipage}\
    \begin{minipage}[c]{0.32\columnwidth}
        \centerline{\includegraphics[width=\columnwidth]{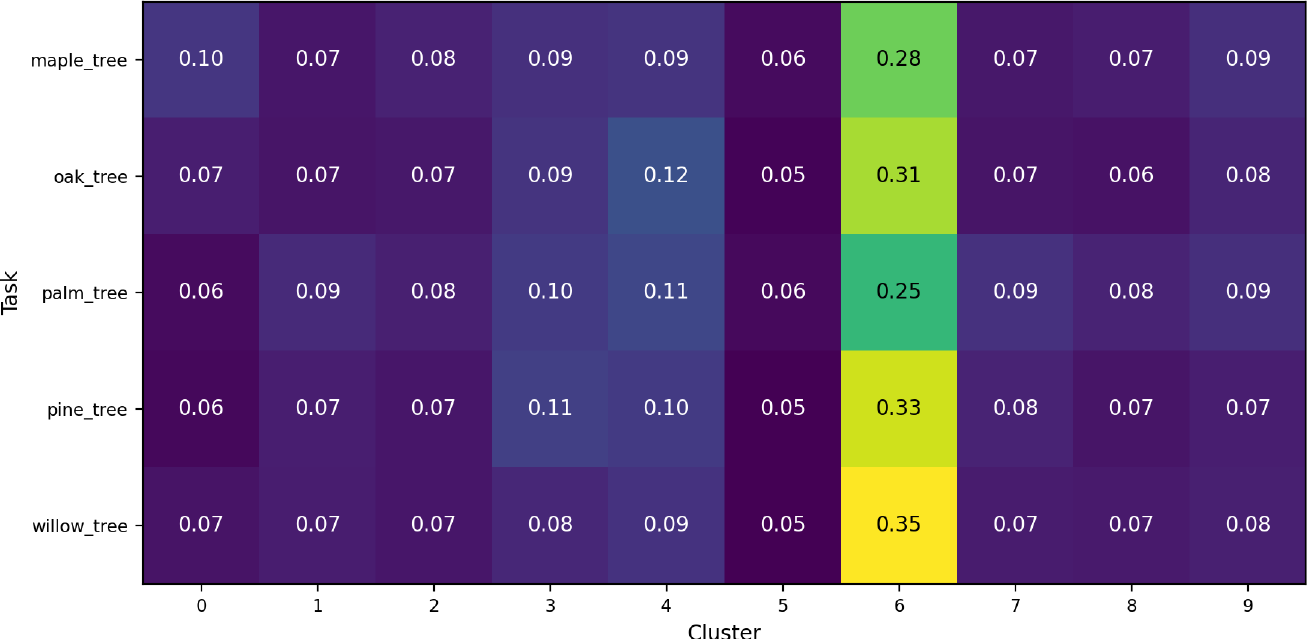}}
        \subcaption{trees}\label{fig_G3_10C_trees}
    \end{minipage}\\
    \begin{minipage}[c]{0.32\columnwidth}
        \centerline{\includegraphics[width=\columnwidth]{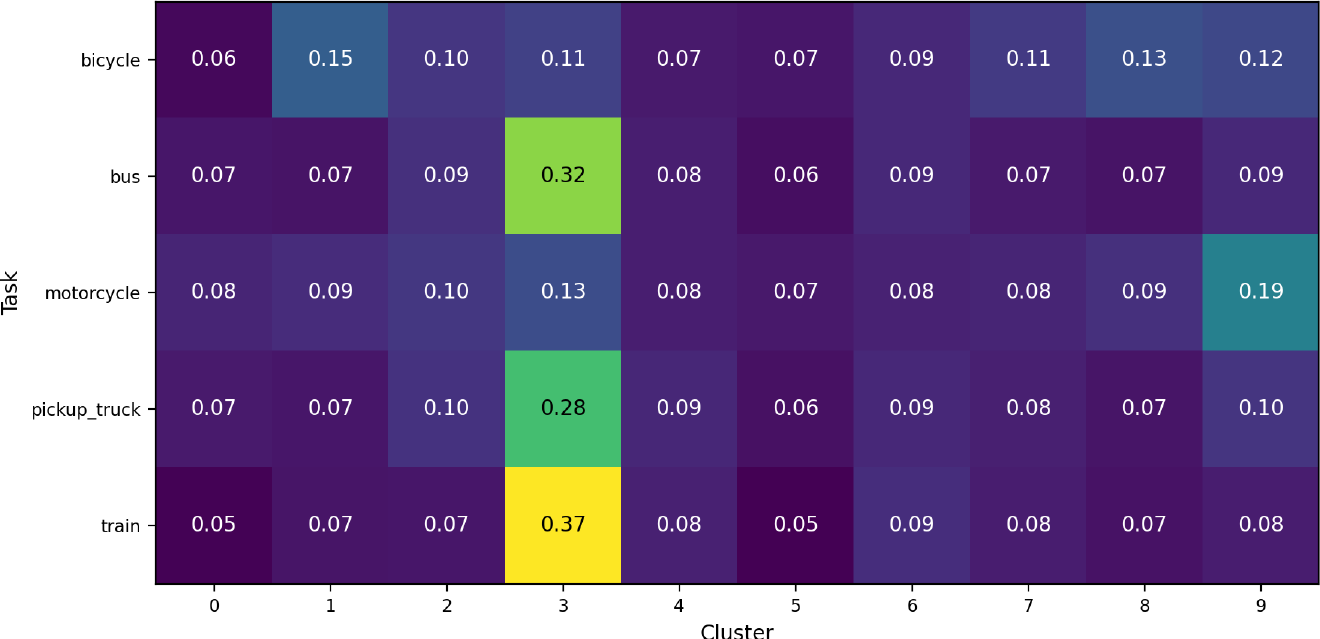}}
        \subcaption{vehicles 1}\label{fig_G3_10C_vehicles_1}
    \end{minipage}\
    \begin{minipage}[c]{0.32\columnwidth}
        \centerline{\includegraphics[width=\columnwidth]{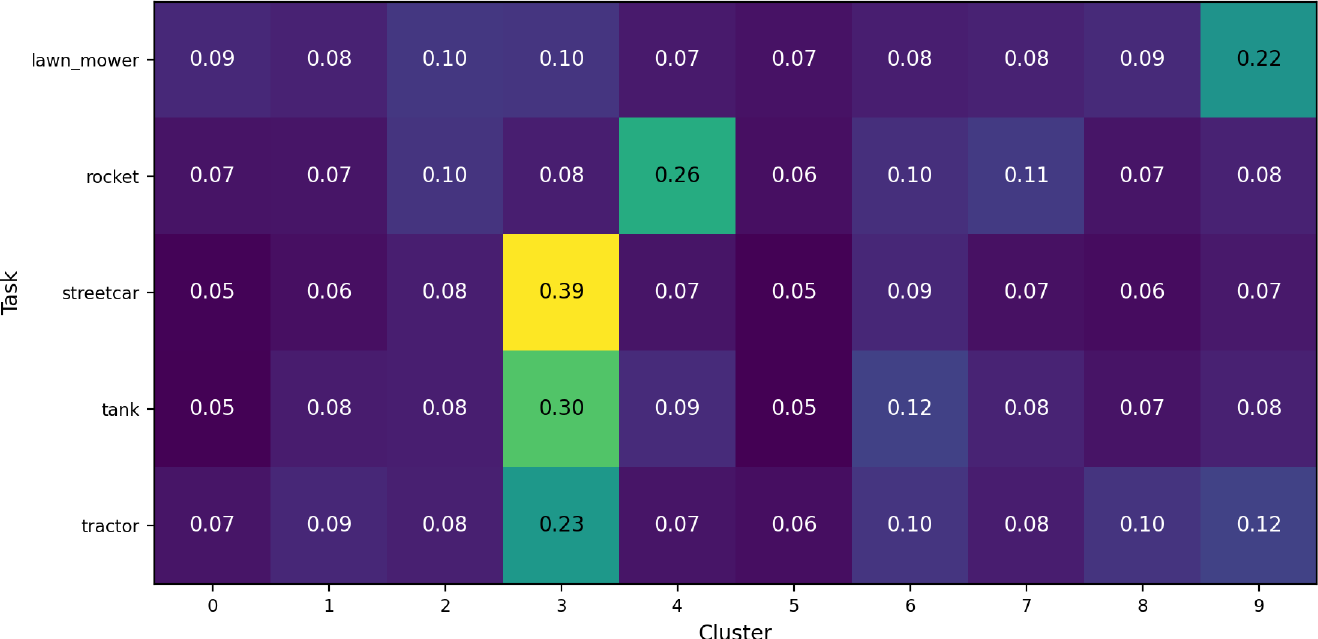}}
        \subcaption{vehicles 2}\label{fig_G3_10C_vehicles_2}
    \end{minipage}\\
    \caption{Task grouping of G3 (100 Tasks of CIFAR100) into $10$ clusters with $F=2$ using Data Maps from RESNET18}\label{fig_G3_10C}
\end{figure*}

In this appendix, we present detailed insights into the task clustering results of 100 tasks, building upon the experimental setup outlined in Section \ref{section_clustering_results}. Figure \ref{fig_G3_20C} showcases the clustering results using $20$ clusters, while Figure \ref{fig_G3_10C} illustrates the results with $10$ clusters. Each image in the figures represents the clustering outcomes for one superclass from CIFAR100.

Notably, the clustering with $20$ clusters demonstrates successful grouping in many categories, such as $\{$People, Trees, Food Containers, Flowers, Household Electrical Devices$\}$. Figures \ref{fig_G3_20C_vehicles_1} and \ref{fig_G3_20C_vehicles_2} highlight the close association between Vehicles 1 and 2, as they are almost merged into the same cluster (Cluster 15). When reducing the number of clusters to $10$, we observe enhanced coherence in the assigned tasks. For instance, the tasks related to ``Fruit and Vegetables'' are nearly clustered together after reducing the number of clusters.

Moreover, our method effectively captures the logical associations between categories. Figures \ref{fig_G3_10C_household_electrical_devices} and \ref{fig_G3_10C_household_furniture} showcase the plausible merge between ``Household Electrical Devices'' and ``Household Furniture.'' Additionally, we observe similar distribution patterns among related categories, such as $\{$ Large Carnivores, Large Omnivores and Herbivores, Medium-Sized Mammals$\}$ in Figures \ref{fig_G3_10C_large_carnivores}, \ref{fig_G3_10C_large_omnivores_and_herbivores}, and \ref{fig_G3_10C_medium_sized_mammals} respectively.

Overall, these results reveal the qualitative success of our approach in clustering a very large number of tasks, highlighting the effectiveness of our method even in such challenging scenarios.

\begin{figure*}[ht]
    \centering
    \begin{minipage}[c]{0.32\columnwidth}
        \centerline{\includegraphics[width=\columnwidth]{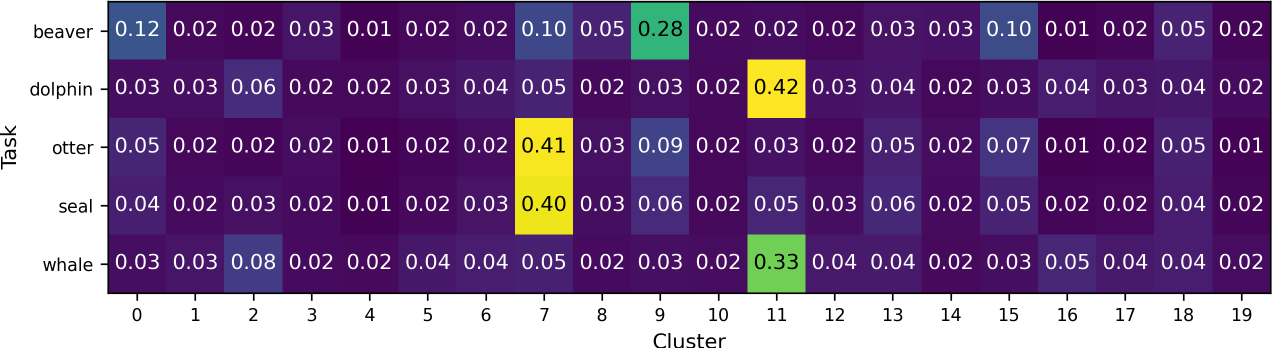}}
        \subcaption{aquatic mammals }\label{fig_G3_20C_custom_aquatic_mammals}
    \end{minipage}\
    \begin{minipage}[c]{0.32\columnwidth}
        \centerline{\includegraphics[width=\columnwidth]{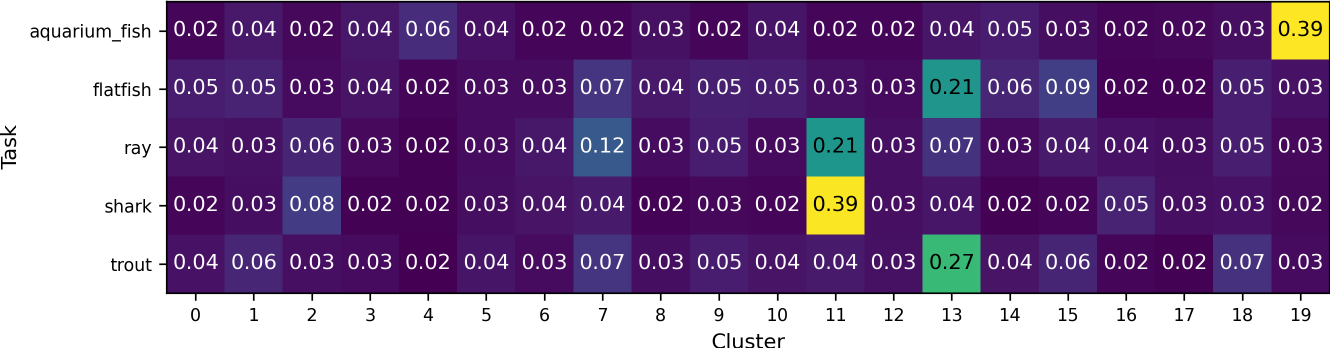}}
        \subcaption{fish}\label{fig_G3_20C_custom_fish}
    \end{minipage}\
    \begin{minipage}[c]{0.32\columnwidth}
        \centerline{\includegraphics[width=\columnwidth]{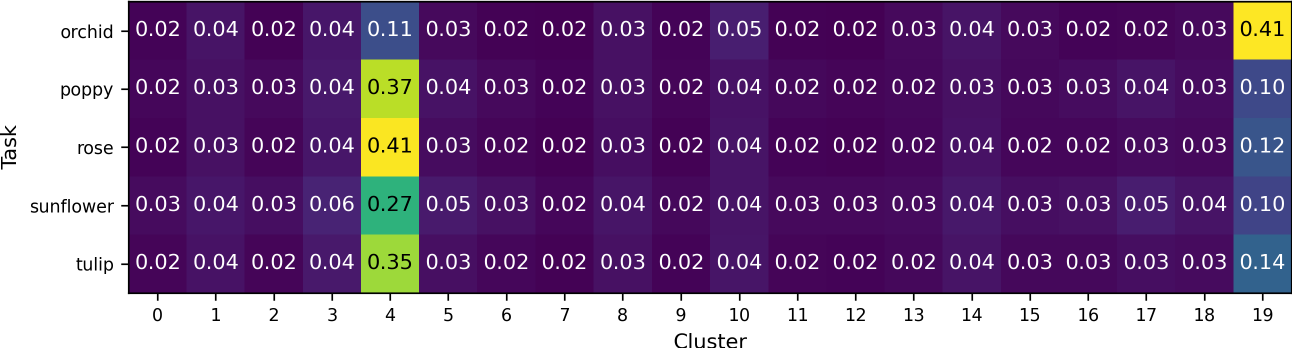}}
        \subcaption{flowers}\label{fig_G3_20C_custom_flowers}
    \end{minipage}\\
    \begin{minipage}[c]{0.32\columnwidth}
        \centerline{\includegraphics[width=\columnwidth]{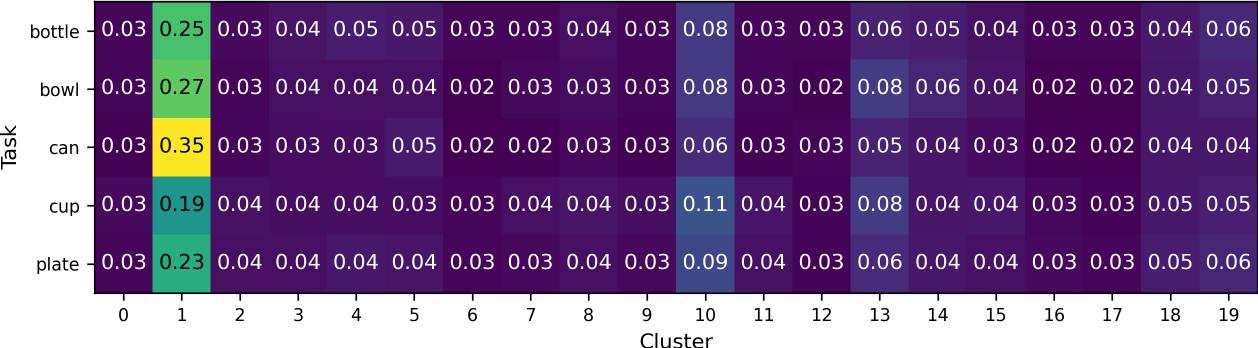}}
        \subcaption{food containers}\label{fig_G3_20C_custom_food_containers}
    \end{minipage}\
    \begin{minipage}[c]{0.32\columnwidth}
        \centerline{\includegraphics[width=\columnwidth]{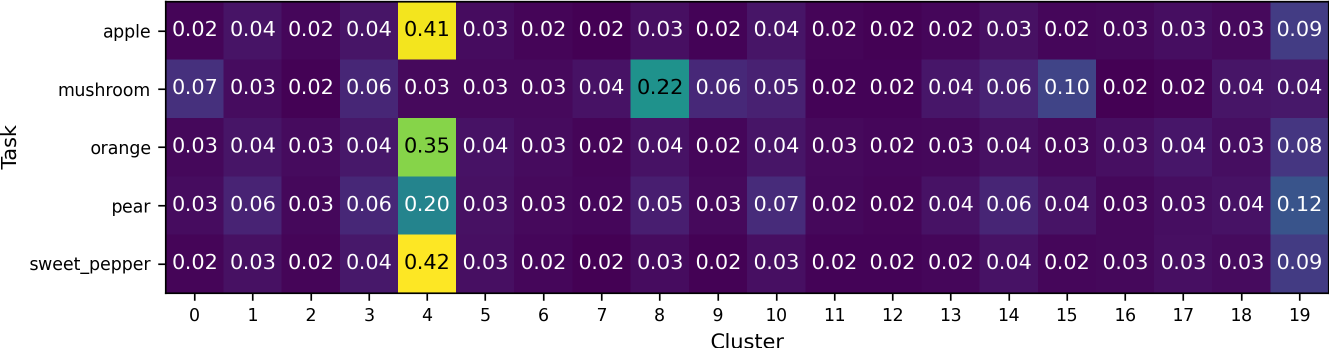}}
        \subcaption{fruit and vegetables}\label{fig_G3_20C_custom_fruit_and_vegetables}
    \end{minipage}\
    \begin{minipage}[c]{0.32\columnwidth}
        \centerline{\includegraphics[width=\columnwidth]{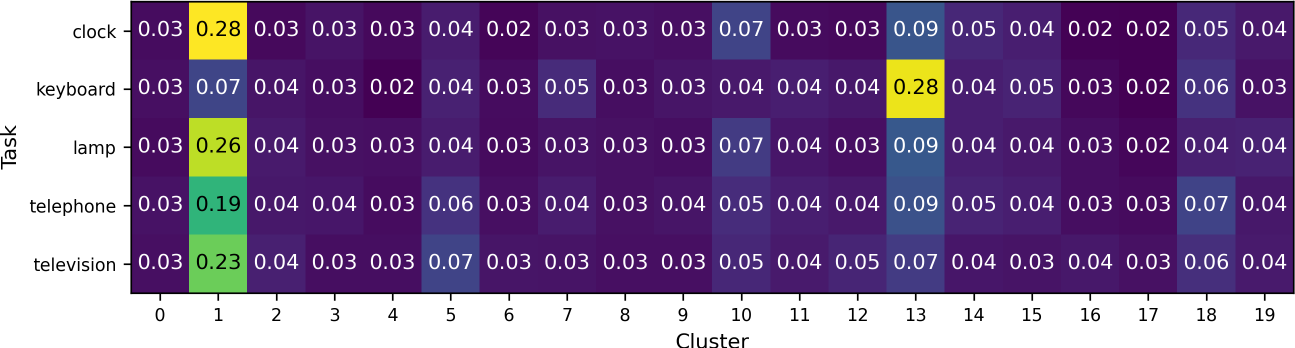}}
        \subcaption{household electrical devices}\label{fig_G3_20C_custom_household_electrical_devices}
    \end{minipage}\\
    \begin{minipage}[c]{0.32\columnwidth}
        \centerline{\includegraphics[width=\columnwidth]{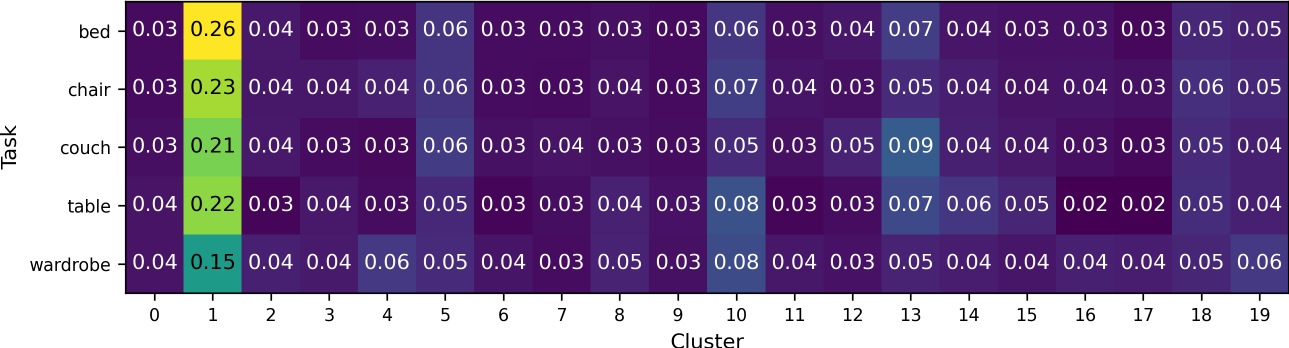}}
        \subcaption{household furniture}\label{fig_G3_20C_custom_household_furniture}
    \end{minipage}\
    \begin{minipage}[c]{0.32\columnwidth}
        \centerline{\includegraphics[width=\columnwidth]{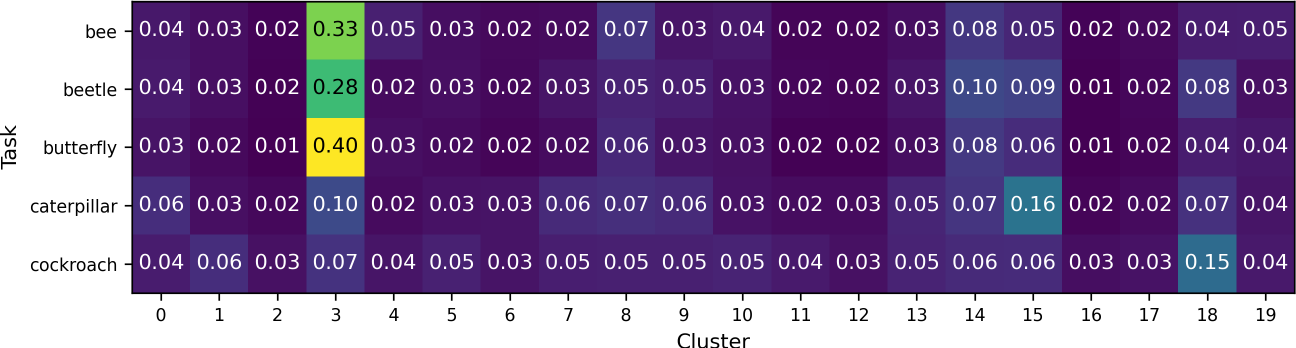}}
        \subcaption{insects}\label{fig_G3_20C_custom_insects}
    \end{minipage}\
    \begin{minipage}[c]{0.32\columnwidth}
        \centerline{\includegraphics[width=\columnwidth]{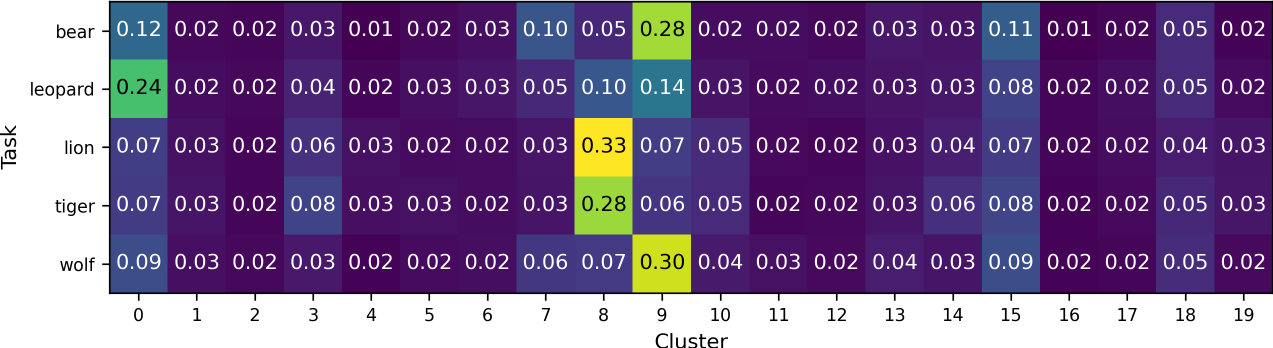}}
        \subcaption{large carnivores}\label{fig_G3_20C_custom_large_carnivores}
    \end{minipage}\\
    \begin{minipage}[c]{0.32\columnwidth}
        \centerline{\includegraphics[width=\columnwidth]{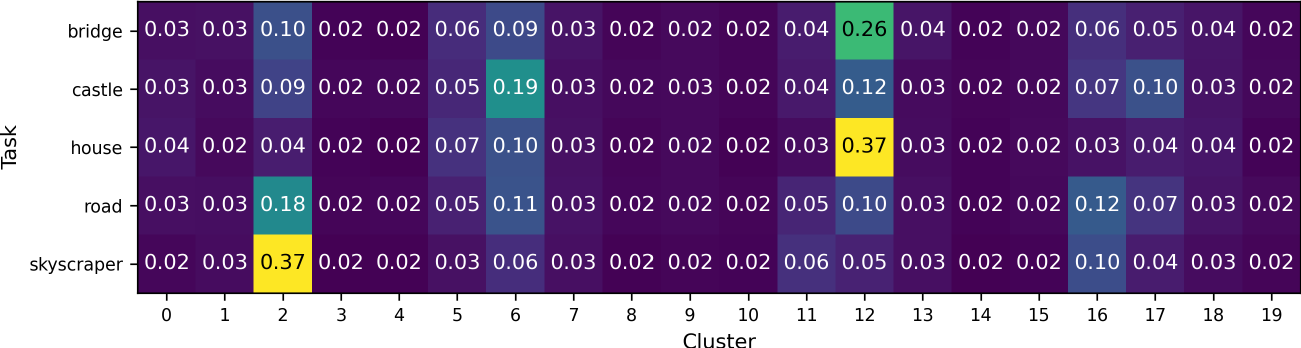}}
        \subcaption{large man-made outdoor things}\label{fig_G3_20C_custom_large_man_made_outdoor_things}
    \end{minipage}\
    \begin{minipage}[c]{0.32\columnwidth}
        \centerline{\includegraphics[width=\columnwidth]{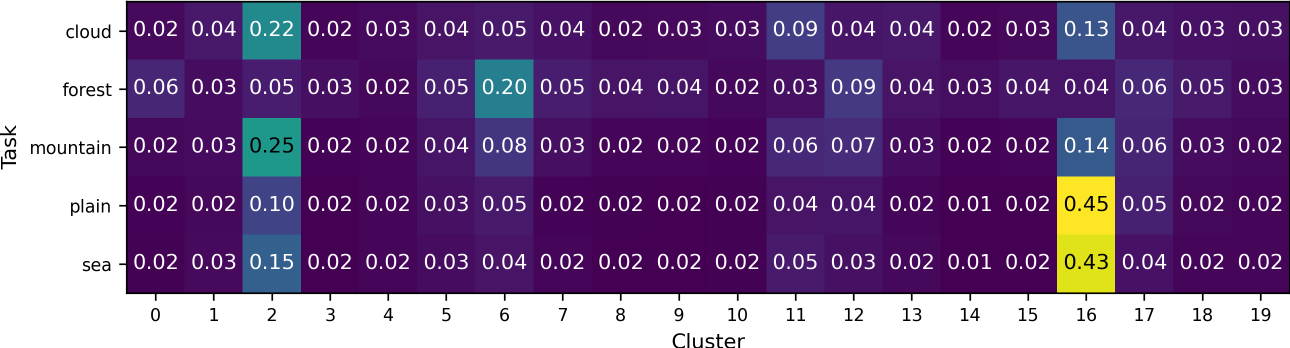}}
        \subcaption{large natural outdoor scenes}\label{fig_G3_20C_custom_large_natural_outdoor_scenes}
    \end{minipage}\
    \begin{minipage}[c]{0.32\columnwidth}
        \centerline{\includegraphics[width=\columnwidth]{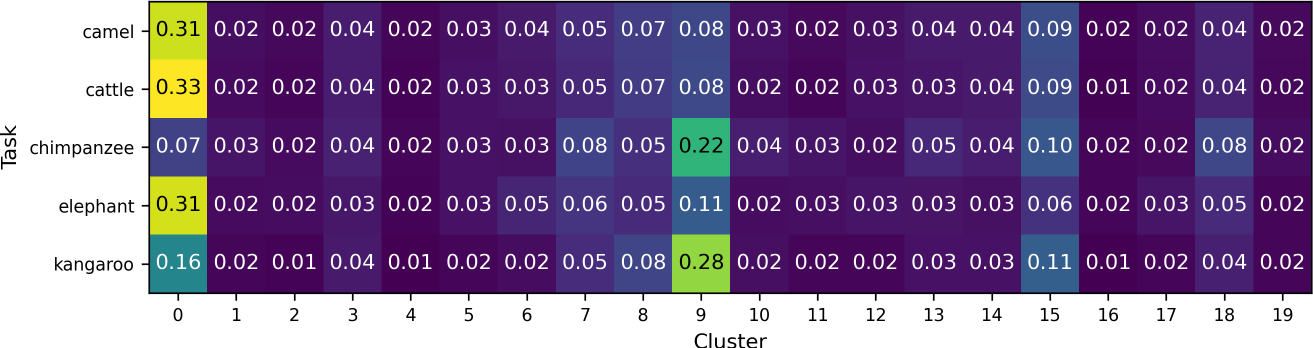}}
        \subcaption{large omnivores and herbivores}\label{fig_G3_20C_custom_large_omnivores_and_herbivores}
    \end{minipage}\\
    \begin{minipage}[c]{0.32\columnwidth}
        \centerline{\includegraphics[width=\columnwidth]{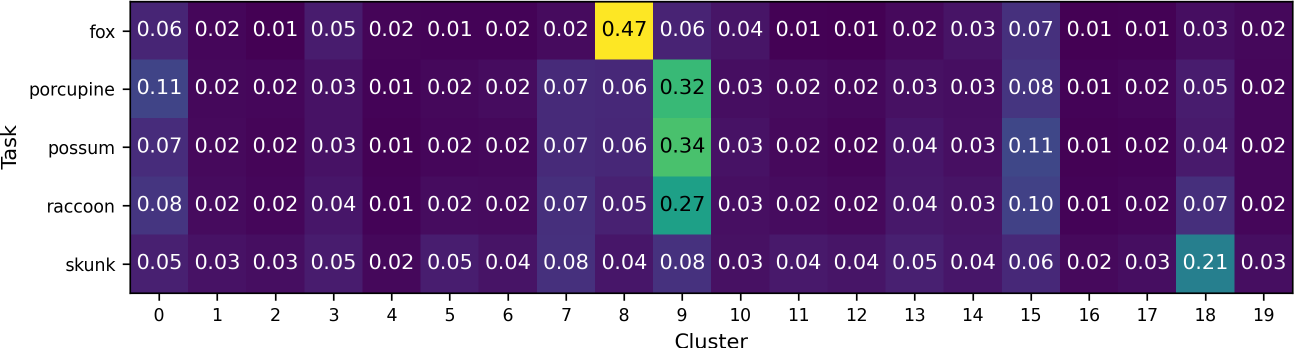}}
        \subcaption{medium-sized mammals}\label{fig_G3_20C_custom_medium_sized_mammals}
    \end{minipage}\
    \begin{minipage}[c]{0.32\columnwidth}
        \centerline{\includegraphics[width=\columnwidth]{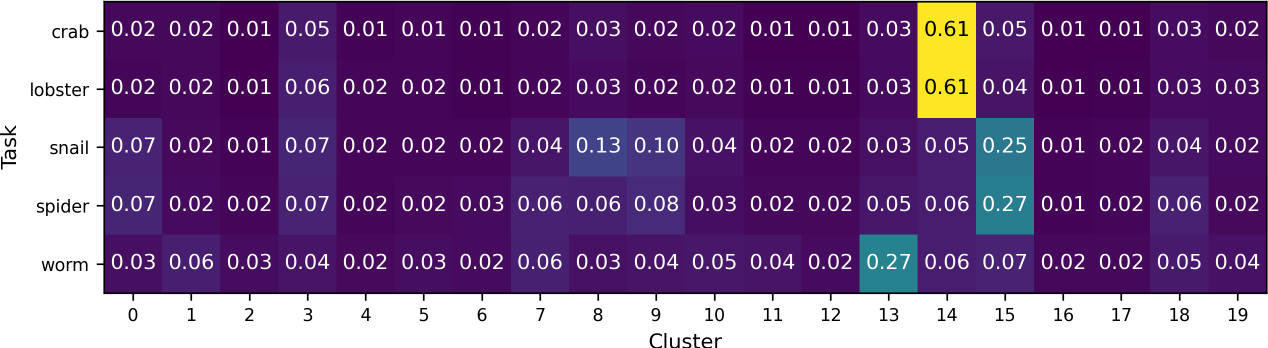}}
        \subcaption{non-insect invertebrates}\label{fig_G3_20C_custom_non_insect_invertebrates}
    \end{minipage}\
    \begin{minipage}[c]{0.32\columnwidth}
        \centerline{\includegraphics[width=\columnwidth]{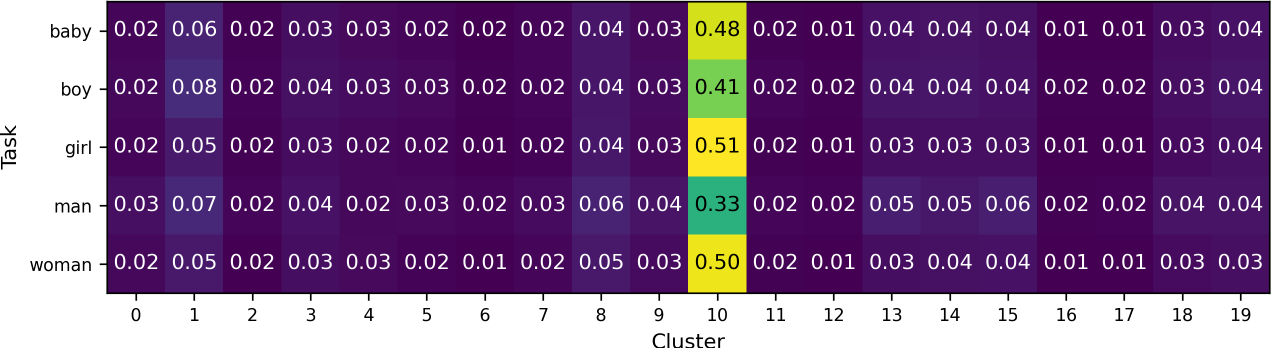}}
        \subcaption{people}\label{fig_G3_20C_custom_people}
    \end{minipage}\\
    \begin{minipage}[c]{0.32\columnwidth}
        \centerline{\includegraphics[width=\columnwidth]{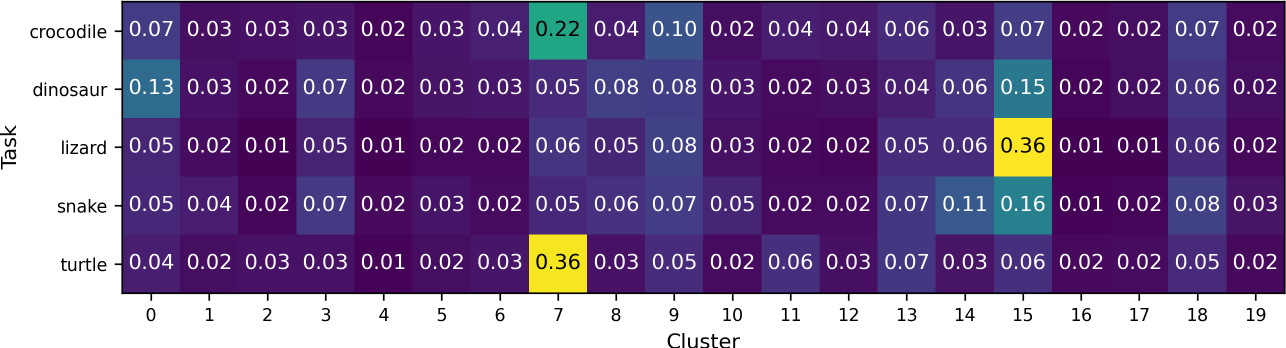}}
        \subcaption{reptiles}\label{fig_G3_20C_custom_reptiles}
    \end{minipage}\
    \begin{minipage}[c]{0.32\columnwidth}
        \centerline{\includegraphics[width=\columnwidth]{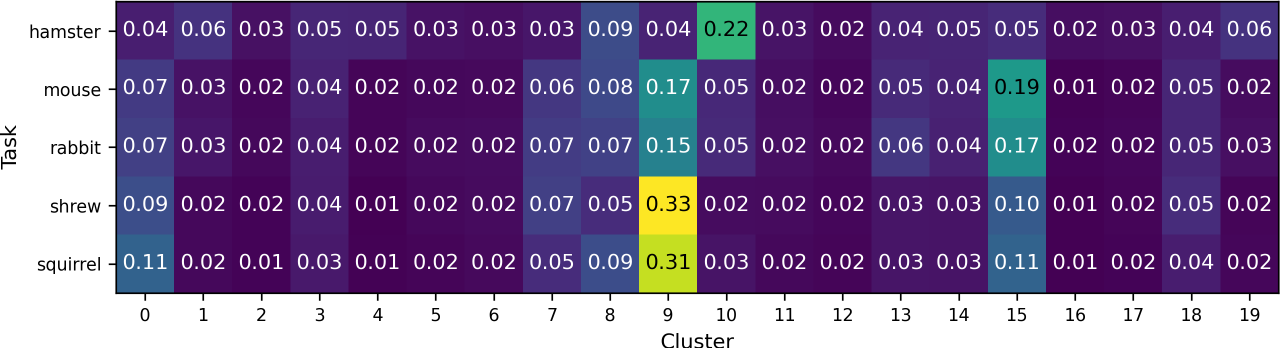}}
        \subcaption{small mammals}\label{fig_G3_20C_custom_small_mammals}
    \end{minipage}\
    \begin{minipage}[c]{0.32\columnwidth}
        \centerline{\includegraphics[width=\columnwidth]{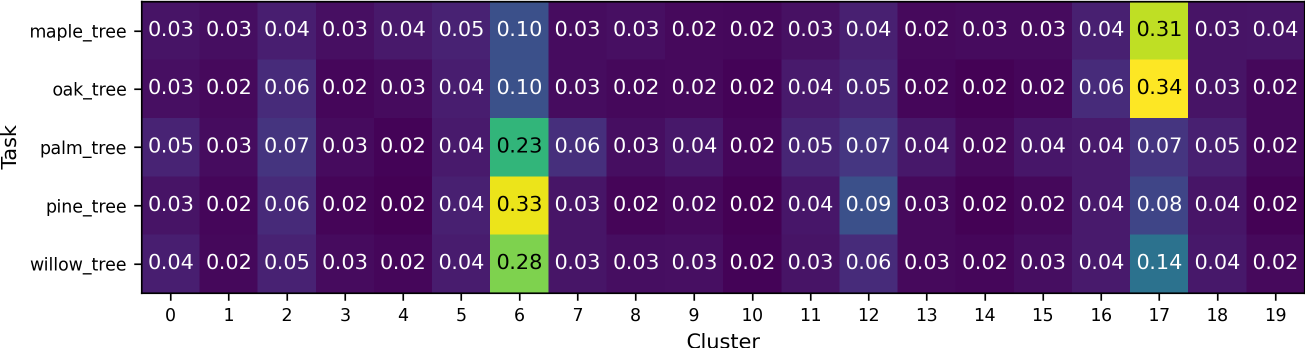}}
        \subcaption{trees}\label{fig_G3_20C_custom_trees}
    \end{minipage}\\
    \begin{minipage}[c]{0.32\columnwidth}
        \centerline{\includegraphics[width=\columnwidth]{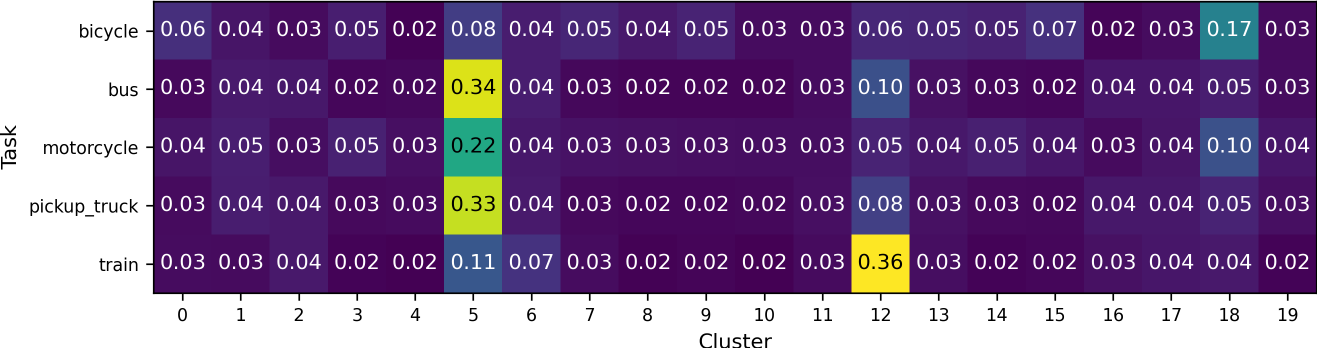}}
        \subcaption{vehicles 1}\label{fig_G3_20C_custom_vehicles_1}
    \end{minipage}\
    \begin{minipage}[c]{0.32\columnwidth}
        \centerline{\includegraphics[width=\columnwidth]{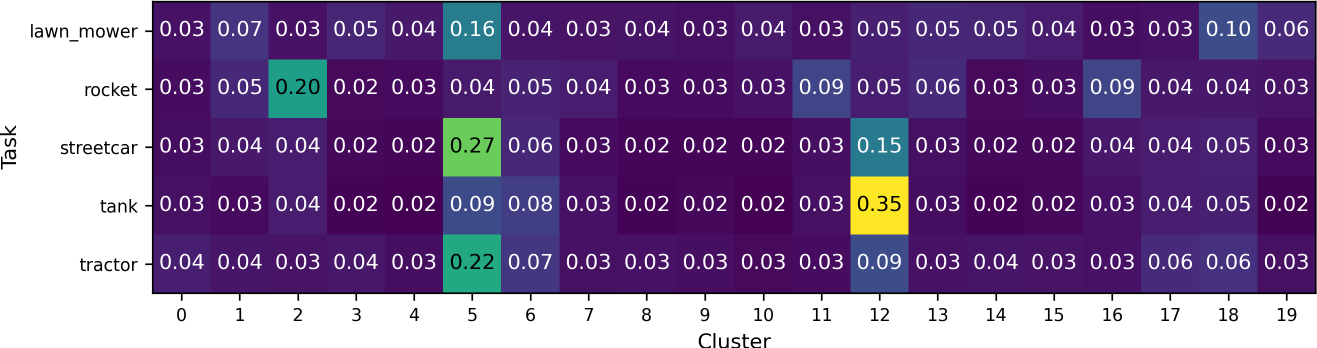}}
        \subcaption{vehicles 2}\label{fig_G3_20C_custom_vehicles_2}
    \end{minipage}\\
    \caption{Task grouping of G3 (100 Tasks of CIFAR100) into $20$ clusters with $F=2$ using Data Maps from the Custom CNN}\label{fig_G3_20C_custom}
\end{figure*}

\subsection{Robust Clustering using Less Complex Architectures} \label{appendix_custom_cnn_exp}

\begin{wraptable}{r}{0.48\textwidth}
\centering
\caption{The custom CNN archtecture used in our experiments. $n$ is the number of tasks.}
\label{table_custom_model}
\begin{tabular}{lcr}
\toprule
Layer                & Kernel Size  & Output        \\
\midrule
Convolution          & $5 \times 5$ & $32$ channels \\
2D BN + ReLU &              &               \\
2D MaxPool           & $2 \times 2$ &               \\
Convolution          & $3 \times 3$ & $64$ channels \\
2D BN + ReLU &              &               \\
Convolution          & $3 \times 3$ & $64$ channels \\
2D BN + ReLU &              &               \\
2D MaxPool           & $2 \times 2$ &               \\
Flatten              &              &               \\
FC      &              & $128$         \\
FC      &              & $n$           \\
\bottomrule
\end{tabular}
\end{wraptable}
To assess the robustness and generalization potential of our approach, we conducted a similar experiment using a simpler CNN architecture. The architecture's details are provided in Table \ref{table_custom_model}, where $n=100$. Notably, the CNN architecture used for extracting the data maps in our clustering process comprises only $472,612$ learnable parameters. In contrast, the ResNet18 architecture, from which we initially derived our clustering results, consists of $11,227,712$ learnable parameters.

\begin{figure*}[ht!]
    \centering
    \begin{minipage}[c]{0.32\columnwidth}
        \centerline{\includegraphics[width=\columnwidth]{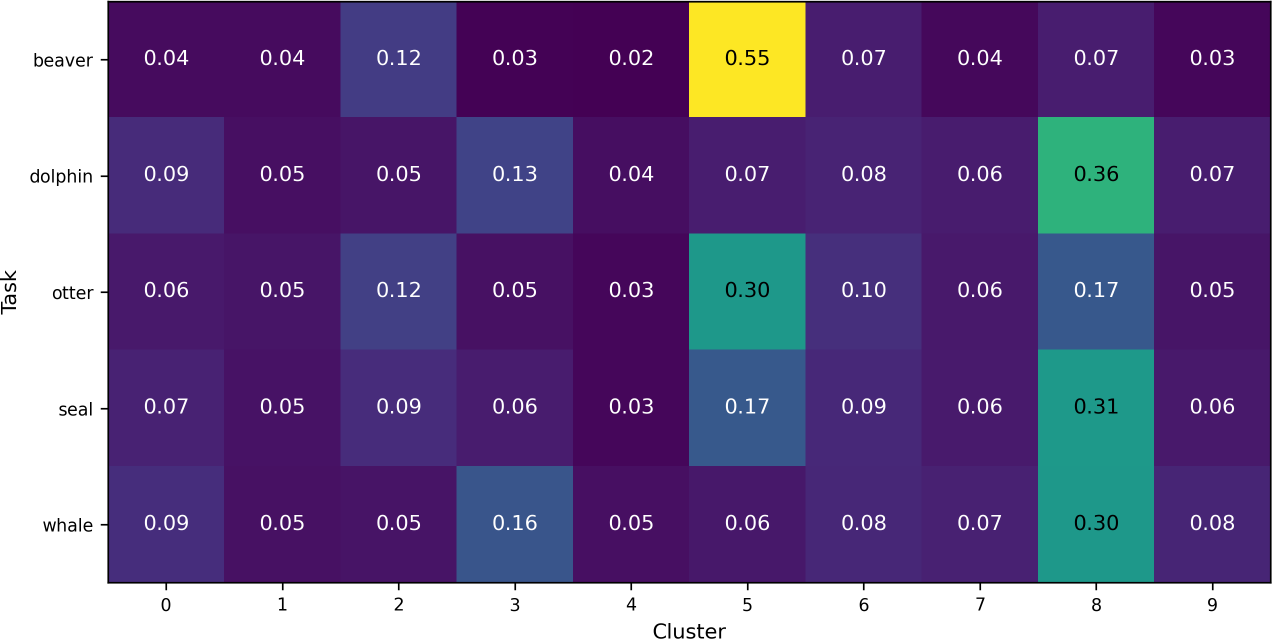}}
        \subcaption{aquatic mammals}\label{fig_G3_10C_custom_aquatic_mammals}
    \end{minipage}\
    \begin{minipage}[c]{0.32\columnwidth}
        \centerline{\includegraphics[width=\columnwidth]{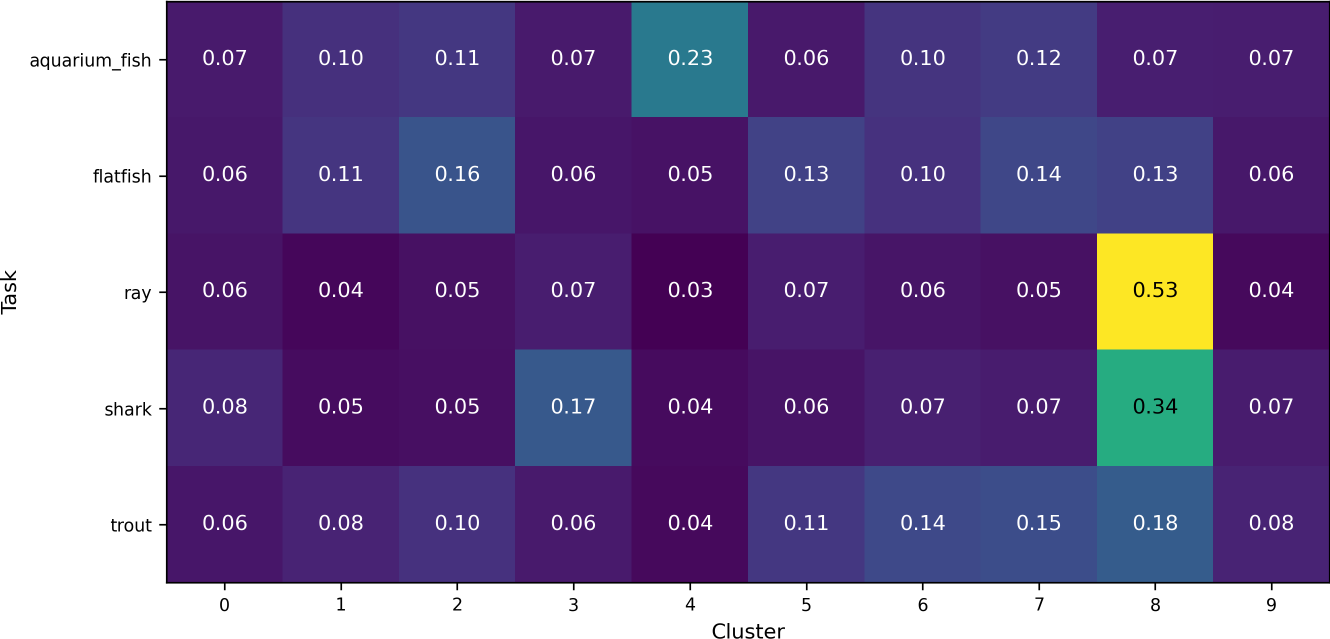}}
        \subcaption{fish}\label{fig_G3_10C_custom_fish}
    \end{minipage}\
    \begin{minipage}[c]{0.32\columnwidth}
        \centerline{\includegraphics[width=\columnwidth]{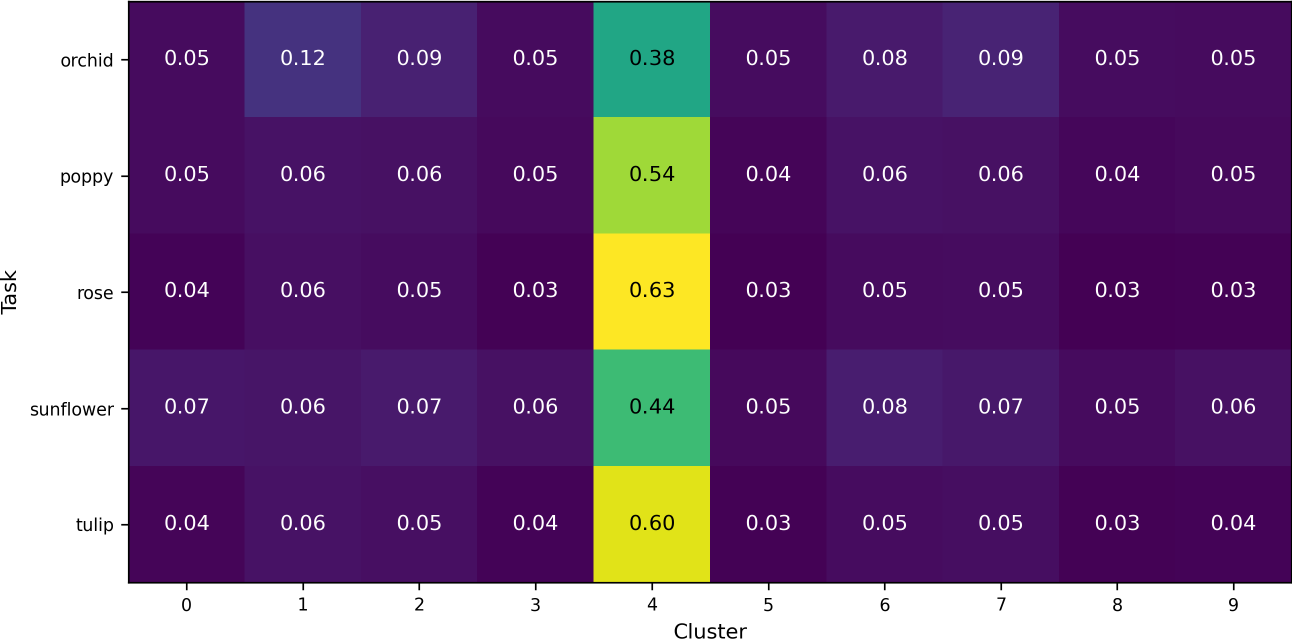}}
        \subcaption{flowers}\label{fig_G3_10C_custom_flowers}
    \end{minipage}\\
    \begin{minipage}[c]{0.32\columnwidth}
        \centerline{\includegraphics[width=\columnwidth]{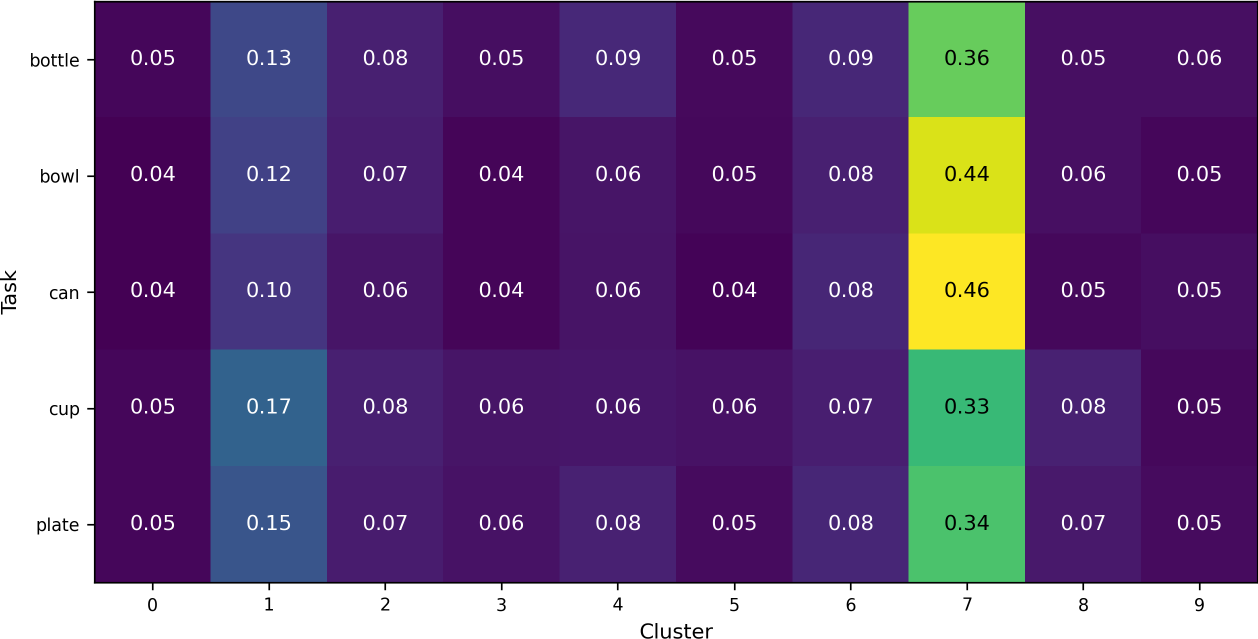}}
        \subcaption{food containers}\label{fig_G3_10C_custom_food_containers}
    \end{minipage}\
    \begin{minipage}[c]{0.32\columnwidth}
        \centerline{\includegraphics[width=\columnwidth]{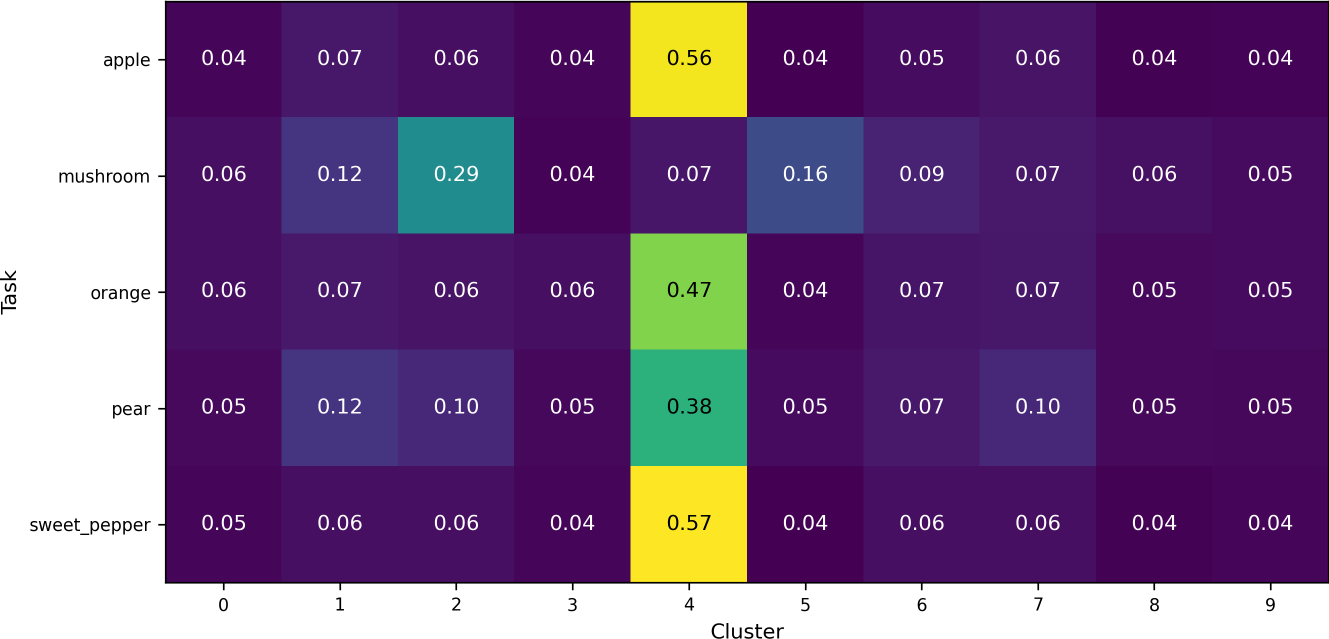}}
        \subcaption{fruit and vegetables}\label{fig_G3_10C_custom_fruit_and_vegetables}
    \end{minipage}\
    \begin{minipage}[c]{0.32\columnwidth}
        \centerline{\includegraphics[width=\columnwidth]{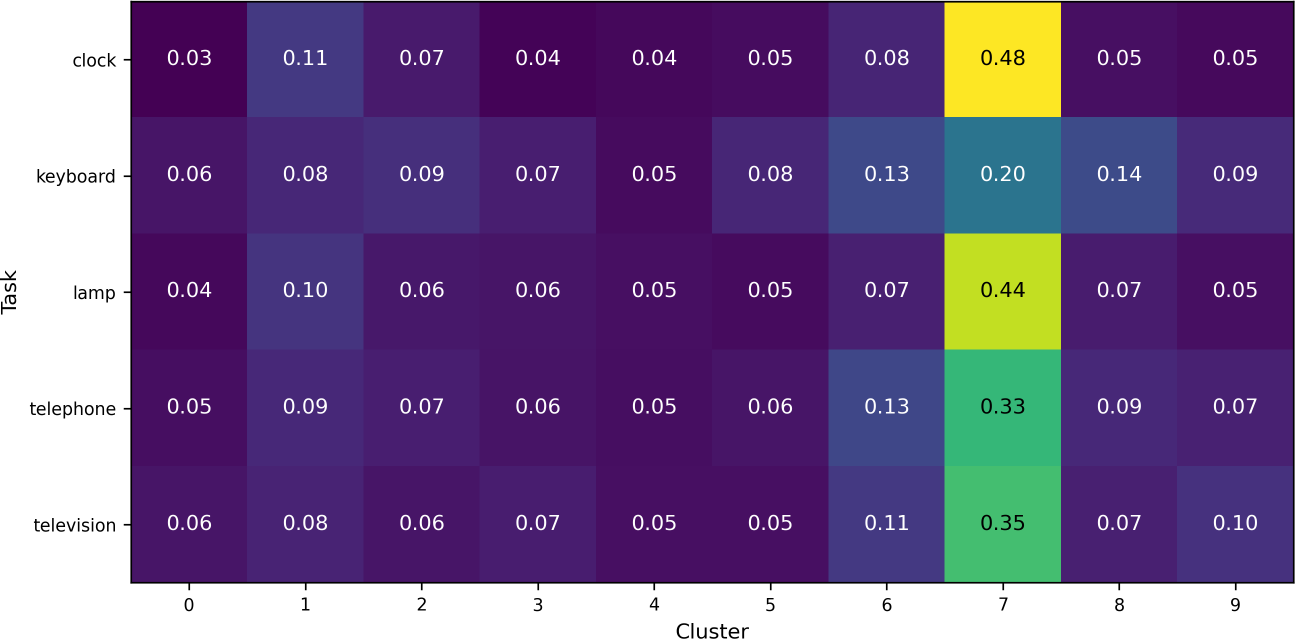}}
        \subcaption{household electrical devices}\label{fig_G3_10C_custom_household_electrical_devices}
    \end{minipage}\\
    \begin{minipage}[c]{0.32\columnwidth}
        \centerline{\includegraphics[width=\columnwidth]{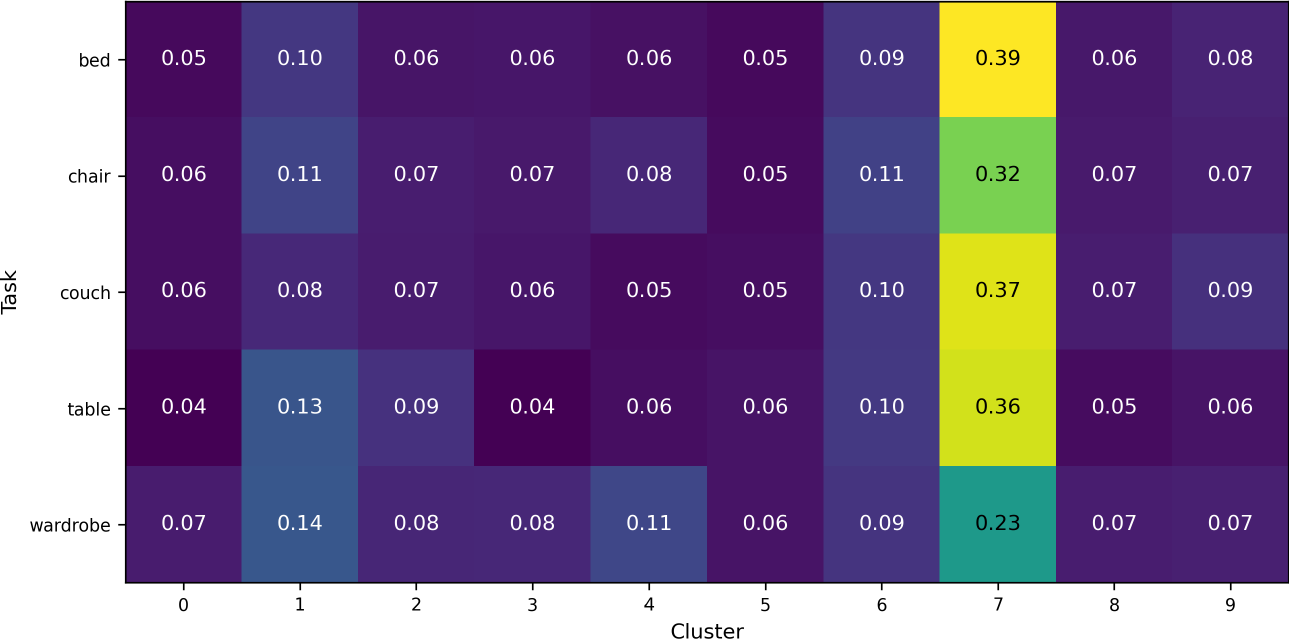}}
        \subcaption{household furniture}\label{fig_G3_10C_custom_household_furniture}
    \end{minipage}\
    \begin{minipage}[c]{0.32\columnwidth}
        \centerline{\includegraphics[width=\columnwidth]{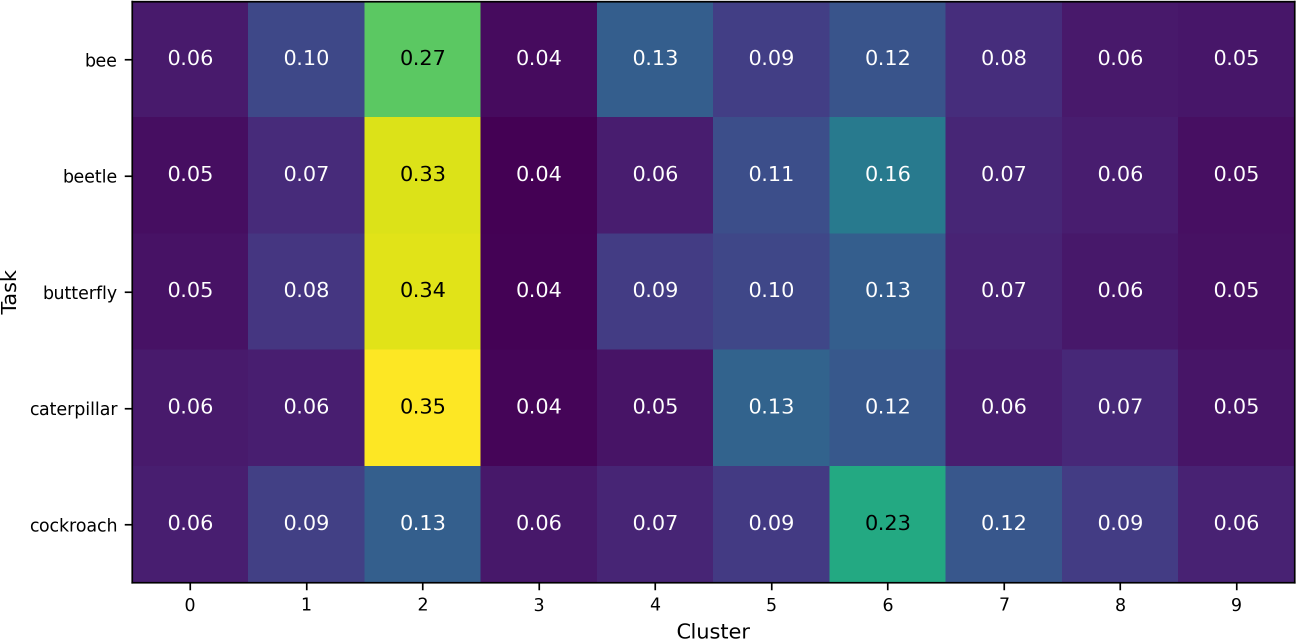}}
        \subcaption{insects}\label{fig_G3_10C_custom_insects}
    \end{minipage}\
    \begin{minipage}[c]{0.32\columnwidth}
        \centerline{\includegraphics[width=\columnwidth]{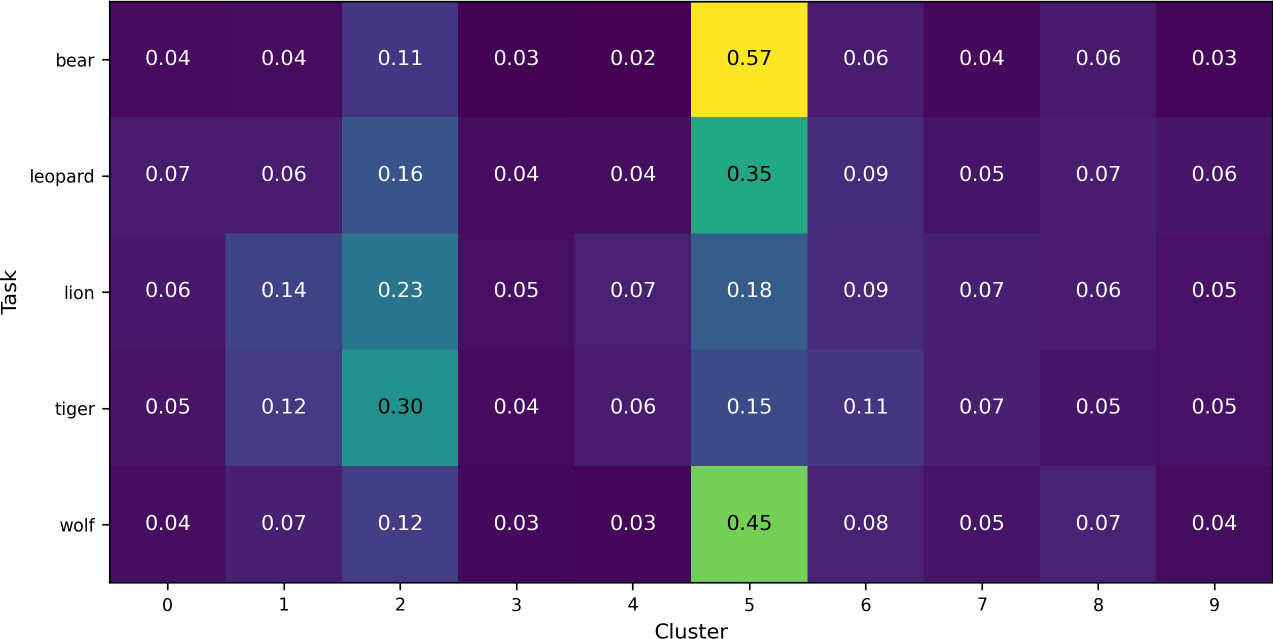}}
        \subcaption{large carnivores}\label{fig_G3_10C_custom_large_carnivores}
    \end{minipage}\\
    \begin{minipage}[c]{0.32\columnwidth}
        \centerline{\includegraphics[width=\columnwidth]{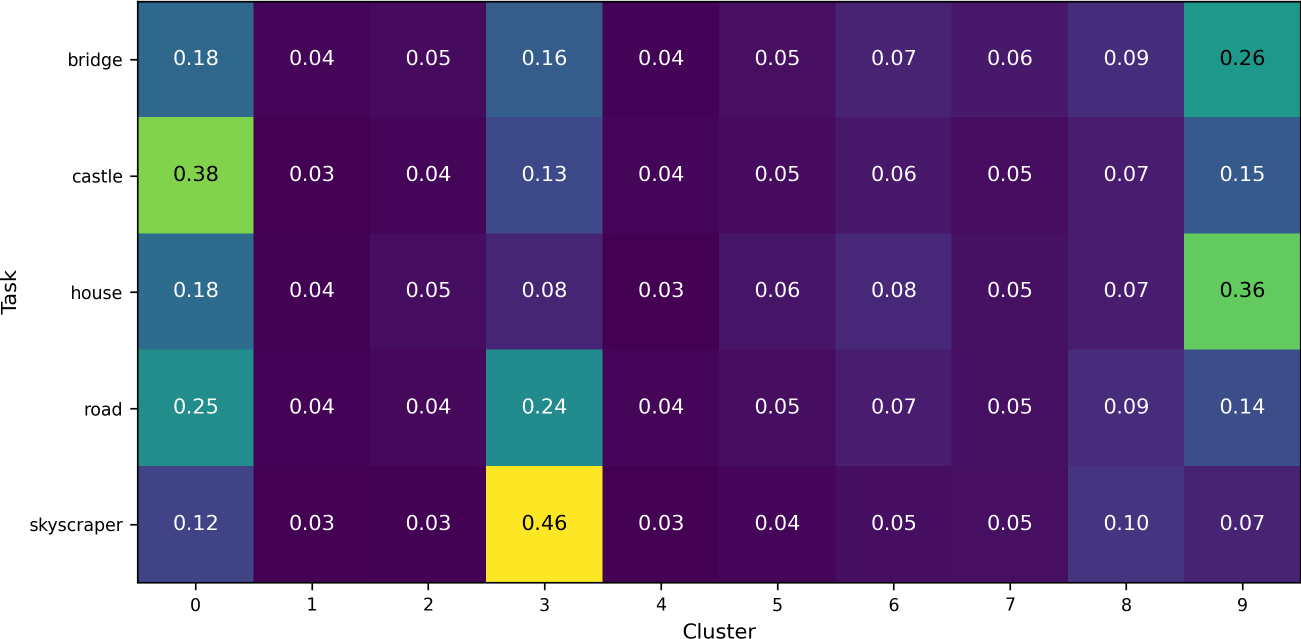}}
        \subcaption{large man-made outdoor things}\label{fig_G3_10C_custom_large_man_made_outdoor_things}
    \end{minipage}\
    \begin{minipage}[c]{0.32\columnwidth}
        \centerline{\includegraphics[width=\columnwidth]{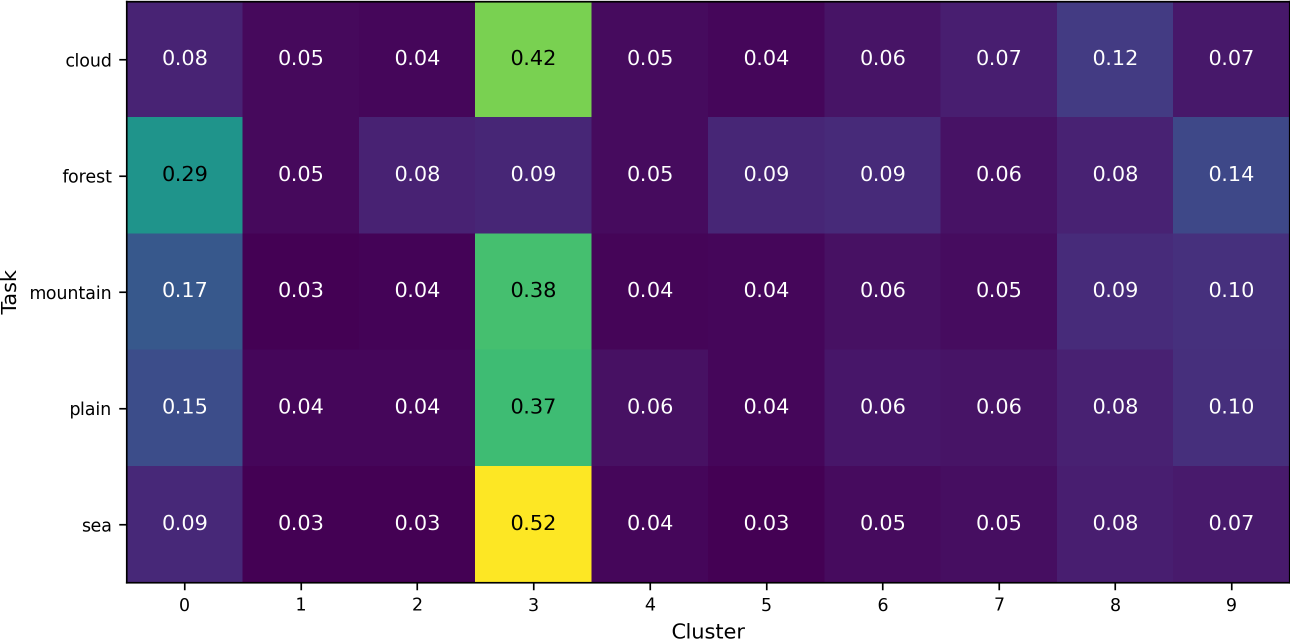}}
        \subcaption{large natural outdoor scenes}\label{fig_G3_10C_custom_large_natural_outdoor_scenes}
    \end{minipage}\
    \begin{minipage}[c]{0.32\columnwidth}
        \centerline{\includegraphics[width=\columnwidth]{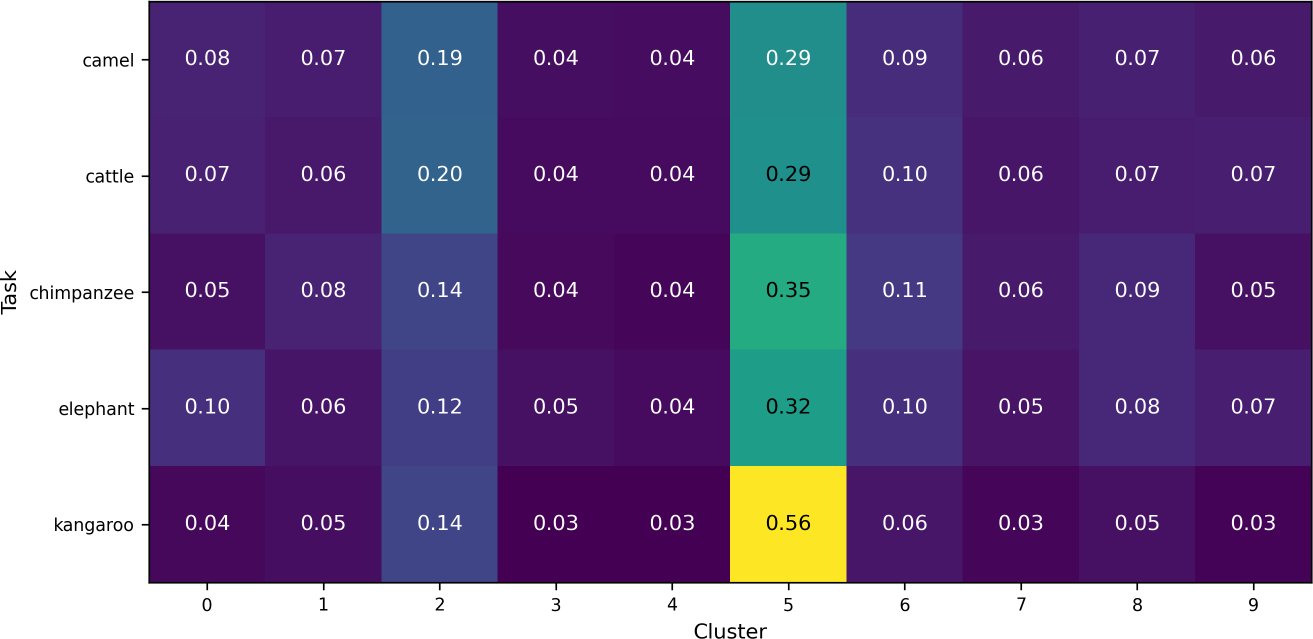}}
        \subcaption{large omnivores and herbivores}\label{fig_G3_10C_custom_large_omnivores_and_herbivores}
    \end{minipage}\\
    \begin{minipage}[c]{0.32\columnwidth}
        \centerline{\includegraphics[width=\columnwidth]{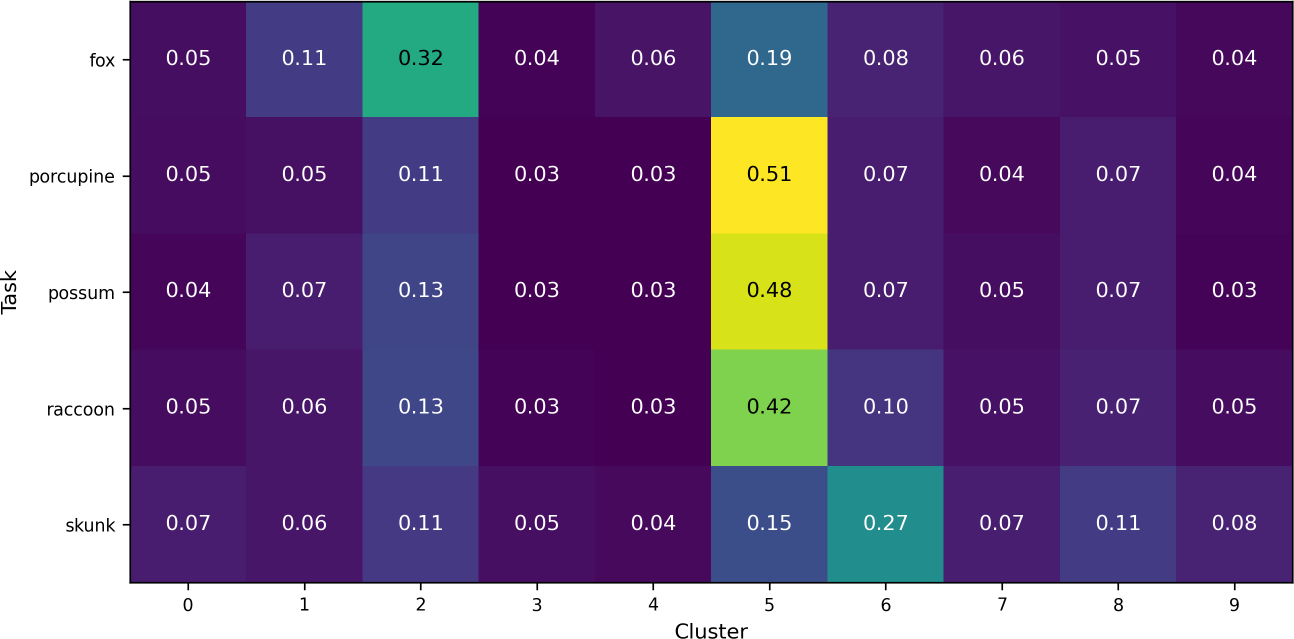}}
        \subcaption{medium-sized mammals}\label{fig_G3_10C_custom_medium_sized_mammals}
    \end{minipage}\
    \begin{minipage}[c]{0.32\columnwidth}
        \centerline{\includegraphics[width=\columnwidth]{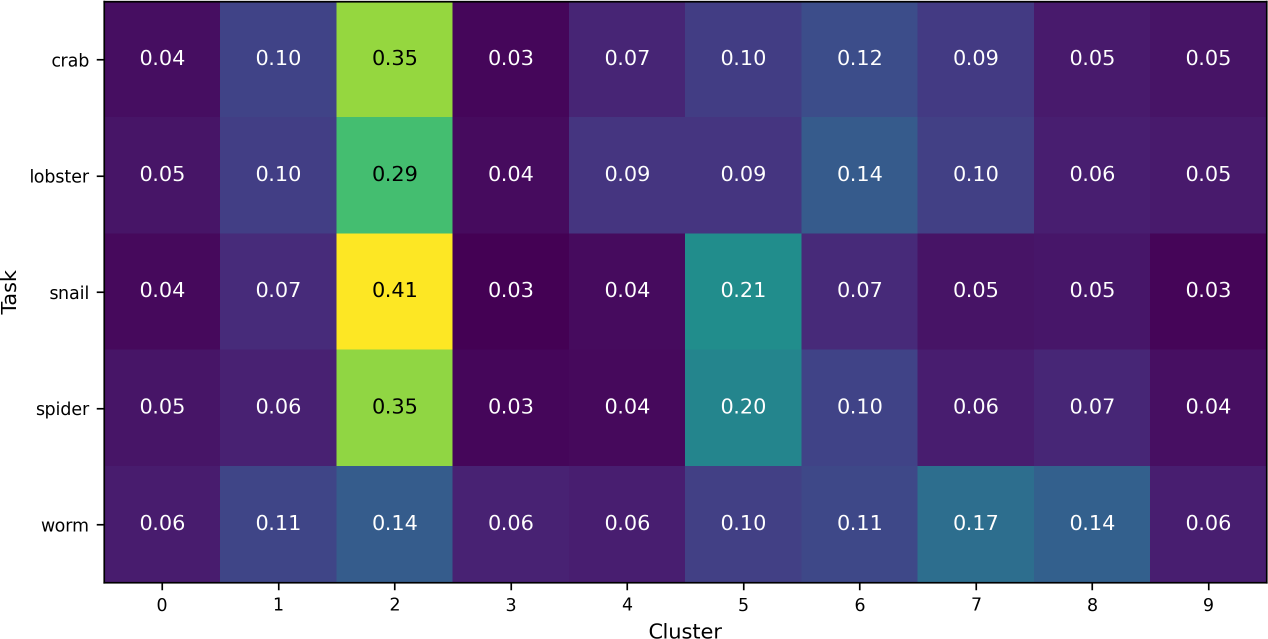}}
        \subcaption{non-insect invertebrates}\label{fig_G3_10C_custom_non_insect_invertebrates}
    \end{minipage}\
    \begin{minipage}[c]{0.32\columnwidth}
        \centerline{\includegraphics[width=\columnwidth]{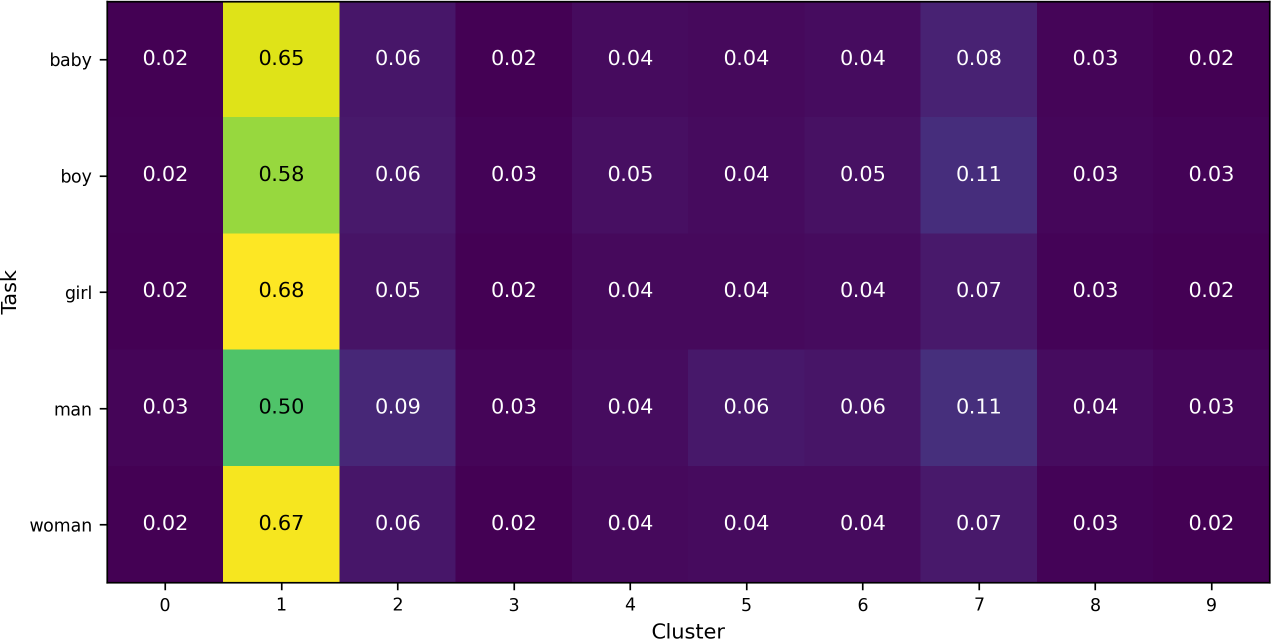}}
        \subcaption{people}\label{fig_G3_10C_custom_people}
    \end{minipage}\\
    \begin{minipage}[c]{0.32\columnwidth}
        \centerline{\includegraphics[width=\columnwidth]{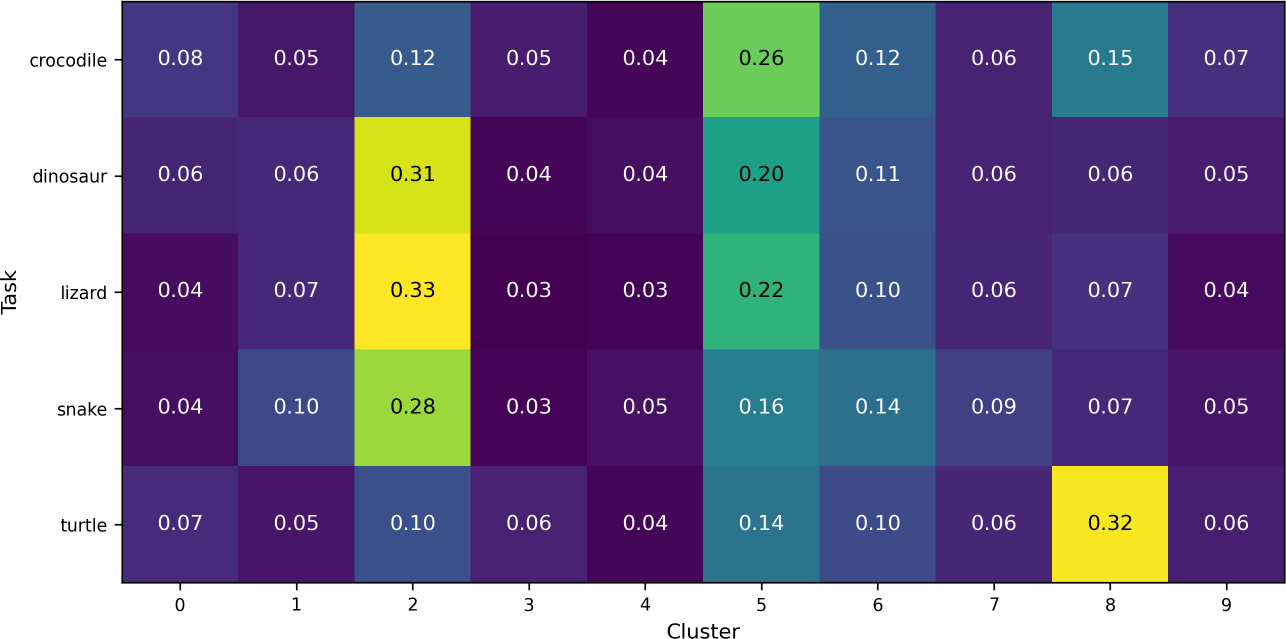}}
        \subcaption{reptiles}\label{fig_G3_10C_custom_reptiles}
    \end{minipage}\
    \begin{minipage}[c]{0.32\columnwidth}
        \centerline{\includegraphics[width=\columnwidth]{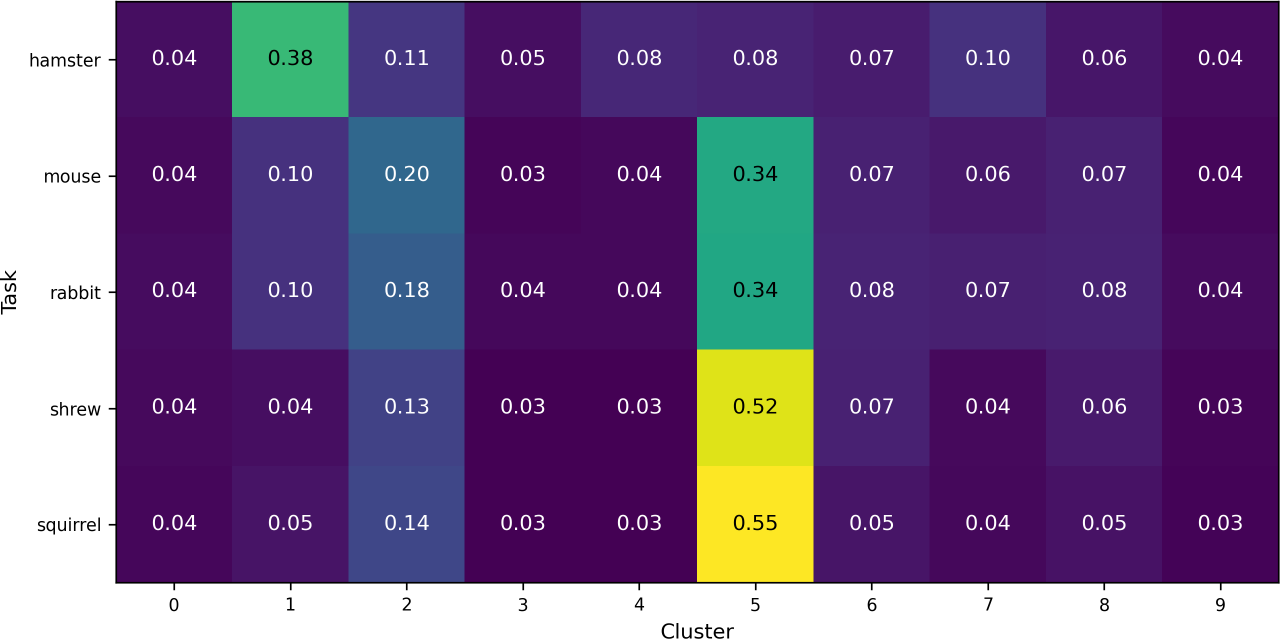}}
        \subcaption{small mammals}\label{fig_G3_10C_custom_small_mammals}
    \end{minipage}\
    \begin{minipage}[c]{0.32\columnwidth}
        \centerline{\includegraphics[width=\columnwidth]{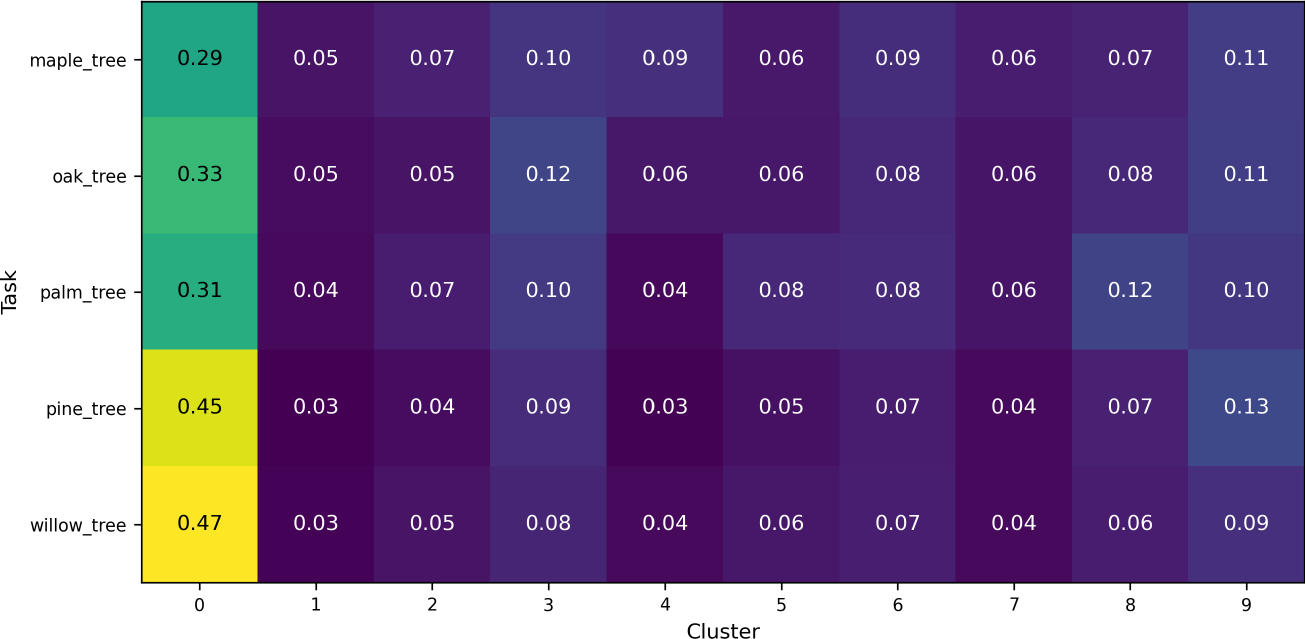}}
        \subcaption{trees}\label{fig_G3_10C_custom_trees}
    \end{minipage}\\
    \begin{minipage}[c]{0.32\columnwidth}
        \centerline{\includegraphics[width=\columnwidth]{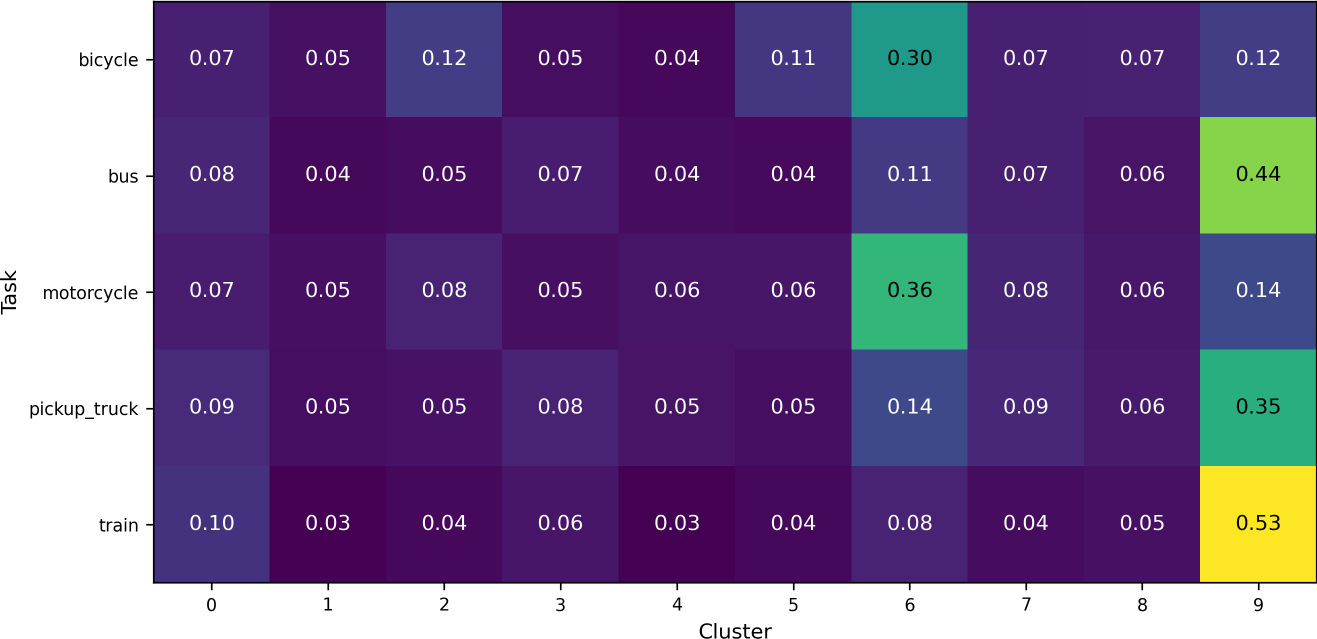}}
        \subcaption{vehicles 1}\label{fig_G3_10C_custom_vehicles_1}
    \end{minipage}\
    \begin{minipage}[c]{0.32\columnwidth}
        \centerline{\includegraphics[width=\columnwidth]{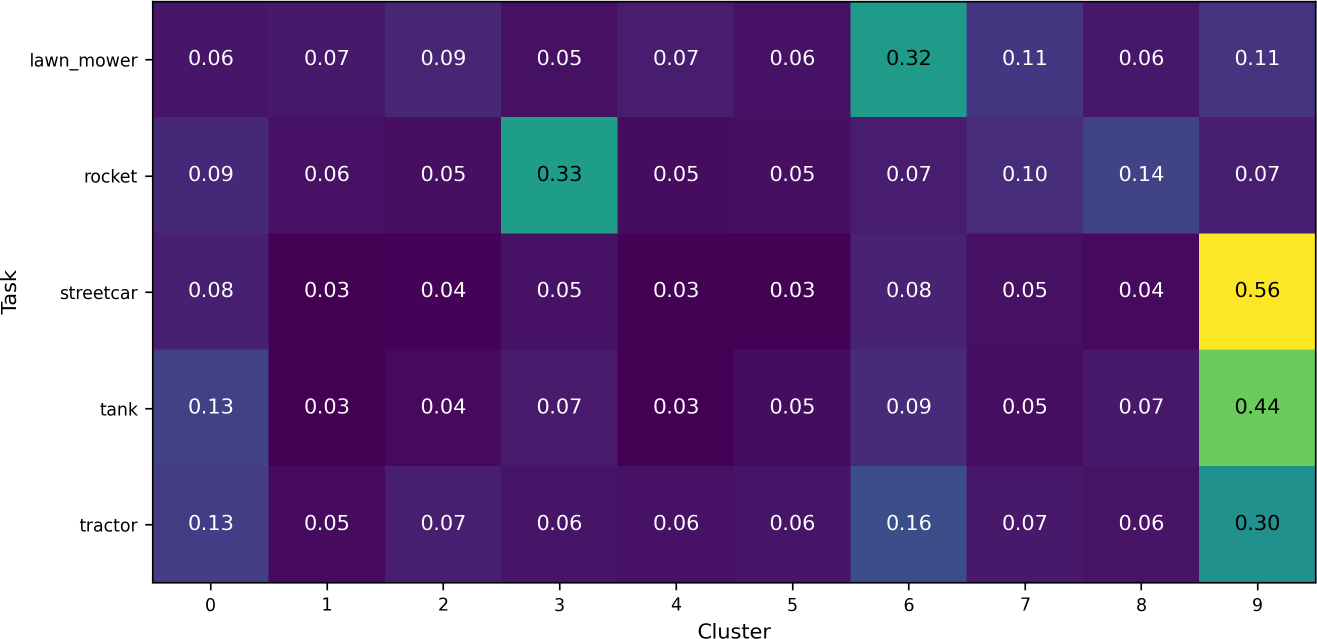}}
        \subcaption{vehicles 2}\label{fig_G3_10C_custom_vehicles_2}
    \end{minipage}\\
    \caption{Task grouping of G3 (100 Tasks of CIFAR100) into $10$ clusters with $F=2$ using Data Maps from the Custom CNN}\label{fig_G3_10C_custom}
\end{figure*}

Figure \ref{fig_G3_20C_custom} and \ref{fig_G3_10C_custom} showcase the task clustering results obtained using the simplified CNN architecture. Surprisingly, the resulting clustering patterns from our custom CNN are remarkably comparable to those achieved with ResNet18. This observation underscores the method's expressiveness, as it effectively captures task relationships without necessitating a high number of learnable parameters.

This intriguing finding opens the possibility of extrapolating clustering outcomes from less expressive models to more complex ones. However, further investigations are required to ascertain the extent and implications of this approach.

\section{Modularity in the Code Structure} \label{appendix_code_structure}

\begin{figure*}[ht]
    \centering
    \centerline{\includegraphics[width=0.8\textwidth]{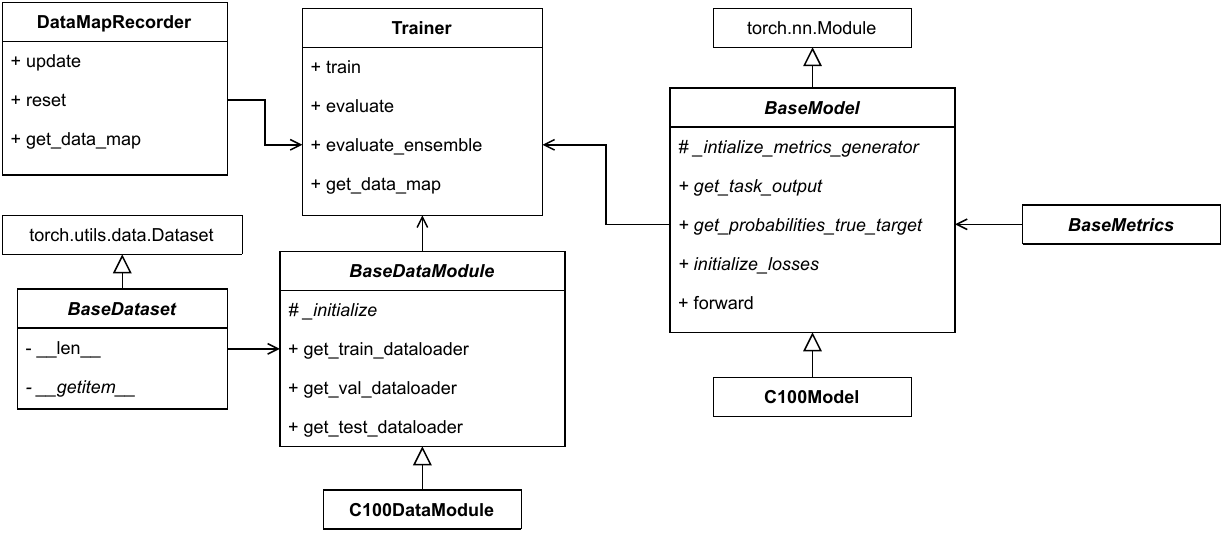}}
    \caption{UML class diagram of our pipeline}
    \label{fig_pipeline}
\end{figure*}

In designing our method, we placed significant emphasis on scalability and modularity. Ensuring scalability without requiring substantial modifications whenever model architecture or tasks change, code modularity, is crucial. To achieve this, we designed a modular pipeline that facilitates easy adaptation to various scenarios with minimal adjustments. The class diagram for our pipeline is depicted in Figure \ref{fig_pipeline}.

Our code architecture is composed of several classes, each with distinct roles. The main class responsible for training models on datasets is the \texttt{Trainer}. Notably, the \texttt{Trainer} employs an object from the \texttt{DataMapRecorder} class, which dynamically updates and records data maps after each iteration. To enhance reusability, we introduce an abstract class called \texttt{BaseModel}, which serves as the blueprint for all models. Similarly, the \texttt{Trainer} expects a \texttt{BaseDataModule} as input, an abstract class encompassing dataset information and instructions for building data loaders.
\begin{wrapfigure}{r}{0.5\textwidth}
\includegraphics[width=0.9\linewidth]{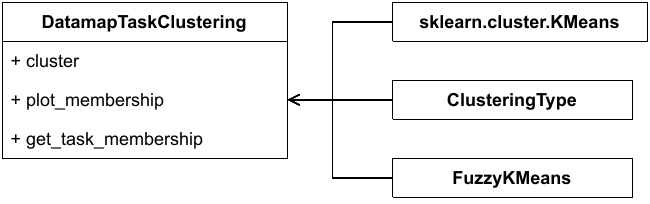} 
\caption{UML class diagram of task clustering classes}
\label{fig_task_clustering}
\end{wrapfigure}
When dealing with new sets of tasks, two steps are necessary. First, we inherit from \texttt{BaseModel}, implementing its abstract methods, italicized in the figure. Second, we provide concrete implementations for the \texttt{BaseDataModule}'s abstract methods. Our code contains examples of these two steps for different datasets such as CIFAR10, CIFAR100, and CelebA.

To train models and extract data maps, only a few lines of code are required, exemplified in Code \ref{code_training}. The resulting data map from training an MTL model is then ready for clustering using our \texttt{DatamapTaskClustering} class, depicted in Figure \ref{fig_task_clustering}, as shown in Code \ref{code_clustering}. This design simplifies the integration process for new tasks, models, or datasets, effectively enhancing the overall modularity of our pipeline and method.
\begin{lstlisting}[language=Python,caption={Start training and record the data maps},label=code_training] 
# create an instance of the data module
data_module = C100DataModule(...)

# Now initialize a model
model = C100Model(...)

# Finally, initialize a trainer object
trainer = Trainer(model, data_module, args= args, ... )

# start training
trainer.train()
\end{lstlisting}

\begin{lstlisting}[language=Python, caption={Generate task memberships from data maps}, label=code_clustering]
# To further see the task membership, we do the below steps

# get the datamap from the trainer
dm_mtl = trainer.get_data_map()

# initialize a cluster estimator
cluster_estimator = DatamapTaskClustering(dm_mtl, n_clusters, ...)

# get the results
mtl_task_weights = cluster_estimator.cluster(...)
\end{lstlisting}
\end{document}